\documentclass[10pt,journal,compsoc]{IEEEtran}
\ifCLASSOPTIONcompsoc
  \usepackage[nocompress]{cite}
\else
  \usepackage{cite}
\fi
\ifCLASSINFOpdf
   \usepackage[pdftex]{graphicx}
   \graphicspath{{figs/}}
   \DeclareGraphicsExtensions{.pdf,.jpeg,.png}
\else
   \usepackage[dvips]{graphicx}
   \graphicspath{{figs/}}
   \DeclareGraphicsExtensions{.eps}
\fi
\usepackage{amsmath}
\usepackage{amsfonts}

\ifCLASSOPTIONcompsoc
 \usepackage[caption=false,font=footnotesize,labelfont=sf,textfont=sf]{subfig}
\else
 \usepackage[caption=false,font=footnotesize]{subfig}
\fi
\usepackage{url}
\usepackage{booktabs}
\usepackage{threeparttable}  % 表格加脚注
\usepackage{multirow}  % 表格中支持跨行
\usepackage{makecell}  % 表格内换行
\usepackage{tabularx}

\usepackage{pifont}

\usepackage{color}

\makeatletter
\let\NAT@parse\undefined
\makeatother
\usepackage{hyperref}  % hyperref still needs to be put at the end!

\begin{document}
%
% paper title
% Titles are generally capitalized except for words such as a, an, and, as,
% at, but, by, for, in, nor, of, on, or, the, to and up, which are usually
% not capitalized unless they are the first or last word of the title.
% Linebreaks \\ can be used within to get better formatting as desired.
% Do not put math or special symbols in the title.
% \title{Survey of Dropout Methods in Neural Models and their Applications in Recommender Systems}
% \title{A Survey on Dropout Methods and Their Applications in Neural Recommendation Models}
\title{A Survey on Dropout Methods and Experimental Verification in Recommendation}

\author{Yangkun~Li,~\IEEEmembership{Member,~IEEE,}
        Weizhi~Ma,~\IEEEmembership{Member,~IEEE,}
        Chong~Chen,~\IEEEmembership{Member,~IEEE,}
        Min~Zhang,~\IEEEmembership{Member,~IEEE,}
        Yiqun~Liu,~\IEEEmembership{Senior~Member,~IEEE,}
        Shaoping~Ma,
        and~Yuekui~Yang% <-this % stops a space
\IEEEcompsocitemizethanks{\IEEEcompsocthanksitem Y. Li, C. Chen, M. Zhang, Y. Liu, S. Ma, and Y. Yang are with Department of Computer Science and Technology, Institute for Artificial Intelligence, Beijing National Research Center for Information Science and Technology, Tsinghua University, Beijing 100084, China.\protect\\
% note need leading \protect in front of \\ to get a newline within \thanks as
% \\ is fragile and will error, could use \hfil\break instead.
E-mail: lyk21@mails.tsinghua.edu.cn, z-m@tsinghua.edu.cn
\IEEEcompsocthanksitem W. Ma is with Institute for AI Industry Research (AIR), Tsinghua University, Beijing 100084, China.
\IEEEcompsocthanksitem Y. Yang is also with Tencent AI Platform Department, China.
\IEEEcompsocthanksitem Min Zhang is the corresponding author.}% <-this % stops an unwanted space
}

% note the % following the last \IEEEmembership and also \thanks - 
% these prevent an unwanted space from occurring between the last author name
% and the end of the author line. i.e., if you had this:
% 
% \author{....lastname \thanks{...} \thanks{...} }
%                     ^------------^------------^----Do not want these spaces!
%
% a space would be appended to the last name and could cause every name on that
% line to be shifted left slightly. This is one of those "LaTeX things". For
% instance, "\textbf{A} \textbf{B}" will typeset as "A B" not "AB". To get
% "AB" then you have to do: "\textbf{A}\textbf{B}"
% \thanks is no different in this regard, so shield the last } of each \thanks
% that ends a line with a % and do not let a space in before the next \thanks.
% Spaces after \IEEEmembership other than the last one are OK (and needed) as
% you are supposed to have spaces between the names. For what it is worth,
% this is a minor point as most people would not even notice if the said evil
% space somehow managed to creep in.

% The paper headers
\markboth{Preprint. Under review.}%
{\MakeLowercase{\textit{Y. Li et al.}}}
% The only time the second header will appear is for the odd numbered pages
% after the title page when using the twoside option.
% 
% *** Note that you probably will NOT want to include the author's ***
% *** name in the headers of peer review papers.                   ***
% You can use \ifCLASSOPTIONpeerreview for conditional compilation here if
% you desire.

% The publisher's ID mark at the bottom of the page is less important with
% Computer Society journal papers as those publications place the marks
% outside of the main text columns and, therefore, unlike regular IEEE
% journals, the available text space is not reduced by their presence.
% If you want to put a publisher's ID mark on the page you can do it like
% this:
%\IEEEpubid{0000--0000/00\$00.00~\copyright~2015 IEEE}
\IEEEpubid{\begin{tabular}[t]{@{}c@{}}\copyright~2021 IEEE. Personal use of this material is permitted. Permission from IEEE must be obtained for all other uses, in any current or future media,\\including reprinting/republishing this material for advertising or promotional purposes, creating new collective works,\\for resale or redistribution to servers or lists, or reuse of any copyrighted component of this work in other works.\end{tabular}}
% or like this to get the Computer Society new two part style.
%\IEEEpubid{\makebox[\columnwidth]{\hfill 0000--0000/00/\$00.00~\copyright~2015 IEEE}%
%\hspace{\columnsep}\makebox[\columnwidth]{Published by the IEEE Computer Society\hfill}}
% Remember, if you use this you must call \IEEEpubidadjcol in the second
% column for its text to clear the IEEEpubid mark (Computer Society jorunal
% papers don't need this extra clearance.)

% use for special paper notices
%\IEEEspecialpapernotice{(Invited Paper)}

% for Computer Society papers, we must declare the abstract and index terms
% PRIOR to the title within the \IEEEtitleabstractindextext IEEEtran
% command as these need to go into the title area created by \maketitle.
% As a general rule, do not put math, special symbols or citations
% in the abstract or keywords.
\IEEEtitleabstractindextext{%
\begin{abstract}

Overfitting is a common problem in machine learning, which means the model too closely fits the training data while performing poorly in the test data. Among various methods of coping with overfitting, dropout is one of the representative ways. From randomly dropping neurons to dropping neural structures, dropout has achieved great success in improving model performances. Although various dropout methods have been designed and widely applied in past years, their effectiveness, application scenarios, and contributions have not been comprehensively summarized and empirically compared by far. It is the right time to make a comprehensive survey.

In this paper, we systematically review previous dropout methods and classify them into three major categories according to the stage where dropout operation is performed. Specifically, more than seventy dropout methods published in top AI conferences or journals (e.g., TKDE, KDD, TheWebConf, SIGIR) are involved. The designed taxonomy is easy to understand and capable of including new dropout methods. Then, we further discuss their application scenarios, connections, and contributions. To verify the effectiveness of distinct dropout methods, extensive experiments are conducted on recommendation scenarios with abundant heterogeneous information. Finally, we propose some open problems and potential research directions about dropout that worth to be further explored.

\end{abstract}

% Note that keywords are not normally used for peerreview papers.
}

% make the title area
\maketitle

% To allow for easy dual compilation without having to reenter the
% abstract/keywords data, the \IEEEtitleabstractindextext text will
% not be used in maketitle, but will appear (i.e., to be "transported")
% here as \IEEEdisplaynontitleabstractindextext when the compsoc 
% or transmag modes are not selected <OR> if conference mode is selected 
% - because all conference papers position the abstract like regular
% papers do.
\IEEEdisplaynontitleabstractindextext
% \IEEEdisplaynontitleabstractindextext has no effect when using
% compsoc or transmag under a non-conference mode.

% For peer review papers, you can put extra information on the cover
% page as needed:
% \ifCLASSOPTIONpeerreview
% \begin{center} \bfseries EDICS Category: 3-BBND \end{center}
% \fi
%
% For peerreview papers, this IEEEtran command inserts a page break and
% creates the second title. It will be ignored for other modes.
\IEEEpeerreviewmaketitle

% !TeX root = ../main.tex

\IEEEraisesectionheading{\section{Introduction}\label{sec:introduction}}

\subsection{Backgrounds}\label{subsec:backgrounds}

\IEEEPARstart{O}{verfitting} is a common problem in the training process of neural network models \cite{dietterich1995overfitting}. Due to the large number of parameters and strong fitting ability, most neural models perform well on the training set, while they may perform poorly on the test set. Some methods have been proposed in previous studies to address the overfitting problem, such as adding a regularization term to penalize the total size of model parameters \cite{van2017l2} and applying Batch Normalization \cite{ioffe2015batch} or Weight Normalization \cite{salimans2016weight} to regularize deep neural networks.

In 2012, Hinton et al. proposed Dropout \cite{hinton2012improving} to cope with overfitting. The idea is to randomly drop neurons of the neural network during training. This is, in each parameter update, only part of the model parameters will be updated. Through this process, it can prevent complex co-adaptations of neurons on training data. It is important to note that in the testing phase, dropout must be disabled, and the whole network is used for prediction. From this beginning, numerous dropout-based training methods have been proposed and achieved better performances.

% wang2013fast, wan2013regularization, goodfellow2013maxout, kingma2015variational, gal2016dropout, smith2016gradual, li2016improved, gal2017concrete, rennie2014annealed, morerio2017curriculum, gomez2018targeted, gomez2019learning, salehinejad2019ising
% ghiasi2018dropblock, khan2018regularization, cai2019effective, hou2019weighted, pham2021autodropout
% gastaldi2017shake, yamada2019shakedrop, larsson2016fractalnet, zoph2018learning, singh2016swapout
% gal2016theoretically, semeniuta2016recurrent, krueger2016zoneout, merity2018regularizing, zolna2018fraternal
% , ghazvininejad2019mask, devlin2019bert, sun2019ernie, cui2019pre, wu2019mask, ye2019align, zhang2020pegasus, zhou2020s3, zhong2020random, singh2017hide, singh2018hide, chen2020gridmask, zhang2017mixup, verma2019manifold, yun2019cutmix, walawalkar2020attentive, chen2018fastgcn, huang2018adaptive, zou2019layer, feng2020graph, velivckovic2017graph, rong2019dropedge
% gomez2019learning, salehinejad2019ising

In the beginning, this dropout-based training method was applied only to  fully connected layers \cite{hinton2012improving, srivastava2014dropout, ba2013adaptive}. Later, it is extended to more network structures such as convolutional layers in convolutional neural networks (CNNs) \cite{tompson2015efficient, wu2015towards, park2016analysis}, residual networks (ResNet)\cite{huang2016deep, kang2016shakeout, li2016whiteout}, recurrent layers in recurrent neural networks (RNNs) \cite{ pham2014dropout, zaremba2014recurrent, moon2015rnndrop}, etc. In terms of the stage where the dropout operation is performed, there are not only dropout of model structure, but also dropout of input information \cite{sennrich2016edinburgh, devries2017improved, hamilton2017inductive} and dropout of embeddings \cite{ volkovs2017dropoutnet, shi2018attention, shi2019adaptive}. In terms of contributions, dropout methods were first used only to prevent overfitting. Besides, many studies have been made to explore other aspects of its usefulness, such as model compression \cite{molchanov2017variational, neklyudov2017structured, gomez2018targeted}, model uncertainty measurement \cite{gal2016dropout}, data augmentation \cite{bouthillier2015dropout}, enhancing data representations in the pre-training phase \cite{ devlin2019bert}, and prevention of over-smoothing problem in graph neural networks \cite{rong2019dropedge}.

Despite their wide applications, a dropout method that works for one model structure may have no significant effect on another. For example, the use of standard dropout in CNNs does not improve the effect significantly \cite{tompson2015efficient}. The same is true for the direct use of standard dropout to the recurrent connections of RNNs \cite{zaremba2014recurrent}. Dropout at different stages of a machine learning task achieves different purposes. Therefore, in this paper, we classify these wide varieties of dropout methods and summarize their effectiveness, application scenarios, connections, and contributions.

Most of the methods with commonality also fail to compare their results under the same scenario. Since dropout methods are applied to different forms of input information and model structures, a proper comparison scenario should have rich heterogeneous input information and a variety of different model structures. With the rapid development of the internet, various recommendation models have been proposed and widely used to extract user and item features to improve user experience in many online scenarios \cite{zhang2016collaborative, zhao2017meta}. They utilize various heterogeneous information, such as: user and item interaction histories, content features, or social connections \cite{hu2018leveraging}. Such a variety of models and heterogeneous input information provides a suitable environment for our comparisons and verification of different dropout methods. Therefore, we evaluate and compare different dropout methods under the recommendation scenario to fairly compare the effect of these dropout methods, providing references for future works related to dropout.

\vspace{-2pt}

\subsection{Contributions}\label{subsec:contributions}
Our contributions in this paper are three fold:

First, we provide a comprehensive review of more than seventy dropout methods. We propose a new taxonomy based on the stage where dropout operations are performed in machine learning tasks. Each category is then supplemented with operation granularity and application scenarios for more detailed classification and discussion.
% These methods perform dropout operations on different stages of machine learning tasks: on model structure, on embeddings, and on the input information. Accordingly, we classify them into three major categories: dropping model structure, dropping embeddings, and dropping input information.

Second, we discuss the connections between dropout methods of different categories and summarize their various contributions other than preventting overfitting.

Third, we experimentally investigate the effect of dropout methods in recommendation scenarios. The rich heterogeneous input information makes recommendation scenarios suitable for the comparison between different types of dropout methods. We verify and compare dropout methods' effectiveness under a unified experimental environment. And finally, we provide potential research directions about dropout methods.
 
\subsection{Outline}\label{subsec:outline}
The organization of this paper is as follows: Section \ref{sec:related works} introduces background concepts of dropout methods and recommendation systems. In Section \ref{sec:survey}, we review dropout methods according to the stage where the dropout operation is performed in a machine learning task. We summarize their applications on different neural models and discuss their connections. In Section \ref{sec:discussion}, we analyze the contributions of dropout methods other than preventing overfitting. In Section \ref{sec:experiments}, we present an experimental verification of dropout methods on recommendation models.
% We select five representative recommendation models, for each of which we verify and compare the effectiveness of three major types of dropout methods. 
In Section \ref{sec:discussion2}, we provide further discussions on dropout methods and analyze potential research directions in this field. Finally, we conclude the entire paper in Section \ref{sec:conclusion}.

\vspace{-2pt}

\section{Background Concepts}\label{sec:related works}

This section introduces the fundamental knowledge of dropout and recommender systems by summarizing related works of these two fields.

\subsection{Dropout methods}

Dropout is a class of training methods effectively coping with overfitting. Hinton et al. \cite{hinton2012improving} proposed the original dropout, whose idea is to randomly drop neurons of the neural network during training. Through this process, it prevents complex co-adaptations of neurons on training data. From this beginning, numerous drop-based methods have been proposed, achieving better results and higher performances. For exapmle, DropConnect \cite{wan2013regularization} randomly drops neuron connections instead of neurons, while Annealed Dropout \cite{rennie2014annealed} and Curriculum Dropout \cite{morerio2017curriculum} adjust dropout ratio throughout the training process. % Targeted Dropout \cite{gomez2018targeted} and Ising-Dropout \cite{salehinejad2019ising} drop certain neurons for neural pruning.

Besides dropping individual neurons, a series of dropout methods that drop neuron groups are proposed for certain neural model structures. For CNNs, SpatialDropout \cite{tompson2015efficient} randomly drops feature map, and DropBlock \cite{ghiasi2018dropblock} drops continuous region of neurons to prevent overfitting. For RNNs, early applications of dropout \cite{pham2014dropout} only drop feed-forward connections, in order to preserve the memory ability of RNN. Later approaches including RNNDrop \cite{moon2015rnndrop} and Recurrent Dropout \cite{semeniuta2016recurrent} allow dropping recurrent connections as well. For ResNets, there are also specified dropout methods such as Stochastic Depth \cite{huang2016deep} and ShakeDrop \cite{yamada2019shakedrop}.

Except for dropping model structure during training, dropping input information or embeddings of input data is also applied in several scenarios. DropoutNet \cite{volkovs2017dropoutnet} and ACCM \cite{shi2018attention} drop embeddings of users and items in recommendation scenarios to handle the cold-start problem. BERT \cite{devlin2019bert} randomly masks tokens at pre-training stage, enhancing data representations in NLP tasks. CutOut \cite{devries2017improved} and GridMask \cite{chen2020gridmask}  randomly drop part of the input images during training, serving as a regularization and data augmentation technique. GraphSAGE \cite{hamilton2017inductive} and DropEdge \cite{rong2019dropedge} randomly drop nodes or edges during GCN training, preventing overfitting and over-smoothing problem in GCN.

The former survey about dropout methods \cite{labach2019survey} was made by Labach et al. in 2019, which is the only comprehensive survey about this topic. It reviews dropout methods from the aspect of neural models that dropout methods are performed on, including fully connected layers, convolutional layers and recurrent layers.
% Then supplemented with other two applications of dropout methods: model compression and uncertainty measurement. 

Our work has three major differences with this former one. 
First, we cover a wider range of dropout methods, including those proposed in recent three years. Especially new methods published in the top AI conferences.
Second, we present a more precise and general classification. In the former survey, dropout methods were classified according to the neural models they perform on, which means classification requires that when a new model structure appears, the corresponding dropout methods are classified into a new category. Nevertheless, dropout methods themselves may be similar to the ones that already existed before, so it seems cumbersome to put them into a new category just because their applications change. Our work classifies dropout methods according to the stage where dropout operations are performed in machine learning tasks. With this setting, new methods must fit into one existing category, making our classification more reasonable. We also review these methods from the perspective of application scenarios, their interconnections, and contributions other than preventing overfitting. 
Third, there is no experimental comparison of the effectiveness of dropout methods in \cite{labach2019survey}, while we experimentally verify and compare their effectiveness under recommendation scenarios.

\subsection{Recommender Systems}

In terms of the input information types, recommender systems are mainly categrized into collaborative filtering (CF) based, content-based (CB), and hybrid \cite{adomavicius2005toward}. CF based recommender systems make predictions based on the interaction histories of users and items \cite{su2009survey, li2018deep}, while CB recommender systems use content feature of users and items \cite{sun2019research}. Hybrid recommender systems use multiple types of input information to extract interaction similarity as well as content similarity, such as users' social networks \cite{massa2007trust, jamali2009trustwalker} and item reviews \cite{zheng2017joint, xu2018exploiting}. In recent years, some recommendation algorithms specified for certain tasks utilize specific form of input data, such as in sequential recommendation the input data is structured as temporal sequences \cite{hidasi2015session, kang2018self}, and in graph recommendation the input data is structured as graphs \cite{wang2019knowledge, he2020lightgcn}. However, the input information in real world scenarios may not be sufficient for the recommender systems to make good predictions, and this problem is called \textit{Cold Start}. Works addressing cold-start problem have been emerging in recent years \cite{xu2021multi, qian2020attribute, zhu2019addressing, zhang2020deep, li2019both, lu2020meta}.

Such rich variety of input information in recommendation scenario provides a good environment for the verification and comparison of different dropout methods. Besides conducting a survey on dropout methods, we also do experimental verification and put them under the same scenario for comparison.

% !TeX root = ../main.tex

\section{Survey of Dropout Methods}\label{sec:survey}

\begin{figure}
  \centering
  \includegraphics[width=\linewidth]{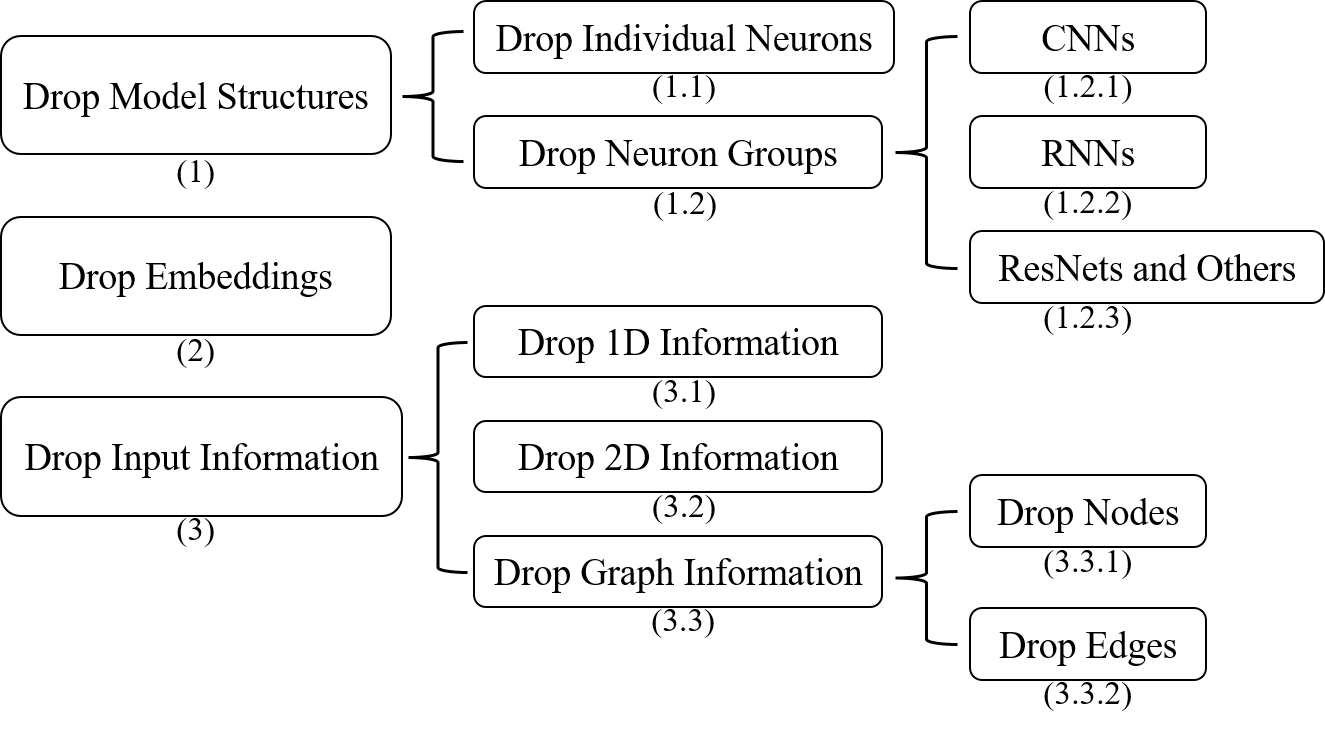}
  \vspace{-16pt}
  \caption{Classification of dropout methods. The number below each box is referring to the corresponding category of dropout methods.}
  \label{fig:classification}
  \vspace{-4pt}
\end{figure}

In this section, we review papers of dropout methods. Based on the stage where the dropout operation is performed in a machine learning algorithm, we classify them into three main categories: drop model structures (Section \ref{subsec:drop structures}), drop embeddings (Section \ref{subsec:drop embeddings}), and drop input information (Section \ref{subsec:drop inputs}) as shown in Figure \ref{fig:classification}. We introduce how these methods perform dropout, their effectiveness, and their applications in different neural models. Finally, we discuss the interconnections between dropout methods in different categories (Section \ref{subsec:interconnections}).

\subsection{Drop Model Structures}\label{subsec:drop structures}

Methods in this category drop model structures, which is, randomly setting part of neuron outputs to zero or other value during training. We first introduce methods dropping individual neurons, then methods dropping neuron groups.

\subsubsection{Drop Individual Neurons}\label{subsubsec:drop single neurons}

Hinton et al. \cite{hinton2012improving} first proposed the standard dropout in 2012, and further detailed in \cite{srivastava2014dropout}. Specifically, for a neural network with $L$ hidden layers, $l\in {1, \dots, L}$, $z^{(l)}$ and $y^{(l)}$ are the input and output of the $l$th layer respectively, and $y^{(0)}$ is the input of the network. $W^{(l)}$ and $b^{(l)}$ are the weight matrix and bias vectors (biases) of the $l$th layer, respectively. The standard feed-forward operation when dropout is not performed is
\begin{equation} \label{eq:dropout 1}
    \begin{aligned}
        z_i^{(l+1)} = \mathbf{w}_i^{(l+1)}\mathbf{y}^{(l)} + b_i^{(l+1)},\quad
        y_i^{(l+1)} = f(z_i^{(l+1)})
    \end{aligned}
    \vspace{-2pt}
\end{equation}
Dropout sets a proportion of $p$ of the neuron outputs to zero during training, where $p\in (0, 1)$ is the dropout ratio. When performing dropout, the feed-forward operation becomes
\begin{equation} \label{eq:dropout 2}
    \begin{aligned}
        r_j^{(l)} &\sim \mathrm{Bernoulli}(p),\quad
        \Tilde{\mathbf{y}}^{(l)} = \mathbf{r}^{(l)} * \mathbf{y}^{(l)} \\
        z_i^{(l+1)} &= \mathbf{w}_i^{(l+1)}\Tilde{\mathbf{y}}^{(l)} + b_i^{(l+1)},\quad
        y_i^{(l+1)} = f(z_i^{(l+1)})
    \end{aligned}
    \vspace{-2pt}
\end{equation}
During testing no dropout is performed so all neurons have output. To keep training and testing conditions consistent, all the weights need to be multiplied by $p$ during testing, i.e., take $\mathbf{W}_{test}^{(l)} = p\mathbf{W}^{(l)}$; or it multiplies all outputs by $1/p$ during training so that the expected output is consistent with testing time. Srivastava et al. \cite{srivastava2014dropout} also mentioned that besides generating the dropout mask from Bernoulli distribution, it is also possible to generate it from a continuous distribution with the same expectation and variance, such as Gaussian distribution $\mathcal{N}\sim (1, \alpha)$, $\alpha=p/(1-p)$. The standard dropout has achieved good performance since it was proposed, and the authors showed the effectiveness of standard dropout on image classification tasks \cite{krizhevsky2012imagenet} in 2013.

Following the standard dropout, a series of methods that randomly drop individual neurons have been proposed.

Ba et al. \cite{ba2013adaptive} proposed Standout in 2013. The method treats dropout as a Bayesian learning process. A Bayesian network (Belief Network) is added above the original network to control the dropout ratio:
\begin{equation} \label{eq:standout 1}
    \begin{aligned}
    y = f(\mathbf{Wx})\circ m, \quad m\sim \mathrm{Bernoulli}(g(\mathbf{W}_s \mathbf{x}))
    \end{aligned}
    \vspace{-4pt}
\end{equation}
where $\mathbf{W_s}$ and $g(\cdot)$ is the weight and the activation function of each layer of the Bayesian network. In application the authors find that $\mathbf{W_s}$ can be chosen as an affine transformation of $\mathbf{W}$, and the test output $y$ is computed as
\begin{equation} \label{eq:standout 2}
    \begin{aligned}
    \mathbf{W}_s = \alpha \mathbf{W} + \beta,\quad
    y = f(\mathbf{Wx})\circ g(\mathbf{W}_s \mathbf{x})
    \end{aligned}
    \vspace{-4pt}
\end{equation}
% So the test output is
% \begin{equation} \label{eq:standout 3}
%     \begin{aligned}
%     y = f(\mathbf{Wx})\circ g(\mathbf{W}_s \mathbf{x})
%     \end{aligned}
%     \vspace{-2pt}
% \end{equation}

Wang and Manning \cite{wang2013fast} proposed Fast Dropout in 2013. In standard dropout, only one of the possible network structures is sampled at a time; a proportion of $p$ neurons are not trained in each epoch, making the network's training slower. Fast Dropout, on the other hand, explains dropout from a Bayesian perspective, showing that the output of a layer that has undergone dropout can be considered as sampling from a potential approximate Gaussian distribution. We can then sample directly from this distribution to obtain results or use its parameters to propagate information about the entire dropout set. This allows for faster training than the standard dropout and is also known as Gaussian Dropout.

\begin{table*}[ht]
    \centering
    \newcolumntype{Y}{>{\raggedleft\arraybackslash}X}  %X单元格居左, Y单元格居右
	\newcolumntype{Z}{>{\centering\arraybackslash}X}  %Z单元格居中
    \caption{Table of methods that drop model structures.}
    \vspace{-6pt}
\begin{threeparttable}
    \begin{tabular}{llllll}
        \toprule
Method & Year & Category & Brief Description & \makecell[l]{Original\\Scenario} & Source \\
        \midrule
Dropout\cite{hinton2012improving, srivastava2014dropout} & 2012 & 1.1\dag & Randomly drop neurons & FCL* & JMLR \\
Standout\cite{ba2013adaptive} & 2013 & 1.1 & Add a Bayesian NN to control the dropout ratio & FCL & NeurIPS  \\
Fast Dropout\cite{wang2013fast} & 2013 & 1.1 & Sample outputs directly from a distribution & FCL & ICML     \\
DropConnect\cite{wan2013regularization} & 2013 & 1.1 & Drop weights instead of neurons & FCL & ICML \\
Maxout\cite{goodfellow2013maxout} & 2013 & 1.1 & Computes several outputs for each input & FCL & ICML \\
Annealed Dropout\cite{rennie2014annealed} & 2014 & 1.1 & Dropout ratio decreases with training epochs & FCL & SLT \\
Variational Dropout\cite{kingma2015variational} & 2015 & 1.1 & Dropout ratio can be learned in training & FCL & NeurIPS \\
\rule{-2.3pt}{12pt}
Monte Carlo Dropout\cite{gal2016dropout} & 2016 & 1.1 & \makecell[l]{Intepret dropout as a Bayesian approximation\\of deep Gaussian process} & FCL & ICML \\
\rule{-2.3pt}{7pt}
DropIn\cite{smith2016gradual} & 2016 & 1.1 & Pass dropped values directly to the next layer & FCL & CVPR \\
Evolutional Dropout\cite{li2016improved} & 2016 & 1.1 & Calculate dropout ratio from input & FCL & NeurIPS  \\
\rule{-2.3pt}{12pt}
Concrete Dropout\cite{gal2017concrete} & 2017 & 1.1 & \makecell[l]{Automatically adjust dropout ratio compared\\to Monte Carlo Dropout} & FCL & NeurIPS  \\
\rule{-2.3pt}{7pt}
Curriculum Dropout\cite{morerio2017curriculum} & 2017 & 1.1 & Dropout ratio increases with training epochs & FCL & ICCV \\
Targeted Dropout\cite{gomez2018targeted, gomez2019learning} & 2018 & 1.1 & Dropout for neural pruning & FCL & NeurIPS \\
Ising-Dropout\cite{salehinejad2019ising} & 2019 & 1.1 & Incorporate Ising model & FCL & ICASSP \\
EDropout\cite{salehinejad2021edropout} & 2021 & 1.1 & Use EBM to decide pruning state & FCL & TNNLS \\
LocalDrop\cite{lu2021localdrop} & 2021 & 1.1 & Based on local Rademacher complexity & FCL & TPAMI \\
SimCSE\cite{gao2021simcse} & 2021 & 1.1 & Data augmentation by dropout twice & FCL & EMNLP \\
Child-Tuning\cite{xu2021raise} & 2021 & 1.1 & Mask gradient when back-propagation & FCL & EMNLP \\
R-Drop\cite{liang2021r} & 2021 & 1.1 & Dropout twice to regularize & FCL & arxiv \\
AS-Dropout\cite{chen2021adaptive} & 2021 & 1.1 & Adaptive sparse dropout & FCL & Neurocomput. \\
SpatialDropout\cite{tompson2015efficient} & 2015 & 1.2.1\dag & Drop feature maps in CNN & CNN & CVPR \\
Max-pooling Dropout\cite{wu2015towards} & 2015 & 1.2.1 & Drop neurons before pooling layer & CNN & NN \\
Convolutional Dropout\cite{wu2015towards} & 2015 & 1.2.1 & Drop neurons before convolutional layer & CNN & NN \\
Max-drop\cite{park2016analysis} & 2016 & 1.2.1 & Drop feature maps with high activations & CNN & ACCV \\
Stochastic Dropout\cite{park2016analysis} & 2016 & 1.2.1 & Dropout ratio sampled from normal distribution & CNN & ACCV     \\
DropBlock\cite{ghiasi2018dropblock} & 2018 & 1.2.1 & Drop contiguous regions on each feature map & CNN & NeurIPS  \\
Spectral Dropout\cite{khan2018regularization} & 2018 & 1.2.1 & Dropout in the frequency domain & CNN & NN \\
Drop-Conv2d\cite{cai2019effective} & 2019 & 1.2.1 & Dropout before convolution instead of BN & CNN & arxiv \\
Weighted Channel Dropout\cite{hou2019weighted} & 2019 & 1.2.1 & Drop weighted feature channels & CNN    & AAAI     \\
CorrDrop\cite{zeng2021correlation} & 2021 & 1.2.1 & Drop neurons according to feature correlation & CNN & PR \\
LocalDrop\cite{lu2021localdrop} & 2021 & 1.2.1 & Based on local Rademacher complexity & CNN & TPAMI \\
AutoDropout\cite{pham2021autodropout} & 2021 & 1.2.1 & Optimize dropout patterns by RL & CNN    & AAAI     \\
Vanilla drop for RNN\cite{pham2014dropout, zaremba2014recurrent} & 2014 & 1.2.2\dag & Drop feed-forward connections only & RNN    & ICFHR \\
RNNDrop\cite{moon2015rnndrop} & 2015 & 1.2.2 & One dropping mask for each layer & RNN & ASRU     \\
Variational RNN Dropout\cite{gal2016theoretically}  & 2015 & 1.2.2 & Variational inference based dropout & RNN & NeurIPS  \\
Recurrent Dropout\cite{semeniuta2016recurrent} & 2016 & 1.2.2 & Drop only the vectors generating hidden states & RNN    & COLING   \\
Zoneout\cite{krueger2016zoneout} & 2016 & 1.2.2 & Residual connections between timestamps & RNN & ICLR \\
Weighted-dropped LSTM\cite{merity2018regularizing} & 2017 & 1.2.2 & Drop weights like DropConnect & RNN    & ICLR     \\
\rule{-2.3pt}{12pt}
Fraternal Dropout\cite{zolna2018fraternal} & 2018 & 1.2.2 & \makecell[l]{Train two identical RNNs with different\\dropout masks} & RNN & ICLR \\
\rule{-2.3pt}{7pt}
Stochastic Depth\cite{huang2016deep} & 2016 & 1.2.3\dag & Drop blocks and retain only residual connections & ResNet & ECCV     \\
Shakeout\cite{kang2016shakeout} & 2016 & 1.2.3 & Assign new weights to neurons & ResNet & AAAI     \\
Whiteout\cite{li2016whiteout} & 2016 & 1.2.3 & Introduce Gaussian noise compared to Shakeout & ResNet & arxiv \\
\rule{-2.3pt}{12pt}
Swapout\cite{singh2016swapout} & 2016 & 1.2.3 & \makecell[l]{A synthesis of standard Dropout and Stochastic\\Depth} & ResNet & NeurIPS  \\
\rule{-2.3pt}{7pt}
DropPath\cite{larsson2016fractalnet} & 2016 & 1.2.3 & Drop subpaths in Fractalnet & DNN & ICLR     \\
Shake-Shake\cite{gastaldi2017shake} & 2017 & 1.2.3 & Assign weights in 3-way ResNet & ResNet & arxiv \\
ShakeDrop\cite{yamada2019shakedrop} & 2018 & 1.2.3 & Improve Shake-Shake to other form of ResNet & ResNet & IEEE Access \\
Scheduled DropPath\cite{zoph2018learning} & 2018 & 1.2.3 & Dropout ratio increases linearly & DNN & CVPR     \\
DropHead\cite{zhou2020scheduled} & 2020 & 1.2.3 & Drop attention heads of Transformer & Transformer & EMNLP     \\
        \bottomrule
    \end{tabular}
\begin{tablenotes}
\footnotesize
\item[\dag] 1.1 refers to dropping individual neurons, 1.2.1 dropping 2D neuron groups, 1.2.2 dropping recurrent connections, and 1.2.3 dropping residual connections or others. 
\item[*] FCL refers to Fully Connected Layers.
\end{tablenotes}
\end{threeparttable}
    \label{tab:drop neuron groups}
    \vspace{-12pt}
\end{table*}

Wan et al. \cite{wan2013regularization} proposed DropConnect in 2013. Compared to standard dropout which randomly zeroes the output of neurons, DropConnect randomly zeroes elements of the weight matrix of each layer:
\begin{equation} \label{eq:dropconnect 1}
    \begin{aligned}
    \mathbf{y} = f((\mathbf{W}\circ \mathbf{M})\mathbf{x}), \quad m_{ij}\sim \mathrm{Bernoulli}(1-p)
    \end{aligned}
    \vspace{-2pt}
\end{equation}
This approach removes ``connections'' in fully connected layers, hence the name ``DropConnect''.
% 在英文论文里引号不是用""，应该是``DropConnect'' （你复制到latex就会知道，这样是有方向的双引号）

Goodfellow et al. \cite{goodfellow2013maxout} proposed Maxout in 2013. Maxout is an improvement of standard dropout \cite{hinton2012improving}. Specifically, the output of each hidden layer is computed as
\begin{equation} \label{eq:maxout 1}
    \begin{aligned}
    h_i(\mathbf{x}) = \max_{j\in [1, k]}{z_{ij}} 
    ,\ \mathrm{where}\ z_{ij} = \mathbf{x}^T \mathbf{W}_{:ij} + b_{ij}
    \end{aligned}
    \vspace{-2pt}
\end{equation}
where the weight matrix $\mathbf{W}\in \mathbb{R}^{d\times m\times k}$ and the bias vector $\mathbf{b}\in \mathbb{R}^{m\times k}$ are training parameters. $\mathbf{x}$, $d$, $m$, and $k$ are the input, the input dimension, the output dimension, and Maxout parameter, respectively. As can be seen, unlike in standard dropout \cite{hinton2012improving} where only one output is computed for each input at each layer, Maxout computes $k$ outputs for each input at each layer. Then it takes the maximum of the $k$ outputs as the output of this layer. This operation makes Maxout essentially a nonlinear activation function, which gives the method its name. Maxout has been applied in computer vision tasks including object detection \cite{zhou2017cad}.

Kingma et al. \cite{kingma2015variational} proposed Variational Dropout in 2015. This work studies Stochastic Gradient Variational Bayesian Inference (SGVB) problem and found its connection with dropout: Gaussian Dropout proposed in \cite{srivastava2014dropout} is a local reparameterization of SGVB. This paper thus proposes Variational Dropout so that the dropout ratio $p$ is not a pre-set hyperparameter that requires human adjusting but a parameter that can be learned through training. In \cite{molchanov2017variational} the authors show that Variational Dropout is a efficient way to perform model compression, which can significantly reduce the number of parameters of neural networks with a negligible decrease of accuracy.

Gal and Ghahramani proposed Monte Carlo Dropout \cite{gal2016dropout} in 2016. The authors interpret dropout as a Bayesian approximation of deep Gaussian processes. The output of a deep Gaussian process is a probability distribution, and using standard dropout in testing phase can estimate some properties of this potential distribution. For example, the estimated variance can be used to characterize the uncertainty of the model output, and this estimating method is called Monte Carlo Dropout. Monte Carlo Dropout has been applied in a series of works \cite{gal2016uncertainty, zhu2017deep, jungo2017towards, lakshminarayanan2017simple}. It can mitigate the problem of representing uncertainty in deep learning more efficiently without sacrificing test accuracy.

Smith et al. \cite{smith2016gradual} proposed DropIn in 2016. The feed-forward operation for each layer performing DropIn is:
\begin{equation} \label{eq:dropin 1}
    \begin{aligned}
    \Tilde{\mathbf{y}}^{(l)} &= \mathbf{r}^{(l)}\circ \mathbf{y}^{(l)},\quad
    \mathbf{z}^{(l+1)} = \mathbf{W}^{(l+1)} \Tilde{\mathbf{y}}^{(l)} + \mathbf{b}^{(l+1)} \\
    \mathbf{y}^{(l+1)} &= f(\mathbf{z}^{(l+1)}) + (1 - \mathbf{r}^{(l)})\circ \mathbf{y}^{(l)}
    \end{aligned}
    \vspace{-2pt}
\end{equation}
As can be seen, in addition to passing the kept outputs ($\mathbf{r}^{(l)}\circ \mathbf{y}^{(l)}$) to the next layer, DropIn also passes the values of dropped positions ($(1 - \mathbf{r}^{(l)})\circ \mathbf{y}^{(l)}$) directly to the next layer without going through the nonlinear activation function. This operation increases the depth of the network while avoiding vanishing gradient problem while serving the same regularization effect as standard dropout.

Li et al. \cite{li2016improved} proposed Evolutional Dropout in 2016. Intuitively, the importance of neurons corresponding to different features in a neural network is different, so their corresponding dropout probabilities should be different. This paper applies this idea to both shallow and deep neural networks: for shallow networks, the dropout ratio is calculated from the second-order statistics of the input data features; for deep networks, the dropout ratio of each layer is calculated in real-time from the output of that layer of each batch. Compared with the standard dropout, Evolutional Dropout improves the accuracy of the results while greatly increasing the convergence speed.

Gal et al. \cite{gal2017concrete} proposed Concrete Dropout in 2017. Concrete Dropout is an improvement on Monte Carlo Dropout \cite{gal2016dropout}. Monte Carlo Dropout can estimate the model uncertainty, achieved by performing a grid search on the dropout ratio parameter. This is unfeasible for deeper models (e.g., those in computer vision tasks) and reinforcement learning models because of the excessive computational time and resources. Based on the development of Bayesian learning, this paper uses a continuous relaxation of the dropout discrete mask. Concrete Dropout proposes a new objective function which allows automatic adjustment of the dropout parameters on large models, reducing the time required for experiments. It also allows the agent in reinforcement learning to dynamically adjust its uncertainty as the training process goes on and more training data is exposed. Experiments show that Concrete Dropout can decrease the time of model training by weeks by automatically learning the dropout probabilities in reinforcement learning compared to conventional dropout methods.

Rennie et al. \cite{rennie2014annealed} proposed Annealed Dropout in 2014. As the name implies, ``annealed'' dropout is the decrease of dropout ratio as the number of training epochs increases:
\begin{equation} \label{eq:annealed dropout 1}
    \vspace{-2pt}
    \begin{aligned}
    p[t] = p[t-1] + \alpha_t(\theta)
    \end{aligned}
    \vspace{-2pt}
\end{equation}
$\alpha_t(\theta)$ is the parameter that controls the dropout ratio. A simple approach is to decrease linearly: the initial dropout ratio is $p[0]$ and decreases to $0$ after $N$ rounds:
\begin{equation} \label{eq:annealed dropout 2}
    \vspace{-2pt}
    \begin{aligned}
    p[t] = \max(0, 1 - \dfrac{t}{N}) p[0]
    \end{aligned}
    \vspace{-2pt}
\end{equation}
The explanation for this approach is that at the beginning when we are exposed to little training data, we only need to ``explain'' the data with a simple model, i.e., we only make fewer neurons work, and more neurons are dropped out. Later on, when more training data is exposed, we can allow a more ``complex'' model to ``explain'' the data, reducing the dropout ratio and making more neurons work.

Morerio et al. \cite{morerio2017curriculum} proposed Curriculum Dropout in 2017. Inspired by curriculum learning \cite{bengio2009curriculum}, in contrast to Annealed Dropout \cite{rennie2014annealed}, Curriculum Dropout increases dropout ratio as the number of training epochs increases. The explanation for this approach is to simulate the learning process from easy to difficult in human learning: the dropout ratio is small at the beginning, introducing less noise and analogous to the ``easy'' task; then the dropout ratio increases, introducing more noise and making the task ``harder''.

Gomez et al. \cite{gomez2018targeted, gomez2019learning} proposed Targeted Dropout in 2018. Neural pruning is a neural network compression method \cite{han2015deep} that can be used to reduce the number of network parameters and improve training efficiency. The Targeted Dropout selects and drops those neurons whose absence can make the model most suitable for neural network pruning, facilitating model compression. Targeted Dropout is able to achieve better performance with only half of total number of parameters compared to the original networks without dropout.

Salehinejad and Valaee \cite{salehinejad2019ising} proposed Ising-Dropout in 2019. Borrowing the concept of Ising model in physics, Ising-Dropout adds an image Ising model to a neural network to detect and drop out those neurons that are least useful. It could compress the number of parameters up to 41.18\% and 55.86\% for the classification task on the MNIST and Fashion-MNIST datasets respectively. The authors also proposed EDropout \cite{salehinejad2021edropout} working for neural pruning in 2021. It utilizes an Energy-Based Model (EBM) to decide the pruning state. 

In 2020, {\.I}rsoy and Alpayd{\i}n \cite{irsoy2021dropout} proposed a dropout method for hierarchically gated models \cite{titsias2002mixture, qi2016hierarchically} to prevent overfitting in decision-tree-like models. Ragusa et al. \cite{ragusa2021random} employ dropout on Internet of Things (IoT) models.

Gao et al. \cite{gao2021simcse} proposed SimCSE in 2021. SimCSE is a contrastive learning method for NLP. Specifically, it performs dropout twice for the same instance to get two positive samples and treats all other in-batch instances as negative. Performing data augmentation by dropout twice achieves good results on this contrastive learning task. Child-Tuning \cite{xu2021raise} is another application of dropout in NLP. It randomly masks gradient when back-propagation.

Liang et al. \cite{liang2021r} proposed R-Drop (``R'' for ``Regularized'') in 2021. R-Drop generalized the idea of ``dropout twice'' \cite{gao2021simcse} from contrastive learning to general tasks. A training instance goes through the network twice with random dropout. On the one hand, we want the two predictions as close as possible to the label, by which we compute cross-entropy loss $\mathcal{L}^{(CE)}$, on the other hand, we want the two predictions as close as possible to each other, by which we compute Kullback-Leibler divergence loss $\mathcal{L}^{(KL)}$. Thus we get final loss function consists of two parts:
\begin{equation} \label{eq:r-drop 1}
    \begin{aligned}
    \mathcal{L} = \mathcal{L}^{(CE)} + \alpha \mathcal{L}^{(KL)}
    \end{aligned}
    \vspace{-4pt}
\end{equation}
and the second part is ``regularizing'' the first part.

Chen and Yi \cite{chen2021adaptive} proposed AS-Dropout (Adaptive Sparse Dropout) in 2021. AS-Dropout calculates dropout probability adaptively according to the neuron's activation function, such that only a small proportion of neurons are active in each training epoch. 

Dropout methods that drop individual neurons are summarized in Table \ref{tab:drop neuron groups}. The application scenario for the dropout of individual neurons is usually fully connected layers. The basic operation of this type of methods is easy to implement and can be applied to a wide range of neural models. It is likely to lead to a stable improvement on the model performance in most cases, as will be shown in Section \ref{sec:experiments}. The generality and effectiveness has made it the most popular type of dropout methods. However, it is not suitable for models with specific structure (e.g. CNN, RNN, Transformer), for which the dropout methods usually drop neuron groups.

\subsubsection{Drop Neuron Groups}\label{subsubsec:drop neuron groups}

Dropout methods in \ref{subsubsec:drop single neurons} are mainly performed on fully connected layers. For neural networks with special structures, such as convolutional neural networks, residual networks, and recurrent neural networks, neurons are aggregated together forming specific structures to perform certain functions. Directly dropping individual neurons randomly may not have expected effect on these networks, so a series of dropout methods have been proposed specified for them.

\paragraph{\textbf{CNNs}} \label{para:drop cnn}

Tompson et al. \cite{tompson2015efficient} first proposed SpatialDropout specifically for convolutional neural networks in 2014. In CNNs, for the same feature map, all pixel features within the coverage of the same convolutional kernel are used to compute the output of the next layer, resulting a strong gradient correlation between adjacent pixels. The standard dropout removes individual pixel features randomly, which has little effect on reducing the interdependence between neurons, and is therefore ineffective. In contrast, for a feature tensor with size $n_{\mathrm{feats}}\times \mathrm{height}\times \mathrm{ width}$, SpatialDropout selects only $n_{\mathrm{feats}}$ dropout values, i.e., the whole feature map is either dropped for all or kept for all. This reduces the interdependence between neurons in CNN and has a good regularization effect.

Wu and Gu \cite{wu2015towards} proposed Max-pooling Dropout in 2015. For a neural network containing convolutional layers and pooling layers, if the $l$th layer is immediately followed by a pooling layer, the feed-forward operation is expressed as
\begin{equation} \label{eq:max-pooling dropout 1}
    \begin{aligned}
    a_j^{(l+1)} = pool(a_1^{(l)}, \dots, a_i^{(l)}, \dots, a_n^{(l)}), \quad i\in R_j^{(l)}
    \end{aligned}
    \vspace{-3pt}
\end{equation}
where $R_j^{(l)}$ is the $j$th pooling region of the $l$th layer, $a_i^{(l)}$ is the activation value of each neuron, and $pool()$ is the pooling function. Two common choices are average pooling, which averages the outputs of all neurons, and max-pooling, which takes the maximum of all neuron outputs. The authors take the latter one. Max-pooling Dropout randomly drops out the output $a_i^{(l)}$ of individual neurons during training phase, which is then passed into the pooling layer:
\begin{equation} \label{eq:max-pooling dropout 2}
    \begin{aligned}
    \hat{a}_i^{(l)} &= m_i^{(l)} * a_i^{(l)}, \quad m_i^{(l)} \sim \mathrm{Bernoulli}(p) \\
    a_j^{(l+1)} &= pool(\hat{a}_1^{(l)}, \dots, \hat{a}_i^{(l)}, \dots, \hat{a}_n^{(l)}), \quad i\in R_j^{(l)}
    \end{aligned}
    \vspace{-2pt}
\end{equation}
\begin{figure}
  \centering
  \includegraphics[width=0.55\linewidth]{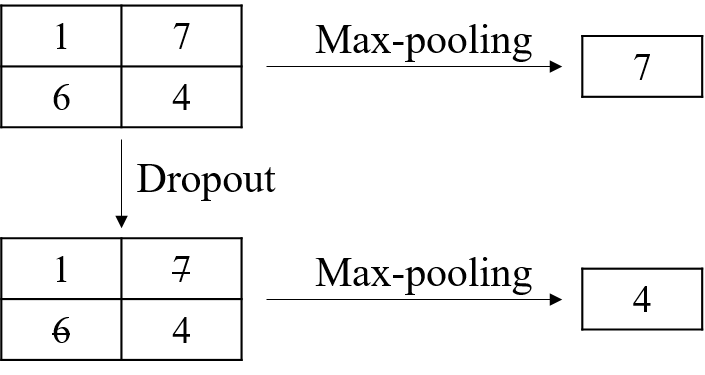}
  \vspace{-6pt}
  \caption{Max-pooling Dropout \cite{wu2015towards} drops neuron outputs before they are passed to pooling layer.}
  \label{fig:max-pooling}
  \vspace{-10pt}
\end{figure}
as shown in Figure \ref{fig:max-pooling}. This is therefore equivalent to selecting neuron outputs from a multinomial distribution:
\begin{equation} \label{eq:max-pooling dropout 3}
    \begin{aligned}
    \mathrm{Pr}(a_j^{(l+1)} = a^{\prime (l)}_i) = p_i = pq^{n-i}, \quad i = 1, 2, \dots, n
    \end{aligned}
    \vspace{-2pt}
\end{equation}
% No dropout is performed during testing phase so correction to the neuron output is required:
% \begin{equation} \label{eq:max-pooling dropout 4}
%     \begin{aligned}
%     a_j^{(l+1)} = \sum_{i=0}^n p_i a^{\prime (l)}_i = \sum_{i=1}^n p_i a^{\prime (l)}_i
%     \end{aligned}
% \end{equation}
If the $l$th layer is immediately followed by a convolutional layer, this paper also proposes Convolutional Dropout. The feed-forward operation in the training phase is
\begin{equation} \label{eq:max-pooling dropout 5}
    \begin{aligned}
    m_k^{(l)}(i) &\sim \mathrm{Bernoulli}(p),\quad
    \hat{a}_k^{(l)} = a_k^{(l)} \ast m_k^{(l)}, \\
    z_j^{(l+1)} &= \sum_{k=1}^{n^{(l)}} \mathrm{conv}(W_j^{(l+1)}, \hat{a}_k^{(l)}),\\
    a_j^{(l+1)} &= f(z_j^{(l+1)}).
    \end{aligned}
    \vspace{-4pt}
\end{equation}
where $a_k^{(l)}$ is the $k$th feature map of the $l$th layer. No dropout is performed during testing and all neuron outputs need to be multiplied by the retain probability of the training phase.

In 2016, Park and Kwak \cite{park2016analysis} makes two improvements to SpatialDropout \cite{tompson2015efficient}. The first is to select and drop out high activation values on the feature maps or channels; the second is that the dropout ratio is not a fixed value but is obtained by sampling from a normal distribution. The authors refer to these two improved dropout methods as Max-drop and Stochastic Dropout, respectively.

Ghiasi et al. \cite{ghiasi2018dropblock} proposed DropBlock in 2018. For each layer of the feature map, DropBlock randomly drops multiple contiguous regions of size $block\_size\times block\_size$. When $block\_size = 1$, DropBlock is reduced to standard dropout; when $block\_size = feature\_map\_size$, i.e., one block can cover the whole layer of feature map, DropBlock is equivalent to SpatialDropout \cite{tompson2015efficient}.

Khan et al. \cite{khan2018regularization} proposed Spectral Dropout in 2018. A Spectral Dropout operation is added between two layers of a CNN. The operation has three steps: transforming activation values of the previous layer to frequency domain; dropping the components below a certain threshold in frequency domain; and changing back to the original value domain. The feed-forward operation between two layers without Spectral Dropout is expressed in the following equation:
\begin{equation} \label{eq:spectral dropout 1}
    \begin{aligned}
    \mathbf{A}_l^{\prime} = f(\mathbf{F}_l\otimes \mathbf{A}_{l-1} + \mathbf{b}_l)
    \end{aligned}
    \vspace{-2pt}
\end{equation}
When performing Spectral Dropout, the operation becomes
\begin{equation} \label{eq:spectral dropout 2}
    \begin{aligned}
    \mathbf{A}_l = \mathcal{T}^{-1}( \mathbf{M}\circ \mathcal{T}( f(\mathbf{F}_l\otimes \mathbf{A}_{l-1} + \mathbf{b}_l) ) )
    \end{aligned}
    \vspace{-2pt}
\end{equation}
where $\mathcal{T}$ denotes the frequency transformation and $\mathbf{M}$ is the masking matrix of dropout in frequency domain. This approach serves to filter input noises and effectively speeds up the convergence of network training.

In 2019, Cai et al. \cite{cai2019effective} analyze why standard dropout does not work in convolutional neural networks: it conflicts with the effect of Batch Normalization \cite{ioffe2015batch}. It is experimentally verified that putting dropout operation before convolution operation instead of batch normalization operation can effectively improve the dropout effect. Then the authors proposed Drop-Conv2d method to improve the training effect by combining dropout at feature-channel level and dropout at forward-path level in CNN.

Hou and Wang \cite{hou2019weighted} proposed Weighted Channel Dropout for feature channel dropout in 2019. The operation steps are in two stages: scoring feature channels and selecting feature channels. In the first stage, feature channels are scored using Global Average Pooling (GAP) method:
\begin{equation} 
\label{eq:weighted channel dropout 1}
    \begin{aligned}
    score_i = \dfrac{1}{W\times H} \sum_{j=1}^W \sum_{k=1}^H x_i(j, k)
    \end{aligned}
    \vspace{-4pt}
\end{equation}
In the second stage, the weighted random selection (WRS) and random number generation (RNG) steps are used to select feature channels for dropout and retention.

Zeng et al. \cite{zeng2021correlation} proposed CorrDrop in 2021. CorrDrop drops out CNN neurons based on their feature correlation and can do it in both spatial manner and channel manner.

Lu et al. \cite{lu2021localdrop} proposed LocalDrop in 2021. This regularization method is based on theoretical analysis of local Rademacher complexity and can be applied to both fully connected layers and convolutional layers. 

Pham and Le \cite{pham2021autodropout} proposed AutoDropout in 2021. AutoDropout uses reinforcement learning to train a controller selecting the optimal dropout pattern to train the model. The controller eliminates the need of manually adjusting dropout patterns as in previous methods such as DropBlock.

\paragraph{\textbf{RNNs}}\label{para:drop rnn}

Pachitariu et al. \cite{pachitariu2013regularization} applied standard dropout directly to RNNs in 2013 to randomly drop the outputs of neurons in RNNs. Bayer et al. \cite{bayer2013fast}, in the same year, applied Fast Dropout \cite{wang2013fast} directly to RNN.

\begin{figure}
  \centering
  \includegraphics[width=0.37\linewidth]{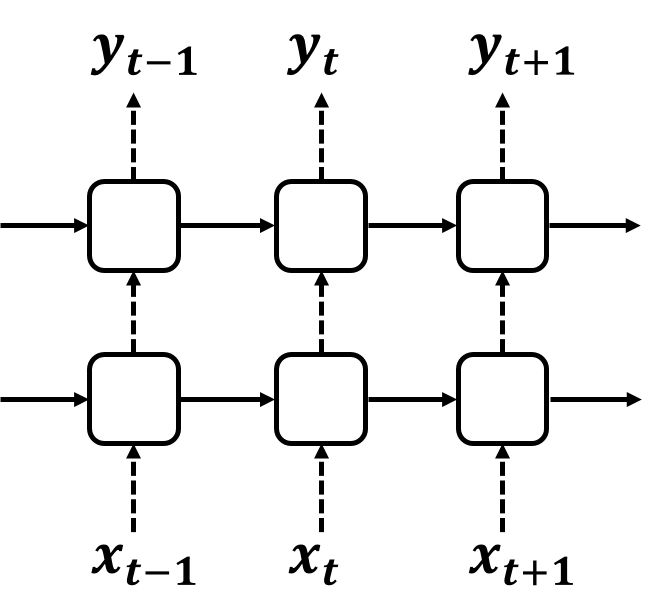}
  \vspace{-6pt}
  \caption{Vanilla dropout for RNNs, which only drops feed-forward connections but not recurrent connections. \cite{zaremba2014recurrent}}
  \label{fig:rnn_regularization}
  \vspace{-10pt}
\end{figure}

Pham et al. \cite{pham2014dropout} first proposed a dropout method specific to RNN structure in 2014, rather than just applying random dropout directly to RNNs. Instead of dropping connections between hidden states at different timestamps (recurrent connections), only the connections from input to output direction (feed-forward connections) are dropped. The dropout is performed in the same way as the standard dropout, i.e., training with a mask $\mathbf{m}$ of Bernoulli distribution and an elementary multiplication of the hidden state vector at each layer, and testing with all neurons working thus multiplying the output by the retention probability $p$:
\begin{equation} \label{eq:rnn dropout 1}
    \begin{aligned}
    \mathbf{h}_{\mathrm{train}} = \mathbf{m}\odot\mathbf{h},\quad
    \mathbf{h}_{\mathrm{test}} = p\mathbf{h}
    \end{aligned}
    \vspace{-2pt}
\end{equation}

Zaremba et al. \cite{zaremba2014recurrent} also elaborated this method more systematically in 2014. For multilayer LSTMs \cite{hochreiter1997long}, only the connections between layers are dropped, not for connections at different timestamps within the same layer. Let $h_t^l$ be the hidden state at moment $t$ of the $l$th layer, $c_t^l$ be the memory unit at moment $t$ of the $l$th layer, and $T_{n, m}: \mathbb{R}^n \rightarrow\mathbb{R}^m$ be an affine transformation, then performing dropout only to the connections between layers is:
% $\mathrm{LSTM}: h_t^{l-1} , h_{t-1}^l, c_{t-1}^l \rightarrow h_t^l, c_t^l$ is:
% \begin{equation} \label{eq:rnn dropout 2}
%     \begin{aligned}
%     \begin{pmatrix}
%     i\\f\\o\\g
%     \end{pmatrix} &= \begin{pmatrix}
%     \mathrm{sigm}\\\mathrm{sigm}\\\mathrm{sigm}\\\tanh
%     \end{pmatrix} T_{2n,4n} \begin{pmatrix}
%     h_t^{l-1}\\h_{t-1}^l
%     \end{pmatrix} \\
%     c_t^l &= f\odot c_{t-1}^l + i\odot g \\
%     h_t^l &= o\odot \tanh(c_t^l)
%     \end{aligned}
% \end{equation}
% Perform dropout only to the connections between layers:
\begin{equation} \label{eq:rnn dropout 3}
    \begin{aligned}
    \begin{pmatrix}
    i\\f\\o\\g
    \end{pmatrix} &= \begin{pmatrix}
    \mathrm{sigm}\\\mathrm{sigm}\\\mathrm{sigm}\\\tanh
    \end{pmatrix} T_{2n,4n} \begin{pmatrix}
    \mathbf{D}(h_t^{l-1})\\h_{t-1}^l
    \end{pmatrix}, \\
    c_t^l &= f\odot c_{t-1}^l + i\odot g,\quad
    h_t^l = o\odot \tanh(c_t^l)
    \end{aligned}
    \vspace{-2pt}
\end{equation}
where $\mathbf{D}$ is dropout operation matrix, which acts only on the output $h_t^{l-1}$ of previous layer at the same timestamp, and not on the output $h_{t-1}^l$ at the previous timestamp of this layer, as shown in Figure \ref{fig:rnn_regularization}\cite{zaremba2014recurrent}.
The dashed lines in Figure \ref{fig:rnn_regularization} indicate the connections to which dropout is applied and may be dropped, while the solid lines indicate the connections retained.

The motivation of both the above articles \cite{pham2014dropout, zaremba2014recurrent} is that the strength of RNN is its memory capacity, but if recurrent connections are dropped, the memory capacity of RNN will be impaired. To test this idea, the authors of \cite{pham2014dropout} did a series of experiments in \cite{bluche2015apply} 2015 to discuss dropout for RNNs and where in the network dropout operation should be performed. The authors examined the effect of adding dropout layers before LSTM input, on LSTM recurrent connection direction, and after LSTM output, respectively. They found that in most cases, adding dropout layers on LSTM input and output directions is better than adding them on recurrent connection direction, which experimentally verifies the ideas in the previous two papers.\

\begin{figure}
  \centering
  \includegraphics[width=0.8\linewidth]{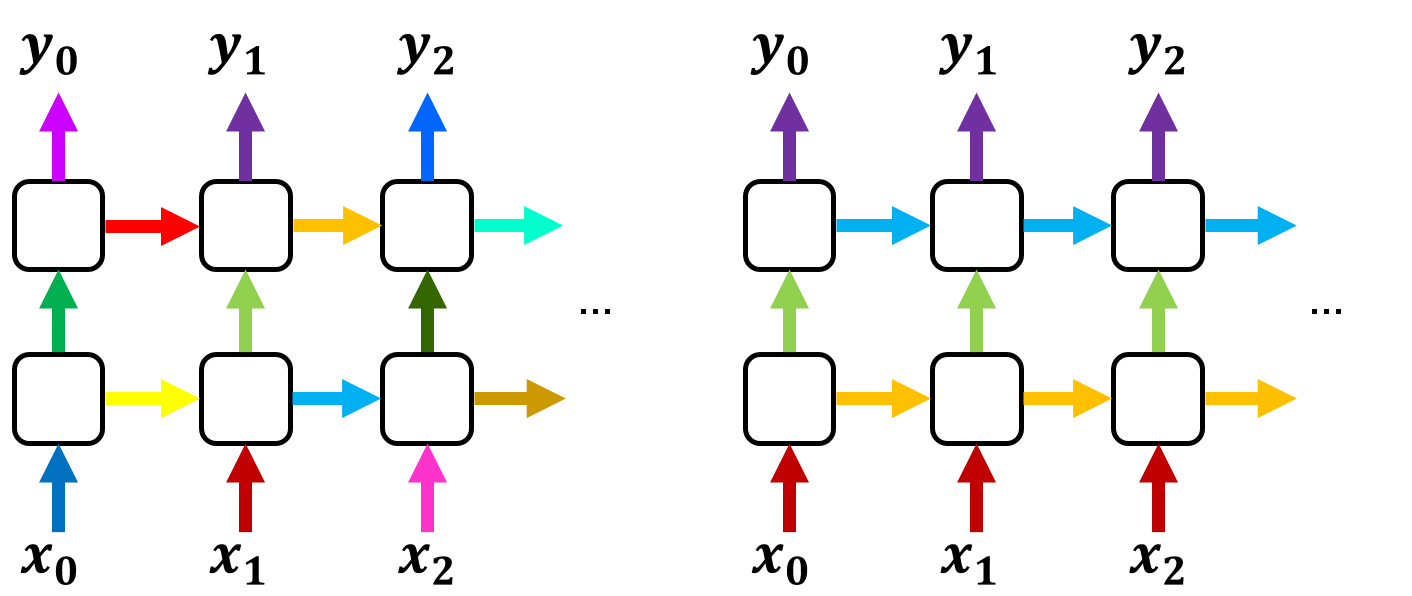}
  \vspace{-6pt}
  \caption{RNNDrop (right) outperforms standard dropout (left) by generating one mask for each layer in RNN and keeping it throughout the sequence. \cite{moon2015rnndrop, labach2019survey}}
  \label{fig:rnndrop}
  \vspace{-10pt}
\end{figure}

Moon et al. \cite{moon2015rnndrop} proposed RNNDrop in 2015, which gives a method for dropping recurrent connections between different timestamps of the same layer. This is done by generating only one dropping mask for each layer and then using this one mask at all timestamps of that layer. In this way, elements that are dropped at the first timestamp will not be used at subsequent timestamps, and elements that are kept at the first timestamp will be passed through to the last timestamp. The schematic is shown in Figure \ref{fig:rnndrop}\cite{labach2019survey}. The left side of the figure shows the dropout of RNNs using random dropout method, and the right shows how RNNDrop working, with the same colors indicating the same dropout masks. In this way, the model is regularized with dropout while retaining RNN memory capability.

Gal and Ghahramani \cite{gal2016theoretically} proposed Variational RNN Dropout in 2015. The authors view dropout as an approximate inference process in Bayesian neural networks, which can also perform dropout of both feed-forward connections and recurrent connections. In this regard, Variational RNN Dropout can be seen as a variant of RNNDrop \cite{moon2015rnndrop}.

\begin{figure}
  \centering
  \includegraphics[width=\linewidth]{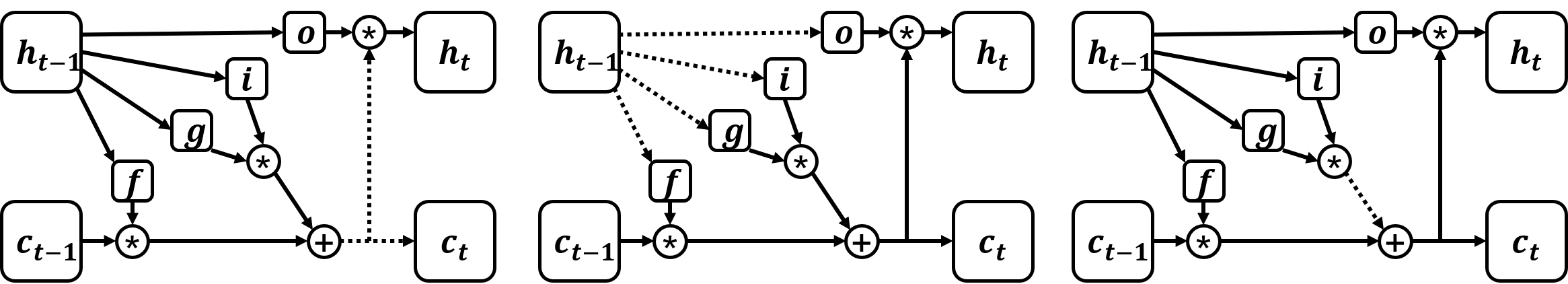}
  \vspace{-12pt}
  \caption{Comparison of RNNRrop, Variational RNN Dropout and Recurrent Dropout \cite{semeniuta2016recurrent}.}
  \label{fig:recurrent_dropout}
  \vspace{-7pt}
\end{figure}

Recurrent Dropout proposed by Semeniuta et al. \cite{semeniuta2016recurrent} in 2016 is another dropout method that preserves memory capacity of RNNs and generates random dropout masks at each step, just like standard dropout does \cite{hinton2012improving, srivastava2014dropout}. This is done by dropping only the vectors used to generate hidden state vectors, but not dropping hidden state vectors themselves. Recurrent Dropout is specialized for gated RNNs, such as LSTM\cite{hochreiter1997long} and GRU\cite{chung2014empirical}. For LSTM in Eq. \ref{eq:recurrent dropout 1},
\begin{equation} \label{eq:recurrent dropout 1}
    \begin{aligned}
    \begin{pmatrix}
    \mathbf{i}_t\\\mathbf{f}_t\\\mathbf{o}_t\\\mathbf{g}_t
    \end{pmatrix} &= \begin{pmatrix}
    \sigma(\mathbf{W}_i[\mathbf{x}_t,\mathbf{h}_{t-1}] + \mathbf{b}_i)\\
    \sigma(\mathbf{W}_f[\mathbf{x}_t,\mathbf{h}_{t-1}] + \mathbf{b}_f)\\
    \sigma(\mathbf{W}_o[\mathbf{x}_t,\mathbf{h}_{t-1}] + \mathbf{b}_o)\\
    f(\mathbf{W}_g[\mathbf{x}_t,\mathbf{h}_{t-1}] + \mathbf{b}_g)
    \end{pmatrix}, \\
    \mathbf{c}_t = &\mathbf{f}_t\ast \mathbf{c}_{t-1} + \mathbf{i}_t\ast \mathbf{g}_t,\quad
    \mathbf{h}_t = \mathbf{o}_t\ast f(\mathbf{c}_t)
    \end{aligned}
    \vspace{-2pt}
\end{equation}
Variational RNN Dropout\cite{gal2016theoretically} performs dropout as Equation \ref{eq:recurrent dropout 2},
\begin{equation} \label{eq:recurrent dropout 2}
    \begin{aligned}
    \begin{pmatrix}
    \mathbf{i}_t\\\mathbf{f}_t\\\mathbf{o}_t\\\mathbf{g}_t
    \end{pmatrix} &= \begin{pmatrix}
    \sigma(\mathbf{W}_i[\mathbf{x}_t,d(\mathbf{h}_{t-1})] + \mathbf{b}_i)\\
    \sigma(\mathbf{W}_f[\mathbf{x}_t,d(\mathbf{h}_{t-1})] + \mathbf{b}_f)\\
    \sigma(\mathbf{W}_o[\mathbf{x}_t,d(\mathbf{h}_{t-1})] + \mathbf{b}_o)\\
    f(\mathbf{W}_g[\mathbf{x}_t,d(\mathbf{h}_{t-1})] + \mathbf{b}_g)
    \end{pmatrix}
    \end{aligned}
    \vspace{-2pt}
\end{equation}
RNNDrop\cite{moon2015rnndrop} performs dropout as Equation \ref{eq:recurrent dropout 3}.
\begin{equation} \label{eq:recurrent dropout 3}
    \begin{aligned}
    \mathbf{c}_t = d(\mathbf{f}_t\ast \mathbf{c}_{t-1} + \mathbf{i}_t\ast \mathbf{g}_t)
    \end{aligned}
    \vspace{-2pt}
\end{equation}
This method, on the other hand, performs dropout as Equation \ref{eq:recurrent dropout 4}.
\begin{equation} \label{eq:recurrent dropout 4}
    \begin{aligned}
    \mathbf{c}_t = \mathbf{f}_t\ast \mathbf{c}_{t-1} + \mathbf{i}_t\ast d(\mathbf{g}_t)
    \end{aligned}
    \vspace{-2pt}
\end{equation}
% For GRU expressed as Equation \ref{eq:recurrent dropout 5},
% \begin{equation} \label{eq:recurrent dropout 5}
%     \begin{aligned}
%     \begin{pmatrix}
%     \mathbf{z}_t\\\mathbf{r}_t
%     \end{pmatrix} &= \begin{pmatrix}
%     \sigma(\mathbf{W}_z[\mathbf{x}_t,\mathbf{h}_{t-1}] + \mathbf{b}_z) \\
%     \sigma(\mathbf{W}_r[\mathbf{x}_t,\mathbf{h}_{t-1}] + \mathbf{b}_r)
%     \end{pmatrix} \\
%     \mathbf{g}_t &= f(\mathbf{W}_g[\mathbf{x}_t,\mathbf{r}_t\ast\mathbf{h}_{t-1}] + \mathbf{b}_g) \\
%     \mathbf{h}_t &= (1 - \mathbf{z}_t)\ast \mathbf{h}_{t-1} + \mathbf{z}_t\ast\mathbf{g}_t
%     \end{aligned}
% \end{equation}
% Recurrent Dropout performs dropout as Equation \ref{eq:recurrent dropout 6}.
% \begin{equation} \label{eq:recurrent dropout 6}
%     \begin{aligned}
%     \mathbf{h}_t = (1 - \mathbf{z}_t)\ast \mathbf{h}_{t-1} + \mathbf{z}_t\ast d(\mathbf{g}_t)
%     \end{aligned}
% \end{equation}
Dropout operations performed by RNNDrop\cite{moon2015rnndrop}, Variational RNN Dropout\cite{gal2016theoretically} and Recurrent Dropout are shown in Figure \ref{fig:recurrent_dropout}\cite{semeniuta2016recurrent}.

Krueger et al. \cite{krueger2016zoneout} proposed Zoneout in 2016. RNNs sometimes face the problem of vanishing gradient \cite{hochreiter1991untersuchungen, bengio1994learning}. Inspired by dropout of residual networks with the method Stochastic Depth\cite{huang2016deep} and Swapout\cite{singh2016swapout}, Zoneout randomly replaces the output at a certain timestamp in RNN with the output of a previous timestamp. Denote the hidden state transfer operation as $h_t = \mathcal{T}(h_{t-1},x_t)$, where $\mathcal{T}$ is often an affine transformation. To perform dropout operation is to replace the original transfer operation $\mathcal{T}$ with a new operation $\Tilde{\mathcal{T}}$. Standard dropout and Zoneout can be expressed as follows, respectively:
\begin{equation} \label{eq:zoneout 1}
    \begin{aligned}
    \mathrm{Dropout:}\quad \Tilde{\mathcal{T}} &= d_t\odot \mathcal{T} + (1-d_t)\odot 0 \\
    \mathrm{Zoneout:}\quad \Tilde{\mathcal{T}} &= d_t\odot \mathcal{T} + (1-d_t)\odot 1 \\
    \end{aligned}
    \vspace{-2pt}
\end{equation}
where $d_t$ is a mask vector obeying Bernoulli distribution. In this way, Zoneout passes the output of previous timestamp to the next timestamp with probability $p$ instead of dropping it with probability $p$. In this way Zoneout solves vanishing gradient problem while acting as a regularizer.

Merity et al. \cite{merity2018regularizing} proposed Weighted-dropped LSTM in 2017. Borrowing the idea of DropConnect\cite{wan2013regularization}, instead of dropping neuron activations, Weighted-dropped LSTM drops elements in the weight matrix. That is, for LSTM in Eq. \ref{eq:recurrent dropout 1},
% \begin{equation} \label{eq:weighted-dropped lstm 1}
%     \begin{aligned}
%     \begin{pmatrix}
%     \mathbf{i}_t\\\mathbf{f}_t\\\mathbf{o}_t\\\mathbf{g}_t
%     \end{pmatrix} &= \begin{pmatrix}
%     \begin{aligned}
%     \sigma(\mathbf{W}_i\mathbf{x}_t &+ \mathbf{U}_i\mathbf{h}_{t-1})\\
%     \sigma(\mathbf{W}_f\mathbf{x}_t &+ \mathbf{U}_f\mathbf{h}_{t-1})\\
%     \sigma(\mathbf{W}_o\mathbf{x}_t &+ \mathbf{U}_o\mathbf{h}_{t-1})\\
%     \tanh(\mathbf{W}_g\mathbf{x}_t &+ \mathbf{U}_g\mathbf{h}_{t-1})
%     \end{aligned}
%     \end{pmatrix} \\
%     \mathbf{c}_t &= \mathbf{f}_t\odot \mathbf{c}_{t-1} + \mathbf{i}_t\odot \mathbf{g}_t \\
%     \mathbf{h}_t &= \mathbf{o}_t\odot \tanh(\mathbf{c}_t)
%     \end{aligned}
% \end{equation}
Weighted-dropped LSTM drops weight matrix 
% (recurrent weights) $[\mathbf{U}_i, \mathbf{U}_f, \mathbf{U}_o, \mathbf{U}_g]$ used for the hidden state transfer, and also drops several other weight matrices (non- recurrent weights) 
$[\mathbf{W}_i, \mathbf{W}_f, \mathbf{W}_o]$ instead of hidden state $\mathbf{h}_{t-1}$.
% In contrast, Variational RNN Dropout\cite{gal2016theoretically} drops hidden state $\mathbf{h}_{t-1}$, as shown in Equation \ref{eq:recurrent dropout 2}.

In 2017, Melis et al.\cite{melis2018state} did a comprehensive review about the effectiveness of RNN applications in language models. The covered dropout methods perform on the input of RNN, feed-forward connections, recurrent connections, and the output of the last layer.

Zolna et al. \cite{zolna2018fraternal} proposed Fraternal Dropout in 2018. According to Ma et al.'s analysis\cite{ma2016dropout}, for networks trained with dropout, the expected outputs of training and prediction will differ. Using different dropout masks results in different outputs, i.e., the outputs are related to dropout masks, which is not what we want. We want model outputs to be irrelevant to dropout masks, they should be as same as possible under different masks, and their variance be as small as possible. Following this idea, Fraternal Dropout, trains two neural networks with the same structure and shared parameters. The only difference between them is their dropout masks. Fraternal Dropout optimize both objective functions of the two networks and the difference between their outputs.
% The overall objective function is
% \begin{equation} \label{eq:fraternal dropout 1}
%     \begin{aligned}
%     \mathcal{L}_{FD}(\mathbf{X}, \mathbf{Y}) = 
%     \sum_{t=1}^T &\dfrac{1}{2} \Big(\mathcal{L}^t\big(\mathbf{p}^t(\mathbf{z}^t, s_i^t; \theta), \mathbf{Y}\big) \\
%     &+ \mathcal{L}^t\big(\mathbf{p}^t(\mathbf{z}^t, s_j^t; \theta), \mathbf{Y}\big) \Big) \\
%     &+ \dfrac{\kappa}{mT}\sum_{t=1}^T \mathcal{R}_{FD}(\mathbf{z}^t; \theta), \\
%     \mathcal{R}_{FD}(\mathbf{z}^t; \theta) := \mathbb{E}_{s_i^t,s_j^t}& \left[||\mathbf{p}^t(\mathbf{z}^t, s_i^t; \theta) - \mathbf{p}^t(\mathbf{z}^t, s_j^t; \theta)||_2^2\right]
%     \end{aligned}
% \end{equation}
The authors also prove that the upper bound of the regularization term here is the objective function of expected linear dropout in \cite{ma2016dropout}.

\paragraph{\textbf{ResNets and others}}\label{para:drop resnet}

Residual network (ResNet) \cite{he2016deep} is a structure designed to solve problems such as vanishing gradient \cite{hochreiter1991untersuchungen, bengio1994learning} and long training time caused by excessive network depth. Let the output of the $l$th layer of network be $h_l$ and the transfer function from the $l-1$th layer to the $l$th layer be $f_l()$ (which may contain one or more convolution functions, batch normalization functions, and activation functions), then the feed-forward operation containing residual block is
\begin{equation} \label{eq:resnet 1}
    \begin{aligned}
    h_l = \mathrm{ReLU}\big(f_l(h_{l-1}) + \mathrm{id}(h_{l-1})\big)
    \end{aligned}
    \vspace{-2pt}
\end{equation}
where $\mathrm{id}$ denotes identity function, i.e., $h_{l-1}$ is passed to the $l$th layer directly. A series of dropout methods are also proposed for models with ResNet structure.

Huang et al. \cite{huang2016deep} proposed Stochastic Depth in 2016. Stochastic Depth randomly drops some operation blocks and retains only residual connections of that layer:
\begin{equation} \label{eq:stochastic depth 1}
    \begin{aligned}
    h_l = \mathrm{ReLU}\big(b_l f_l(h_{l-1}) + \mathrm{id}(h_{l-1})\big)
    \end{aligned}
    \vspace{-2pt}
\end{equation}
where $b_l\sim \mathrm{Bernoulli}(p_l)$ is retention probability. When $b_l=0$, the layer has only one identity function, which is equivalent to directly copying results of the previous layer, i.e., the network becomes shallower. In this way, it is possible to train a network with a desired shallow depth and use the whole deep network during testing, alleviating the vanishing gradient problem \cite{hochreiter1991untersuchungen, bengio1994learning} and the problem of long training time.

Kang et al. \cite{kang2016shakeout} proposed Shakeout in 2016. Different from standard dropout, Shakeout acts on the neurons not by making them choose between 0 and the original value, but between two new weights.
% as shown in Figure \ref{fig:shakeout}\cite{kang2016shakeout}.
% \begin{figure}
%   \centering
%   \includegraphics[width=\linewidth]{figs/shakeout.png}
%   \caption{Shakeout does not make neurons choose between $0$ and the original value like standard dropout (above), but between two new weights (below). \cite{kang2016shakeout}}
%   \label{fig:shakeout}
% \end{figure}
% In Figure \ref{fig:shakeout}, $\hat{r}$ is the mask vector, $\hat{r}_j\sim \dfrac{1}{1-\tau}\mathrm{Bernoulli}(1-\tau)$, $\tau\in[0, 1]$ is dropout probability; $\alpha=1/(1-\tau)$, $\beta(w) = c\cdot\mathrm{sign}(w)$, $c>0$; $w$ and $\Tilde{w}$ are the original and after-dropout weights, respectively. It can be seen that Shakeout has two parameters, $\tau$ and $c$, compared to the standard dropout which has only one parameter. 
The authors also show that Shakeout regularization combines three regularization terms, $L_0$, $L_1$ and $L_2$.

Li and Liu \cite{li2016whiteout} proposed Whiteout in 2016. Both standard dropout \cite{hinton2012improving, srivastava2014dropout} and Shakeout \cite{kang2016shakeout} only introduce Bernoulli noise, while Whiteout introduces Gaussian noise into training process. This is done by adding additive or multiplicative Gaussian noise to the output of each neuron. 
% For the $j$th point of the $l$th layer, Whiteout adds additive noise as
% \begin{equation} \label{eq:whiteout 1}
%     \begin{aligned}
%     \Tilde{X}_j^{(l)} &= X_j^{(l)} + e_{jk},\\
%     &\mathrm{where}\ e_{jk} \overset{\mathrm{ind}}{\sim} N\left(0, \sigma^2|w_{jk}^{(l)}|^{-\gamma} + \lambda \right)
%     \end{aligned}
% \end{equation}
% , adds mutiplicative noise as
% \begin{equation} \label{eq:whiteout 2}
%     \begin{aligned}
%     \Tilde{X}_j^{(l)} &= X_j^{(l)} \epsilon_{jk},\\
%     &\mathrm{where}\ \epsilon_{jk} \overset{\mathrm{ind}}{\sim} N\left(1, \sigma^2|w_{jk}^{(l)}|^{-\gamma} + \lambda \right) \\
%     &= X_j^{(l)} + e^{\prime}_{jk},\\
%     &\mathrm{where}\ e^{\prime}_{jk} \overset{\mathrm{ind}}{\sim} N\left(0, \big(X_j^{(l)}\big)^2 \big(\sigma^2|w_{jk}^{(l)}|^{-\gamma} + \lambda \big) \right)
%     \end{aligned}
% \end{equation}
Whiteout is the first noise injection regularization technique (NIRT) that imposes an extensive $L_\gamma$, $\gamma\in (0, 2)$ sparse regularization without involving $L_2$ regularization.

Gastaldi \cite{gastaldi2017shake} proposed Shake-Shake in 2017. Shake-Shake is used for three-way ResNet. The original feed-forward operation is
\begin{equation} \label{eq:shake-shake 1}
    \begin{aligned}
    x_{i+1} = \sigma(x_i + \mathcal{F}(x_i, \mathcal{W}_i^{(1)}) + \mathcal{F}(x_i, \mathcal{W}_i^{(2)}))
    \end{aligned}
    \vspace{-2pt}
\end{equation}
Shake-Shake introduces random variables $\alpha_i\sim U(0, 1)$ to assign weights to the two-way transfer function:
\begin{equation} \label{eq:shake-shake 2}
    \begin{aligned}
    x_{i+1} = \sigma(x_i + \alpha_i\mathcal{F}(x_i, \mathcal{W}_i^{(1)}) + (1-\alpha_i)\mathcal{F}(x_i, \mathcal{W}_i^{(2)}))
    \end{aligned}
    \vspace{-2pt}
\end{equation}
It takes a weight of 0.5 for each path when testing:
\begin{equation} \label{eq:shake-shake 3}
    \begin{aligned}
    x_{i+1} = \sigma(x_i + 0.5\mathcal{F}(x_i, \mathcal{W}_i^{(1)}) + 0.5\mathcal{F}(x_i, \mathcal{W}_i^{(2)}))
    \end{aligned}
    \vspace{-2pt}
\end{equation}

Yamada et al. \cite{yamada2019shakedrop} proposed ShakeDrop in 2018. Stochastic Depth \cite{huang2016deep} simply drops or retains a layer, Shake-Shake \cite{gastaldi2017shake } can assign weights to different pathways but is applied only to a three-way ResNet. ShakeDrop combines both functions. It has two parameters $\alpha$ and $\beta_l$ to control the assigned weights. $b_l\sim \mathrm{Bernoulli}(p_l)$ is the dropout ratio. ShakeDrop is expressed as
\begin{equation} \label{eq:shakedrop 1}
    G(x) = \begin{cases}
    x + (b_l + \alpha - b_l\alpha)F(x),\quad \text{in train-fwd} \\
    x + (b_l + \beta - b_l\beta)F(x),\quad \text{in train-bwd} \\
    x + \mathbb{E}[b_l + \alpha - b_l\alpha]F(x),\quad \text{in test}
    \end{cases}
    \vspace{-2pt}
\end{equation}

Larsson et al. \cite{larsson2016fractalnet} proposed DropPath in 2016. The authors first proposed a network structure Fractalnet, which achieves extremely deep neural networks based on self-similarity of structure.
% as shown in Figure \ref{fig:fractalnet}\cite{larsson2016fractalnet}.
% \begin{figure}
%   \centering
%   \includegraphics[width=\linewidth]{figs/fractalnet.png}
%   \caption{Fractalnet achieves extremely deep network structure based on self-similarity. \cite{larsson2016fractalnet}}
%   \label{fig:fractalnet}
% \end{figure}
It is shown that Fractalnet can also alleviate vanishing gradient problem in deep neural networks, just as ResNet does. The authors then proposed a regularization approach for Fractalnet, which is randomly dropping sub-paths between input and output within each fractal block. Just as standard dropout can reduce dependencies between neurons, this operation reduces dependencies between sub-paths to act as a regularizer \cite{larsson2016fractalnet}.
% \begin{figure}
%   \centering
%   \includegraphics[width=\linewidth]{figs/droppath.png}
%   \caption{DropPath randomly drops subpaths in FractalNet. \cite{larsson2016fractalnet}}
%   \label{fig:droppath}
% \end{figure}

Zoph et al. \cite{zoph2018learning} proposed Scheduled DropPath in 2018 as an improvement to DropPath \cite{larsson2016fractalnet}. Dropout ratio increases linearly with the number of training epochs rather than a fixed value, and Scheduled DropPath achieves better performance than the original DropPath.

Singh et al. \cite{singh2016swapout} proposed Swapout in 2016, a synthesis of standard dropout and Stochastic Depth. Let $X$ be the input of a block in neural network, and the block computes the output $Y = F(X)$. The output of the $u$th neuron in the block is noted as $F^{(u)}(X)$. $\Theta$ is a tensor with the same shape as $F(X)$ whose elements obey Bernoulli distribution. Standard dropout makes 
% computes its output as
% \begin{equation} \label{eq:swapout 1}
%     \begin{aligned}
%     Y = \Theta \odot F(X)
%     \end{aligned}
%     \vspace{-2pt}
% \end{equation}
% That is, 
the output of each neuron choosed from $\{0$, $F^{(u)}(X)\}$. Stochastic Depth is such that the output of each neuron is choosed from $\{X^{(u)}$, $X^{(u)} + F^{(u)}(X)\}$. Swapout, on the other hand, extends the range of possible output values. For a layer with $N$ blocks $F_1, \dots, F_N$, define $N$ independent Bernoulli tensor $\Theta_1, \dots, \Theta_N$ such that the output is computed as
\begin{equation} \label{eq:swapout 2}
    \begin{aligned}
    Y = \sum_{i=1}^N \Theta_i \odot F_i(X)
    \end{aligned}
    \vspace{-2pt}
\end{equation}
In this way, the output of each neuron has $2^N$ possible values. Consider the simplest case of $N=2$ in ResNet, 
\begin{equation} \label{eq:swapout 3}
    \begin{aligned}
    Y = \Theta_1 \odot X + \Theta_2 \odot F(X)
    \end{aligned}
    \vspace{-2pt}
\end{equation}
The output of each neuron can take 4 values: $\{0$, $X^{(u)}$, $F^{(u)}(X)$, $X^{(u)} + F^{(u)}(X)\}$. Each value corresponds to neuron's state as
\begin{enumerate}
    \vspace{-4pt}
    \item $0$: dropped
    \item $X^{(u)}$: skipped by the residual connection
    \item $F^{(u)}(X)$: normal
    \item $X^{(u)} + F^{(u)}(X)$: a complete residual unit
    \vspace{-4pt}
\end{enumerate}

Zhou et al. proposed DropHead \cite{zhou2020scheduled} in 2020, dropping attention heads in multi-head attention mechanism, which is a core component of Tranformer. It also adaptively adjusts dropout ratio during training to achieve better performance.

Dropout methods that drop neuron groups are summarized in Table \ref{tab:drop neuron groups}. The application scenarios of dropping neuron groups are usually the models with certain structures such as CNN, RNN, ResNet, or Transformer. This type of dropout methods can boost the performance of these models, while its implementation is also subject to the model structures.

\subsection{Drop Embeddings}\label{subsec:drop embeddings}

In some machine learning tasks, the input data is first converted to embeddings then goes through the model. Dropout methods introduced in this section drop embeddings during training. For example, in recommendation, some dropout methods drop embeddings of user and item interaction histories during training to cope with the cold-start problem, while others randomly drop user and item feature embeddings to handle the possible missing information problem in real scenarios.

% In recommendation tasks, the model predicts possible future interactions based on interaction histories of users and items. When a new user or item appears, the absence of its interaction history makes it difficult for the model to predict. This scenario is called \textit{Cold Start}. Works addressing cold-start problem have been emerging in recent years \cite{xu2021multi, qian2020attribute, zhu2019addressing, zhang2020deep, li2019both, lu2020meta}. Some logistic regression-based models, such as factorization machine and its successors, utilize attributes of users and items to make predictions, not only relying on the interaction history, alleviating cold-start problem to some extent. However, in real world scenarios, the values of user and item attributes may also be missing, which still leads to model prediction difficulties.
% For such problems of missing information, there are also drop-based methods in recommendation scenarios. They drop embeddings of users and items during training. The motivation is to simulate the possible missing information situation during testing. Therefore, the final model is less dependent on certain kinds of information and better coping with the missing information problem in real scenarios.

Volkovs et al. \cite{volkovs2017dropoutnet} proposed DropoutNet in 2017. The embedding matrices of users and items are noted as $\mathbf{U}$ and $\mathbf{I}$. The embedding vectors of the $u$th user and $i$th item are $\mathbf{U}_u$ and $\mathbf{I}_i$, respectively.
% After DNN they are represented as $\hat{\mathbf {U}}$ and $\hat{\mathbf{I}}$.
The context information matrices of users and items are $\mathbf{\Phi}^{\mathcal{U}}$ and $\mathbf{\Phi}^{\mathcal{I}}$, respectively. DropoutNet proposed a new form of objective dealing with the problem of missing interaction history. Previous models would add extra terms of context information into objective, hoping these newly added terms can be useful when the terms of interaction history are not available. However, it is difficult to determine the weights of these two components. DropoutNet's objective handles this problem automatically:
\begin{equation} \label{eq:dropoutnet 1}
    \begin{aligned}
    L &= \sum_{u, i}\big(\mathbf{U}_u\mathbf{I}_i^T - f_{\mathcal{U}}(\mathbf{U}_u, \mathbf{\Phi}_u^{\mathcal{U}})f_{\mathcal{I}}(\mathbf{I}_i, \mathbf{\Phi}_i^{\mathcal{I}})^T \big)^2
    % &= \sum_{u, i}\big(\mathbf{U}_u\mathbf{I}_i^T - \hat{\mathbf{U}}_u\hat{\mathbf{I}}_i^T\big)^2
    \end{aligned}
    \vspace{-2pt}
\end{equation}
During training, DropoutNet randomly drops a portion of input $\mathbf{U}_u$ or $\mathbf{I}_i$ by setting them to zero. For training instances that are kept, the objective will make the model ignore context information part ($\mathbf{\Phi}^{\mathcal{U}}$ and $\mathbf{\Phi}^{\mathcal{I}}$) as much as possible like Equation \ref{eq:dropoutnet 1} shows. For training instances with user or item inputs dropped, the objective will make the model rely as much as possible on context information part, as shown in Equation \ref{eq:dropoutnet 2}.
\begin{equation} \label{eq:dropoutnet 2}
    \begin{aligned}
    \textit{u}\text{ cold start: }
    L_{ui} &= \big(\mathbf{U}_u\mathbf{I}_i^T - f_{\mathcal{U}}(\mathbf{0}, \mathbf{\Phi}_u^{\mathcal{U}}) f_{\mathcal{I}}(\mathbf{I}_i, \mathbf{\Phi}_i^{\mathcal{I}})^T \big)^2 \\
    \textit{i}\text{ cold start: }
    L_{ui} &= \big(\mathbf{U}_u\mathbf{I}_i^T - f_{\mathcal{U}}(\mathbf{U}_u, \mathbf{\Phi}_u^{\mathcal{U}}) f_{\mathcal{I}}(\mathbf{0}, \mathbf{\Phi}_i^{\mathcal{I}})^T \big)^2
    \end{aligned}
    \vspace{-2pt}
\end{equation}
% For the dropped instances, besides replacing embeddings with $\mathbf{0}$, DropoutNet can also replace them with the average of their neighbors' embeddings, i.e.:
% \begin{equation} \label{eq:dropoutnet 3}
%     \begin{aligned}
%     \mathbf{U}_u &\rightarrow \mathrm{mean}_{i\in\mathcal{I}(u)} \mathbf{I}_i \\
%     \mathbf{I}_i &\rightarrow \mathrm{mean}_{u\in\mathcal{U}(i)} \mathbf{U}_u
%     \end{aligned}
% \end{equation}

% 看有无2021的
\begin{table*}[ht]
    \centering
    \caption{Table of methods that drop embeddings or input information.}
    \vspace{-6pt}
\begin{threeparttable}
    \begin{tabular}{llllll}
        \toprule
Method & Year & Category & Brief Description & \makecell[l]{Original\\Scenario} & Source \\
        \midrule
DropoutNet\cite{volkovs2017dropoutnet} & 2017 & 2\dag & Randomly drop interactions & Recom. & NeurIPS \\
ACCM\cite{shi2018attention} & 2018 & 2 & Drop interactions \& use attention mechanism & Recom. & CIKM \\
AFS\cite{shi2019adaptive} & 2019 & 2 & Drop interactions and attribute values & Recom. & CIKM \\
WordDropout\cite{sennrich2016edinburgh} & 2016 & 3.1\dag & Drop words in machine translation & NLP & WMT16 \\
BERT\cite{devlin2019bert} & 2018 & 3.1 & Mask and predict tokens in pre-training phase & NLP & NAACL \\
Mask-Predict\cite{ghazvininejad2019mask} & 2019 & 3.1 & Mask and regenerate words in machine translation & NLP & EMNLP \\
ERNIE\cite{sun2019ernie} & 2019 & 3.1 & Incorporate human knowledge into pre-training & NLP & arxiv \\
Whole Word Masking\cite{cui2019pre} & 2019 & 3.1 & Randomly mask chinese words & NLP & arxiv \\
Mask and Infill\cite{wu2019mask} & 2019 & 3.1 & Mask and infill tokens in pre-training & NLP & IJCAI \\
AMS\cite{ye2019align} & 2019 & 3.1 & Incorporate general knowledge using ConceptNet & NLP & arxiv \\
PEGASUS\cite{zhang2020pegasus} & 2019 & 3.1 & Mask sentences for summary generation & NLP & ICML \\
Token Drop\cite{zhang2020token} & 2020 & 3.1 & Drop tokens instead of words & NLP & COLING \\
Selective Masking \cite{gu2020train} & 2020 & 3.1 & Introduce a task-guided pre-training stage & NLP & EMNLP \\
S3-Rec\cite{zhou2020s3} & 2020 & 3.1 & Enhance interaction sequence like BERT & Recom. & CIKM \\
CutOut\cite{devries2017improved} & 2017 & 3.2\dag & Drop a square region on the input image & CV & arxiv \\
Random Erasing\cite{zhong2020random} & 2017 & 3.2 & Drop a rectangular region on the input image & CV & AAAI \\
Hide-and-Seek\cite{singh2017hide} & 2017 & 3.2 & Drop several square regions & CV & ICCV \\
Mixup\cite{zhang2017mixup} & 2017 & 3.2 & Take linear interpolations of training instances as input & CV & ICLR \\
Manifold Mixup\cite{verma2019manifold} & 2019 & 3.2 & Generalize Mixup to feature level & CV & ICML \\
CutMix\cite{yun2019cutmix} & 2019 & 3.2 & Replace regions of one image with another's & CV & ICCV \\
GridMask\cite{chen2020gridmask} & 2020 & 3.2 & Drop regularly tiled square regions & CV & arxiv \\
Attentive CutMix\cite{walawalkar2020attentive} & 2020 & 3.2 & Improve CutMix with attention mechanism  & CV & ICASSP \\
MAE\cite{he2021masked} & 2021 & 3.2 & Mask and reconstruct patches of input images & CV & arxiv \\
GraphSAGE\cite{hamilton2017inductive} & 2017 & 3.3\dag & Randomly sample nodes & Graph & NeurIPS \\
FastGCN\cite{chen2018fastgcn} & 2018 & 3.3 & Sample nodes from the whole graph & Graph & ICLR \\
AS-GCN\cite{huang2018adaptive} & 2018 & 3.3 & Node sampling layer by layer & Graph & NeurIPS \\
GAT\cite{velivckovic2017graph} & 2018 & 3.3 & Attention on edges & Graph & ICLR \\
LADIES\cite{zou2019layer} & 2019 & 3.3 & Adaptively sample nodes by layer & Graph & NeurIPS \\
GRAND\cite{feng2020graph} & 2020 & 3.3 & Random propagation on graph & Graph & NeurIPS \\
SGAT\cite{ye2021sparse} & 2020 & 3.3 & Learn sparse attention coefficients on graph & Graph & TKDE \\
DropEdge\cite{rong2019dropedge} & 2020 & 3.3 & Randomly drop edges & Graph & ICLR \\
        \bottomrule
    \end{tabular}
\begin{tablenotes}
\footnotesize
\item[\dag] 2 refers to dropping embeddings, 3.1 dropping one-dimensional information, 3.2 dropping two-dimensional information, and 3.3 dropping graph information. 
\end{tablenotes}
\end{threeparttable}
    \label{tab:drop embeddings}
    \vspace{-12pt}
\end{table*}

Shi et al. \cite{shi2018attention} proposed ACCM model in 2018. DropoutNet \cite{volkovs2017dropoutnet}, while automatically processing both interaction history information and context information, is implemented in such a way that its objective function is completely backward to one side. When interaction history is available, the model tends to completely rely on it and ignore attribute information; when interaction history is missing, the model tends to completely rely on attribute information. ACCM model, on the other hand, achieves flexible control of the weights of the two components by using attention mechanism \cite{vaswani2017attention}.
% \begin{figure}
%   \centering
%   \subcaptionbox{ACCM structure\label{fig:ACCM_1}}
%     {\includegraphics[width=0.49\linewidth]{figures/ACCM_1.png}}
%   \subcaptionbox{ACCM training\label{fig:ACCM_2}}
%     {\includegraphics[width=0.49\linewidth]{figures/ACCM_2.png}}
%   \caption{ACCM\cite{shi2018attention}}
%   \label{fig:ACCM}
% \end{figure}

The model contains a user part and an item part. Each part computes embeddings by both interaction history and context information. For the user part, the attention network computes attention weights for two kinds of information separately, and then obtains the final embedding $\mathbf{u}$:
\begin{equation} \label{eq:accm 1}
    \begin{aligned}
    h_{CF}^u = \mathbf{h}^T \tanh&(\mathbf{W}\mathbf{u}_{CF} + \mathbf{b}),\ 
    h_{CB}^u = \mathbf{h}^T \tanh(\mathbf{W}\mathbf{u}_{CB} + \mathbf{b}) \\
    a_{CF}^{u} &= \dfrac{\exp(h_{CF}^u)}{\exp(h_{CF}^u) + \exp(h_{CB}^u)} = 1 - a_{CB}^{u} \\
    \mathbf{u} &= a_{CF}^{u}\mathbf{u}_{CF} + a_{CB}^{u}\mathbf{u}_{CB}
    \end{aligned}
    \vspace{-2pt}
\end{equation}
Item embedding $\mathbf{v}$ is generated in the same way as $\mathbf{u}$. The model prediction is
\begin{equation} \label{eq:accm 2}
    \begin{aligned}
    y = b_g + b_u + b_v + \mathbf{u}\mathbf{v}
    \end{aligned}
    \vspace{-2pt}
\end{equation}
Like DropoutNet, the interaction history embeddings are randomly dropped during training, replaced with random vectors:
% as shown in Figure \ref{fig:ACCM_2}.
\begin{equation} \label{eq:accm 3}
    \begin{aligned}
    \mathbf{u} &= a_{CF}^{u}[(1-c^u)\mathbf{u}_{CF} + c^u\mathbf{u}_r] + a_{CB}^{u}\mathbf{u}_{CB} \\
    \mathbf{v} &= a_{CF}^{v}[(1-c^v)\mathbf{v}_{CF} + c^v\mathbf{v}_r] + a_{CB}^{v}\mathbf{v}_{CB} \\
    y &= b_g + c^u b_u + c^v b_v + \mathbf{u}\mathbf{v}
    \end{aligned}
    \vspace{-2pt}
\end{equation}
where $c^u, c^v \sim \mathrm{Bernoulli}(p)$, $p$ is the dropout probability. $\mathbf{u}_r, \mathbf{v}_r$ are random vectors with the same initial distribution of user and item vectors. By using attention mechanism and randomly dropping embddings, ACCM better solves cold start problem in recommendation.

Similar to missing interaction history, missing content information is sometimes encountered in recommendation. Since cold-start problem can be better solved by embedding dropout and attention mechanism together, missing attribute problem can also be solved in a similar way, which is the idea of AFS (Adaptive Feature Sampling) \cite{shi2019adaptive}. AFS drops a portion of user and item context information randomly during training to simulate missing attribute values, making the model more robust when testing.

Dropout methods that drop embeddings are summarized in Table \ref{tab:drop embeddings}. Recommendation models are usually the application scenarios of dropping embeddings, whose input information usually needs to be converted into vector representations for model operations. Dropping embeddings can be effective in such scenarios, while it can only be used when there are embeddings of input data.

\subsection{Drop Input Information}\label{subsec:drop inputs}

Some dropout methods drop part of input information directly during training, which serves for various purposes under different scenarios, such as regularization, data augmentation, or data representation enhancement of pre-training stage.

\subsubsection{One-dimensional Information}\label{subsubsec:drop 1d info}

Sennrich et al. \cite{sennrich2016edinburgh} in 2016 use WordDropout in machine translation to drop out words from input data.

Ghazvinine et al. \cite{ghazvininejad2019mask} proposed Mask-Predict in 2019. While most machine translation systems generate text from left to right, Mask-Predict uses a masking approach to train the model. It first predicts all target words and then iteratively masks and regenerates a subset of words in which the model has least confidence. Unlike BERT \cite{devlin2019bert}, this paper does not use masked language model for pre-training but uses it directly with Mask-Predict to generate text.

Zhang et al. \cite{zhang2020token} in 2020 proposed Token Drop mechanism for neural network machine translation. WordDropout \cite{sennrich2016edinburgh} randomly drops words from sentences, while Token Drop method drops tokens.

Devlin et al. \cite{devlin2019bert} proposed BERT in 2019. In the pre-training phase, 15\% of the tokens are randomly masked. These masked tokens are then predicted in both directions using a self-attentive transformer to enhance the data representation in pre-training phase. After BERT came out, many BERT-like or BERT-based methods have been proposed for enhancing data representation in NLP pre-training.

Cui et al. \cite{cui2019pre} in 2019 proposed Whole Word Masking for Chinese language models. BERT randomly masks words for English language models, but randomly masking Chinese characters is less appropriate because Chinese characters may not be a complete semantic unit. Whole Word Masking masks words instead of Chinese characters when training Chinese language models.

Sun et al. \cite{sun2019ernie} proposed ERNIE in 2019. ERNIE introduces human knowledge into word vector training. It achieves this by considering masking operations of three levels. The first level, like BERT, randomly masks English or Chinese words. The second level randomly masks phrases identified by existing toolkits, incorporating phrase information into training. The third level randomly masks entities pre-defined by human, incorporating prior human knowledge into the training of word vectors.

Wu et al. \cite{wu2019mask} proposed Mask and Infill in 2019. The pre-training phase is divided into a masking phase and a filling phase used to accomplish the task of sentiment transfer. %The masking phase masks only word tokens with affective overtones (e.g., positive/negative), and the task of filling phase is to predict these words.

Ye et al. \cite{ye2019align} proposed AMS method in 2019. A general knowledge Q\&A dataset is generated through ConceptNet \cite{speer2017conceptnet}, and the general knowledge concepts in each utterance of this dataset are masked and predicted so that the model learns general knowledge through this process.

Zhang et al. \cite{zhang2020pegasus} in 2019 proposed PEGASUS for summary generation tasks. During pre-training, not only word tokens are randomly masked, but also important sentences. These sentences are part of the summary to be generated and need to be predicted by the model.

Wang et al. \cite{wang2020semantic} in 2019 imitates BERT to introduce dropout in speech recognition, which randomly masks acoustic signals and features in the input audio.

Selective Masking \cite{gu2020train} proposed by Gu et al. in 2020 introduced a task-guided pre-training stage between general pre-training and fine-tuning stage. 

There are similarities between the input of sequential recommendation and the input of NLP tasks, for both of them are temporal one-dimensional information. So there are also recommendation models that borrows the idea of BERT: Zhou et al. \cite{zhou2020s3} proposed S3-Rec in 2020. It divides the recommendation task into a pre-training stage and a fine-tuning stage just like BERT. S3-Rec randomly masks a portion of item ids and attributes in pre-training phase to enhance the representation between item ids, item attributes, and item sequences.

Dropout methods that drop one-dimentional input information are summarized in Table \ref{tab:drop embeddings}, which are mainly applied in NLP or sequential recommendation tasks, where input data is organized as temporal one-dimensional sequences.

% \begin{table*}[ht]
%     \renewcommand{\arraystretch}{1.1}
%     \centering
%     \newcolumntype{Y}{>{\raggedleft\arraybackslash}X}  %X单元格居左, Y单元格居右
% 	\newcolumntype{Z}{>{\centering\arraybackslash}X}  %Z单元格居中
%     \caption{Drop One-dimentional Information}
%     \begin{tabularx}{0.8\textwidth}{llXlX}
%         \toprule
%             Method & Year & Category & Application & Source \\
%         \midrule
%         WordDropout\cite{sennrich2016edinburgh}              & 2016 & Drop 1d Info   & NLP    & WMT16    \\
% Mask-Predict\cite{ghazvininejad2019mask}             & 2019 & Drop 1d Info   & NLP    & EMNLP    \\
% BERT\cite{devlin2019bert}                     & 2018 & Drop 1d Info   & NLP    & NAACL    \\
% ERNIE\cite{sun2019ernie}                    & 2019 & Drop 1d Info   & NLP    & arxiv \\
% Whole Word Masking\cite{cui2019pre}       & 2019 & Drop 1d Info   & NLP    & arxiv \\
% Mask and Infill\cite{wu2019mask}          & 2019 & Drop 1d Info   & NLP    & arxiv \\
% AMS\cite{ye2019align}                      & 2019 & Drop 1d Info   & NLP    & arxiv \\
% PEGASUS\cite{zhang2020pegasus}                  & 2019 & Drop 1d Info   & NLP    & ICML     \\
% S3-Rec\cite{zhou2020s3}                   & 2020 & Drop 1d Info   & Sequential Recommendation & CIKM     \\
%         \bottomrule
%     \end{tabularx}
%     \label{tab:drop 1d info}
% \end{table*}

\subsubsection{Two-dimensional Information}\label{subsubsec:drop 2d info}

Existing data enhancement methods can be broadly classified into three categories: spatial transformation, color distortion, and information dropping. Dropping two-dimensional input information is generally regarded as a data enhancement method of information dropping.

DeVries and Taylor \cite{devries2017improved} proposed CutOut in 2017. For every training image, CutOut randomly selects a square region and sets the pixel values within this region to zero. The difference of CutOut with methods in Section \ref{subsubsec:drop neuron groups} such as SpatialDropout \cite{tompson2015efficient} or DropBlock \cite{ghiasi2018dropblock} is that CutOut is performed at the level of input information. Compared to dropping model structure, it is easier to implement by directly dropping a part of the input image. In addition, dropping input information is equivalent to generating a new training sample, so there is no need to multiply the neuron output by a factor to eliminate the bias during testing as the methods in Section \ref{subsec:drop structures} does.

Zhong et al. \cite{zhong2020random} proposed Random Erasing in 2017. Similar to CutOut \cite{devries2017improved}, the training image is covered with a rectangular box with random position and random size. The pixel values within the rectangular box are also random.

Singh et al. \cite{singh2017hide} proposed Hide-and-Seek in 2017. CutOut \cite{devries2017improved} and Random Erasing \cite{ zhong2020random} drop only one rectangular region for each input image, while Hide-and-Seek divides the image into $S\times S$ small squares, each square is dropped with $p_{hide}$ probability. The purpose of Hide-and-Seek is to let the model be capable of extracting features from other parts of the image after the most discriminative part has been dropped, preventing the model from relying too much on certain parts.

\begin{figure}
  \centering
  \includegraphics[width=\linewidth]{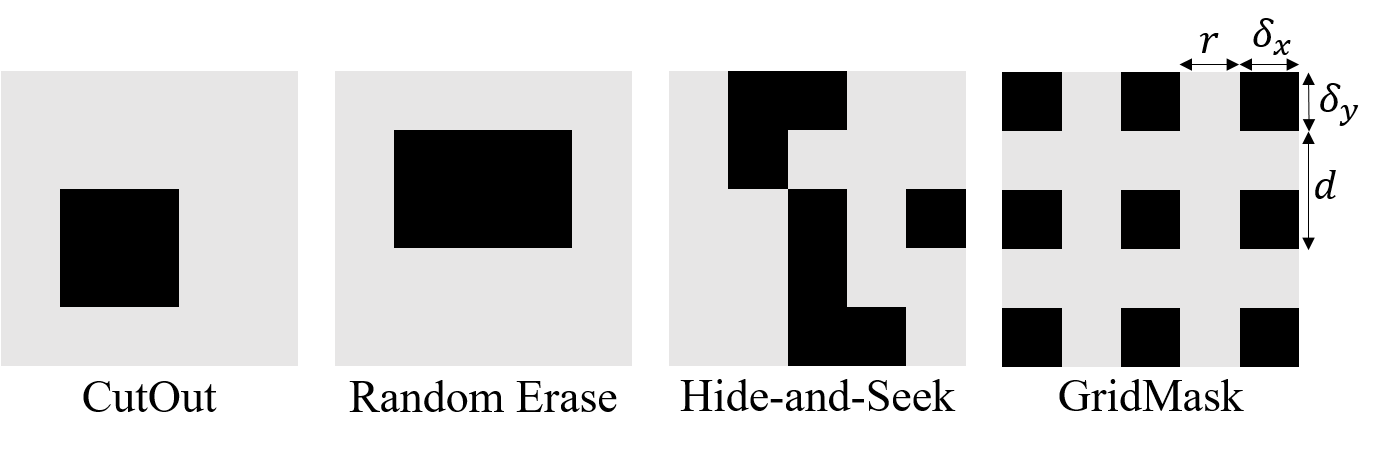}
  \vspace{-16pt}
  \caption{Comparison of dropout patterns of CutOut \cite{devries2017improved}, Random Erasing \cite{zhong2020random}, Hide-and-Seek \cite{singh2017hide} and GridMask \cite{chen2020gridmask}.}
  \label{fig:gridmask}
  \vspace{-10pt}
\end{figure}

Chen et al. \cite{chen2020gridmask} proposed GridMask in 2020. Dropout pattern of GridMask is a number of equally spaced square regions tiled on a plane, determined by four parameters $(r, d, \delta_x, \delta_y)$. The dropout pattern of GridMask is more regular compared to CutOut, Random Erasing and Hide-and-Seek. Compared to AutoAugment \cite{cubuk2019autoaugment} which employs reinforcement learning to search for dropout patterns, GridMask consumes much less training cost. The above four methods are schematically shown in Figure \ref{fig:gridmask}.

\emph{Dropping} input data can also be seen as \emph{introducing noise} to input data, while this noise is Bernoulli noise. Some of the following data enhancement methods do not necessarily \emph{drop} input data, but \emph{introduce noise} to input data.

Zhang et al. \cite{zhang2017mixup} proposed Mixup in 2017. Mixup augments data in a simple way: the linear interpolations of training instances are also taken as training instances. Specifically, for training instances $(\mathbf{x}_i, y_i)$ and $(\mathbf{x}_j, y_j)$,
\begin{equation} \label{eq:mixup 1}
    \begin{aligned}
    \Tilde{\mathbf{x}} &= \lambda \mathbf{x}_i + (1 - \lambda)\mathbf{x}_j \\
    \Tilde{y} &= \lambda y_i + (1 - \lambda)y_j
    \end{aligned}
    \vspace{-4pt}
\end{equation}
Mixup takes $(\Tilde{\mathbf{x}}, \Tilde{y})$ as a training instance as well.

Verma et al. \cite{verma2019manifold} proposed Manifold Mixup in 2019. Manifold Mixup generalizes Mixup \cite{zhang2017mixup} operation to feature level. The motivation is that features have higher-order semantic information, and interpolation in feature level could yield more meaningful samples.

\begin{figure}
  \centering
  \includegraphics[width=0.8\linewidth]{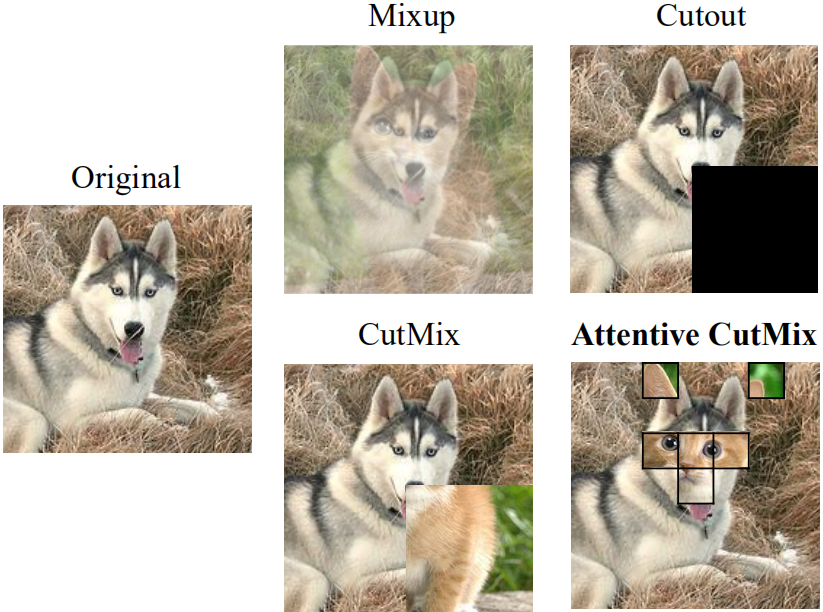}
  \vspace{-6pt}
  \caption{Comparison of dropout patterns of Mixup \cite{verma2019manifold}, CutOut \cite{devries2017improved}, CutMix \cite{yun2019cutmix} and Attentive CutMix. \cite{walawalkar2020attentive}}
  \label{fig:attentivecutmix}
  \vspace{-10pt}
\end{figure}

Yun et al. \cite{yun2019cutmix} proposed CutMix in 2019, which is an improvement on Mixup \cite{zhang2017mixup} and Cutout \cite{devries2017improved}. Cutout fills part of the image with meaningless regions, which is not conducive to the model making full use of training data. Mixup uses linear interpolation to augment the data, however these newly produced images are not natural images. CutMix, on the other hand, randomly selects some rectangular regions of the image $\mathbf{x}_A$ and replaces them with regions at the same locations of image $\mathbf{x}_B$. The corresponding labels are replaced by a combination of the two images: 
\begin{equation} \label{eq:cutmix 1}
    \begin{aligned}
    \Tilde{\mathbf{x}} &= \mathbf{M}\odot \mathbf{x}_A + (1 - \mathbf{M})\odot \mathbf{x}_B \\
    \Tilde{y} &= \lambda y_A + (1 - \lambda)y_B
    \end{aligned}
    \vspace{-4pt}
\end{equation}
where $\mathbf{M}$ is the masking matrix.
%This performs a random substitution in such a way that all pixels are real pixels.
Walawalkar et al. \cite{walawalkar2020attentive} proposed Attentive CutMix in 2020, which further improved CutMix by using an attention mechanism to select the most discriminative regions for replacement. Operations of Mixup, CutOut, CutMix and Attentive CutMix are shown in Figure \ref{fig:attentivecutmix} from the original paper\cite{walawalkar2020attentive}.

He et al. \cite{he2021masked} proposed Masked Autoencoders (MAE) in 2021. It randomly masks patches of the input image and reconstructs the missing pixels during pre-training.

Dropout methods that drop two-dimensional information are summarized in Table \ref{tab:drop embeddings}. They are mainly applied in computer vision tasks, where input data is organized as pixel matrices.

% \begin{table*}[ht]
%     \renewcommand{\arraystretch}{1.1}
%     \centering
%     \newcolumntype{Y}{>{\raggedleft\arraybackslash}X}  %X单元格居左, Y单元格居右
% 	\newcolumntype{Z}{>{\centering\arraybackslash}X}  %Z单元格居中
%     \caption{Drop Two-dimentional Information}
%     \begin{tabularx}{0.75\textwidth}{XlXlX}
%         \toprule
%             Method & Year & Category & Application & Source \\
%         \midrule
%         CutOut\cite{devries2017improved}                   & 2017 & Drop 2d Info   & CV     & arxiv \\
% Random Erasing\cite{zhong2020random}           & 2017 & Drop 2d Info   & CV     & AAAI     \\
% Hide-and-Seek\cite{singh2017hide, singh2018hide}            & 2017 & Drop 2d Info   & CV     & ICCV \\
% GridMask\cite{chen2020gridmask}                 & 2020 & Drop 2d Info   & CV     & arxiv \\
% Mixup\cite{zhang2017mixup}                    & 2017 & Drop 2d Info   & CV     & ICLR     \\
% Manifold Mixup\cite{verma2019manifold}           & 2019 & Drop 2d Info   & CV     & ICML     \\
% CutMix\cite{yun2019cutmix}                   & 2019 & Drop 2d Info   & CV     & ICCV     \\
% Attentive CutMix\cite{walawalkar2020attentive}         & 2020 & Drop 2d Info   & CV     & arxiv \\
%         \bottomrule
%     \end{tabularx}
%     \label{tab:drop 2d info}
% \end{table*}

\subsubsection{Graph Information}\label{subsubsec:drop graph info}

Graph neural networks (GNNs) have a wide range of applications in various tasks such as node classification, cluster detection, and recommender systems \cite{weng2020gain, huan2021search, wang2021forecasting}. In the training of GNNs, some methods randomly drop nodes or edges and use only part of graph information for training, serving as a regularization technique.

\paragraph{\textbf{Drop Nodes}}

Hamilton et al. \cite{hamilton2017inductive} proposed GraphSAGE (SAmple and AGGreGatE) in 2017. Before this, GCNs were generally trained in the way of \emph{transductive learning}, which requires all nodes to be visible at training time and needs to calculate a node's embedding by all its neighbors. GraphSAGE, on the other hand, adopts \emph{inductive learning}, which requires only some of the neighbors to predict a node's embedding. To achieve this, GraphSAGE does not directly train node embeddings but trains aggregation functions, which compute node embeddings from its neighbors. When a new node is added to the graph during testing, the trained aggregation functions can predict its embedding from its neighbors. Parameters of the aggregation functions are trained with an unsupervised learning objective that makes the representations of closer nodes more similar and farther nodes less similar,
\begin{equation} \label{eq:graphsage 1}
    \begin{aligned}
    J_{\mathcal{G}}(\mathbf{z}_u) = &-\log\big(\sigma(\mathbf{z}_u^T\mathbf{z}_v)\big) \\
    &- Q\cdot \mathbb{E}_{v_n\sim P_n(v)}\log\big(\sigma(-\mathbf{z}_u^T\mathbf{z}_{v_n}) \big)
    \end{aligned}
    \vspace{-2pt}
\end{equation}
$v$ is some node reachable within a fixed number of steps near $u$. $P_n$ is the negative sampling distribution, and $Q$ is the number of negative samples. This objective can be replaced with any task-specific objective for other downstream tasks. This training method of sampling only some nodes makes GCN more generalizable, reduces the computational complexity of training, and improves the model performance.

A series of node-dropout training methods have been proposed after GraphSAGE. Chen et al. \cite{chen2018fastgcn} proposed FastGCN in 2018, whose idea is similar to GraphSAGE \cite{hamilton2017inductive}. The difference is that FastGCN randomly samples all nodes in the whole graph, not only the neighbors of a certain node. Considering node sampling efficiency, FastGCN is significantly faster than the original GCN as well as GraphSAGE, while maintaining comparable prediction performance.

Huang et al. \cite{huang2018adaptive} proposed AS-GCN in 2018. Again, the nodes are sampled during training; its differences with GraphSAGE \cite{hamilton2017inductive} and FastGCN \cite{chen2018fastgcn} are threefold: AS-GCN samples nodes layer by layer instead of independently; the sampler is adaptive; and AS-GCN skips some edges when transferring information between two nodes with long distances, enhancing the efficiency of information propagation. The experiments on running time show that AS-GCN is faster than the original GCN and the former node-wise sampling methods.

Zou et al. \cite{zou2019layer} proposed LADIES in 2019. The previous single-node-based sampling methods suffer from the problem of too many neighbors, while the layer-based sampling methods suffer from too sparse connections. LADIES also samples layer by layer, but it calculates the importance of each node in the next layer and samples the most critical nodes among them. This alleviate the problem of sparse connections while limiting the number of neighbors.

Feng et al. \cite{feng2020graph} proposed GRAND in 2020. It performs data augmentation using Random Propagation method. For the graph feature matrix, the authors compare two dropout methods: random dropout of matrix elements and random dropout matrix rows. The former is equivalent to a standard dropout in feature level, while the latter is dropping nodes. The latter performs better in experimental comparison.

\paragraph{\textbf{Drop Edges}}

Veli{\v{c}}kovi{\'c} et al. \cite{velivckovic2017graph} proposed GAT (Graph ATtention Networks) in 2017. GAT uses attention mechanism to compute the importance of different edges and train GNN using more important information. In 2021 Ye and Ji \cite{ye2021sparse} improved on GAT and proposed SGAT. SGAT learns sparse attention coefficients on the graph and produces an edge-sparsified graph. 

Rong et al. \cite{rong2019dropedge} proposed DropEdge in 2020. GCN training process is prone to overfitting and over-smoothing problems. Over-smoothing is a phenomenon that representations of all nodes on the graph tend to be the same, occurring when GCN is too deep. The authors address these two problems by randomly dropping edges during training: dropping edges can be seen as a data augmentation method introducing noise into input data to prevent overfitting; dropping edges also reduces the information propagation through edges, thus prevents the over-smoothing problem.

Dropout methods that drop graph information are widely used in GCN, and we summarize them in Table \ref{tab:drop embeddings}.

Dropping input information can be an effective way of data augmentation or regularization. As will be shown in \ref{sec:experiments}, it is a good way of augmenting sequences in recommendation. It can be applied to a wide range of scenarios as all machine learning tasks have input information. Meanwhile, it is not performed on model parameters but input data, so its regularization effectiveness is not as stable as dropping model structures.

% \begin{table*}[ht]
%     \renewcommand{\arraystretch}{1.1}
%     \centering
%     \newcolumntype{Y}{>{\raggedleft\arraybackslash}X}  %X单元格居左, Y单元格居右
% 	\newcolumntype{Z}{>{\centering\arraybackslash}X}  %Z单元格居中
%     \caption{Drop Graph Information}
%     \begin{tabularx}{0.7\textwidth}{XlXlX}
%         \toprule
%             Method & Year & Category & Application & Source \\
%         \midrule
%         GraphSAGE\cite{hamilton2017inductive}                & 2017 & Drop Graph Info    & GCN    & NeurIPS  \\
% FastGCN\cite{chen2018fastgcn}                  & 2018 & Drop Graph Info    & GCN    & ICLR     \\
% AS-GCN\cite{huang2018adaptive}                   & 2018 & Drop Graph Info    & GCN    & NeurIPS  \\
% LADIES\cite{zou2019layer}                   & 2019 & Drop Graph Info    & GCN    & NeurIPS  \\
% GRAND\cite{feng2020graph}                    & 2020 & Drop Graph Info    & GCN    & NeurIPS  \\
% GAT\cite{velivckovic2017graph} & 2018 & Drop Graph Info & GCN & ICLR \\
% SGAT\cite{ye2021sparse} & 2020 & Drop Graph Info & GCN & TKDE \\
% DropEdge\cite{rong2019dropedge}     & 2020 & Drop Graph Info    & GCN    & ICLR    \\
%         \bottomrule
%     \end{tabularx}
%     \label{tab:drop graph info}
% \end{table*}

\subsection{Summary and Interconnections between Dropout Methods}\label{subsec:interconnections}

Based on where in a machine learning task the dropout operation performs, we classify commonly used dropout methods into three major categories: dropping model structures, dropping embeddings and dropping input information. Dropping model structures is divided into two subcategories of dropping individual neurons and dropping neuron groups, according to the granularity of dropout operation. Dropping input information is divided into three subcategories of dropping one-dimensional information, two-dimensional information, and graph information according to the form of input information. 

% The training process of a neural network model can be abstracted as in Figure \ref{fig:discussion}. It starts from the input data, which may be converted into vector representations (embeddings); then comes the model; and finally, the output. In training phase, a batch of input data goes through this process in each epoch, and the model gives the corresponding output, after which the gradient is back-propagated, and model parameters and embeddings are updated. In testing phase, the model parameters and embeddings are fixed, and the output is obtained given the input data. This is the general training process for neural network models.
\begin{figure}
  \centering
  \includegraphics[width=\linewidth]{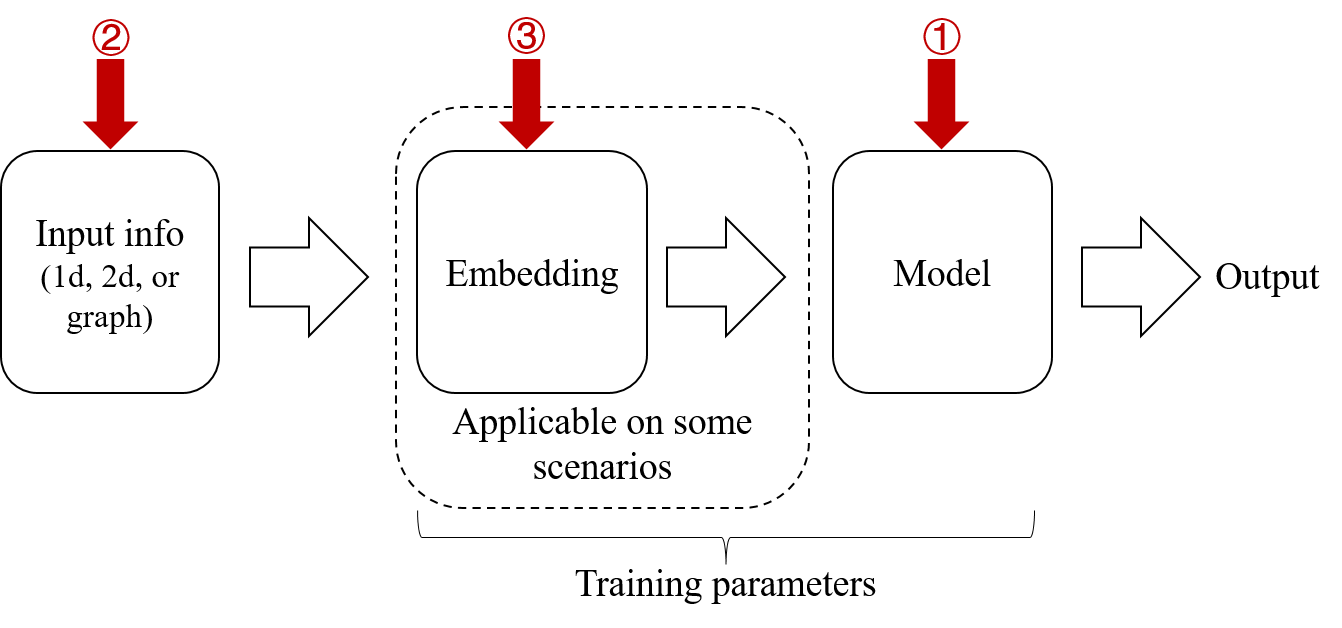}
  \vspace{-16pt}
  \caption{Training procedure and dropout position in neural models. We classify dropout methods into three major category based on where their operations perform in a training process: dropping input information (\ding{172}), dropping embeddings (\ding{173}), and dropping model structures (\ding{174}).}
  \label{fig:discussion}
  \vspace{-10pt}
\end{figure}

The three types of dropout methods, dropout of model structure, dropout of input information, and dropout of embedding, can be represented as \ding{172}, \ding{173}, and \ding{174} in Figure \ref{fig:discussion}, respectively. They are performed at three different stages of the training process, which is our classification criteria.

Within each block, there may also be different layers. For example, for the model part, if we use a deep network, we can perform dropout in every layer or just do it for some layers. If we use a recurrent neural network, we can perform dropout in its feed-forward direction, its recurrent direction, or part of the layers. Since a common implementation of dropout is to zero the neuron outputs, if the output of layer $l$ is considered as the input data of layer $l+1$, then dropping neurons of layer $l$ can be considered as dropping the input data of layer $l+1$. Thus dropping model structure and dropping input data are not clear-cut, they are highly correlated. This is more clear in computer vision tasks where convolutional neural networks are used: methods in Section \ref{para:drop cnn} and Section \ref{subsubsec:drop 2d info} operate at different levels, but their actual implementations are highly similar. By performing dropout operation in Section \ref{para:drop cnn} on the input image level, it becomes the operation in Section \ref{subsubsec:drop 2d info}.

% !TeX root = ../main.tex

\section{Contributions of Dropout Methods}\label{sec:discussion}

In Section \ref{sec:survey}, we classify commonly used dropout methods into three major categories according to the stage at which the dropout operation is performed, discuss their applications in neural models and analyze their interconnections. In this section, we discuss the contributions of dropout methods from the perspective of effectiveness and efficiency. 

\subsection{Improving Effectiveness}\label{subsec:improving_effectiveness}

Dropout makes models to better utilize training data and promotes model effectiveness in many ways.

\noindent $\bullet$ \textbf{Preventing overfitting.} Most of the dropout methods are used as regularization methods \cite{wager2013dropout, helmbold2015inductive} to prevent overfitting. For the methods of dropping model structure, during the training process, a part of neurons is dropped randomly to reduce the interdependence between neurons and prevent overfitting. The method of dropping input information enhances the robustness of the model by introducing noise into the input data. Meanwhile, many dropout methods have contributions other than preventing overfitting.

\noindent $\bullet$ \textbf{Simulating testing phase.} Some methods are used to simulate the possible situations in the testing phase \cite{volkovs2017dropoutnet, shi2018attention, shi2019adaptive}. The testing phase may face an information deficit, and the model needs to give predictions under the absence of information. Therefore, these methods drop part of the information at training time so that the model does not rely excessively on this possibly missing information and improves the performance of the testing phase.

\noindent $\bullet$ \textbf{Data augmentation.} Some methods are used for data augmentation \cite{bouthillier2015dropout, devries2017improved, zhong2020random, chen2020gridmask, yun2019cutmix}. Noise is introduced into the input data to create more training samples and improve the training effect of the model. Dropping training data can be seen as introducing Bernoulli noise into data.

\noindent $\bullet$ \textbf{Enhancing data representation.} Some methods are used to enhance data representation in pre-training phase \cite{sun2019ernie, wu2019mask, zhang2020pegasus, zhou2020s3}. These BERT \cite{devlin2019bert} based methods randomly mask part of the input data and use the unmasked part to predict the masked part to enhance data representation.

\noindent $\bullet$ \textbf{Preventing over-smoothing.} Dropout methods in graph neural networks can also solve the over-smoothing problem \cite{rong2019dropedge}. Over-smoothing occurs when GCN is too deep that the representation of all nodes on the graph tends to be the same. Randomly dropping edges during training can reduce information propagation through edges and prevent the over-smoothing problem.

\subsection{Improving Efficiency}\label{subsec:improving_efficiency}

Besides improving model effectiveness, some dropout methods can also improve model efficiency for certain tasks. 

\noindent $\bullet$ \textbf{Accelerating GCN training.} In GCN scenarios, node sampling technique proposed by GraphSAGE \cite{hamilton2017inductive} efficiently accelerates GCN training. It only needs some of the nodes to perform the training process instead of needing all node neighbors. Later works like FastGCN\cite{chen2018fastgcn} and AS-GCN \cite{huang2018adaptive} improve this sampling technique, making it faster or sample in an adaptive way. 

\noindent $\bullet$ \textbf{Model compression.} Some methods are used for model compression \cite{molchanov2017variational, neklyudov2017structured, gomez2018targeted, gomez2019learning, salehinejad2019ising, salehinejad2021edropout}. These methods make the model structure easier to compress after random dropout of neurons, e.g., easier to perform neural pruning. Model compression reduces model parameters, which can improve training efficiency and prevent overfitting.

\noindent $\bullet$ \textbf{Model uncertainty measurement.} Some methods are used to measure the model uncertainty \cite{gal2016dropout, gal2016uncertainty, gal2017concrete, lakshminarayanan2017simple}. These methods view dropout as a Bayesian learning process. For example, in Monte Carlo Dropout \cite{gal2016dropout}, the authors interpret dropout as a Bayesian approximation of a deep Gaussian process.
% The output of a deep Gaussian process is a probability distribution, and some properties of this potential distribution are estimated in the testing phase using standard dropout. For example, the estimated variance can be used to characterize the uncertainty of the model output. 
Monte Carlo Dropout estimates the uncertainty of the model output by performing a grid search on the dropout rate, which is almost unusable for deeper models (those in computer vision tasks) and reinforcement learning models because of the excessive computational time and computational resources consumed. Thanks to the development of Bayesian learning, Concrete Dropout \cite{gal2017concrete} uses a continuous relaxation of the dropout discrete mask. A new objective function is proposed to automatically adjust the dropout rate on large models, reducing the time required for experiments. It also allows the agent in reinforcement learning to dynamically adjust its uncertainty as the training process proceeds and more training data is observed.

% !TeX root = ../main.tex

\section{Dropout Experiments in Recommendation Models}\label{sec:experiments}

We have reviewed multiple types of dropout methods and discussed their interconnections and contributions. However, each work has its own experiments to verify the effect of its dropout method, so the methods' effectiveness actually has not been investigated under a unified framework and evaluation system. In this section, we experimentally investigate four classes of dropout methods on recommendation models. Choosing recommendation models as our experiment scenario is because they utilize various heterogeneous information, which are transformed into different forms, from input data, to embeddings, and to model structures, covering the range of dropout operations we reviewed in Section \ref{sec:survey}. Such a variety of information sources and forms provides a suitable environment for our comparisons and verification of different dropout methods.

We first introduce the selected recommendation models, the implementations of the four dropout methods on each of them (Section \ref{subsec:implementations}), the datasets and experimental settings (Section \ref{subsec:exp setup}). Then, we analyze the experimental results and present comparison to evaluate the effectiveness of each dropout method (Section \ref{subsec:results_and_analysis}). Finally, we explore the effect of dropout ratio on model performances (Section \ref{subsec:effect_of_dropout_ratio}).

\subsection{Recommendation Models and Implementations of Dropout Methods}\label{subsec:implementations}

We choose five recommendation models belonging to four classes:
\begin{itemize}
    \vspace{-4pt}
    \item Traditional recommendation model: BPRMF\cite{rendle2009bpr}
    \item Neural recommendation model utilizing context information: NFM\cite{he2017neural}
    \item Sequential recommendation model: GRU4Rec\cite{hidasi2015session} and SASRec\cite{kang2018self}
    \item Graph recommendation model: LightGCN\cite{he2020lightgcn}
    \vspace{-4pt}
\end{itemize}

The four dropout methods are dropout of model structure, dropout of input information, dropout of embeddings, and dropout of graph information. Since there are significant differences between graph information and other input information in recommender systems, we treat them as different methods. We elaborate on the implementations of the four dropout methods on each recommendation model in Appendix \ref{append:implementation}.
% \begin{table*}[ht]
%     % \renewcommand{\arraystretch}{1.1}
%     \centering
%     \newcolumntype{Y}{>{\raggedleft\arraybackslash}X}  %X单元格居左, Y单元格居右
% 	\newcolumntype{Z}{>{\centering\arraybackslash}X}  %Z单元格居中
%     \caption{Implementations of dropout methods.}
%     \begin{tabular}{lllll}
%         \toprule
%              & Drop model structure & Drop input info & Drop embedding & Drop graph info \\
%         \midrule
% BPRMF\cite{rendle2009bpr} &  &  &  &  \\
% NFM\cite{he2017neural} &  &  &  &   \\
% GRU4Rec\cite{hidasi2015session} &  &  &  &  \\
% SASRec\cite{kang2018self} &  &   &   &  \\
% LightGCN\cite{he2020lightgcn} &  &   &  &  \\
%         \bottomrule
%     \end{tabular}
%     \label{tab:implementations_of_dropout_methods}
% \end{table*}

\subsection{Datasets and Experiment Settings}\label{subsec:exp setup}

\begin{table}[ht]
    \vspace{-4pt}
    \centering
    \caption{Dataset Statistics}
    \vspace{-4pt}
    \begin{tabular}{lllll}
        \toprule
             & \#user  & \#item  & \#interaction     & density(\%) \\
        \midrule
        MovieLens-1M & 6040 & 3883 & 1000209 & 4.26  \\
        ml1m-cold & 6040 & 3883 & 797675 & 3.40  \\
        Amazon Baby 5-core & 19445 & 7050 & 160792 & 0.117  \\
        \bottomrule
    \end{tabular}
    \label{tab:dataset_attributes}
    \vspace{-2pt}
\end{table}

We use two datasets for experiments: MovieLens-1M-cold and Amazon Baby 5-core 
% \footnote{\url{http://jmcauley.ucsd.edu/data/amazon/index_2014.html}} 
\cite{mcauley2015image, he2016ups}.

MovieLens-1M-cold is obtained by artificially creating a cold-start condition based on MovieLens-1M
% \footnote{\url{https://grouplens.org/datasets/movielens/1m/}} 
\cite{harper2015movielens}. Each user in MovieLens-1M has at least 20 interactions, so there is no cold-start scenario for users. To make our experimental environment closer to real-world recommendation, we randomly select 10\% of users and 10\% of items and remove all of their interactions in training set, constructing a group of cold-start users and items. We also randomly select user and item attributes and set the value to zero (unknown) so that unknown attribute values in the dataset account for 10\% of all attribute values.
% \begin{itemize}
%     \item[-] Randomly select 10\% of users and remove all interaction history from their training set to construct a group of cold-start users;
%     \item[-] Randomly select 10\% of items and remove all interaction history from their training set to construct a group of cold-start items;
%     \item[-] Randomly select user and item attributes and set the value to zero (unknown) so that unknown attribute values in the dataset account for 10\% of all attribute values.
% \end{itemize}
After this process, the resulting dataset is named MovieLens-1M-cold, or ml1m-cold for short.

The statistics of the datasets are shown in Table \ref{tab:dataset_attributes}. The density of Amazon Baby 5-core is only 1/30 of ml1m-cold. We choose a dense and a sparse dataset to make our experiment results more general.

We consider Top-N recommendation task. The parameter values taken for all models in common are shown in Table \ref{tab:ExperimentSettings} in Appendix \ref{append:model_param}.
This ensures a consistent evaluation environment and comparability of evaluation results. According to the analysis of Krichene et al. \cite{krichene2020sampled}, negative sampling during testing can bias the results. Therefore, we conduct non-sampling evaluation for all our experiments. 
%Note that $\mathcal{U}$ is the set of users, $\mathcal{I}$ is the set of items, and for a positive user-item pair $<\mathcal{U}_i, \mathcal{I}_j>$ in the test set, we treat all $<\mathcal{U}_i, \mathcal{I}_k>, k\neq j$ as negative pairs. That is, in this experiment, for ml1m-cold we evaluate a ratio of $1:3882$ for positive and negative instances, and for Amazon Baby 5-core a ratio of $1:7049$.
Parameters specific to each model are shown in Table \ref{tab:ExperimentSettings2} in Appendix \ref{append:model_param}. The training batch size is set to 1024 on ml1m-cold for faster training.

% All parameters are taken to be as consistent as possible with the original paper.

We searched $L_2$-coefficient and choose $1e^{-6}$ for all models on ml1m-cold dataset. On Amazon Baby 5-core, we choose $1e^{-5}$ for BPR, NFM, and GRU4Rec; $1e^{-6}$ for SASRec; and $1e^{-4}$ for LightGCN.
% as shown in Table \ref{tab:l2}.

% \begin{table}[ht]
% \vspace{-4pt}
% % \renewcommand{\arraystretch}{1.1}
% \centering
% \caption{$L_2$ values}
% \vspace{-4pt}
% \begin{tabular}{ccc}
% \toprule
%      & ml1m-cold & Amazon Baby 5-core \\
% \midrule
% BPR      & $1\times 10^{-6}$   & $1\times 10^{-5}$            \\
% NFM      & $1\times 10^{-6}$   & $1\times 10^{-5}$            \\
% GRU4Rec  & $1\times 10^{-6}$   & $1\times 10^{-5}$            \\
% SASRec   & $1\times 10^{-6}$   & $1\times 10^{-6}$            \\
% LightGCN & $1\times 10^{-6}$   & $1\times 10^{-4}$           \\
% \bottomrule
% \end{tabular}
% \label{tab:l2}
% \vspace{-2pt}
% \end{table}

% \subsection{Results and Analysis}\label{subsec:analysis}

% This section will analyze the effect of various dropout methods on each of the five recommendation models. We present experiment results, analyze the effect of dropout ratio, and discuss the proper dropout methods for recommendation models and the appropriate applications for dropout methods. 

\subsection{Results and Analysis} \label{subsec:results_and_analysis}

\begin{figure}
    \centering
    \includegraphics[width=0.9\linewidth]{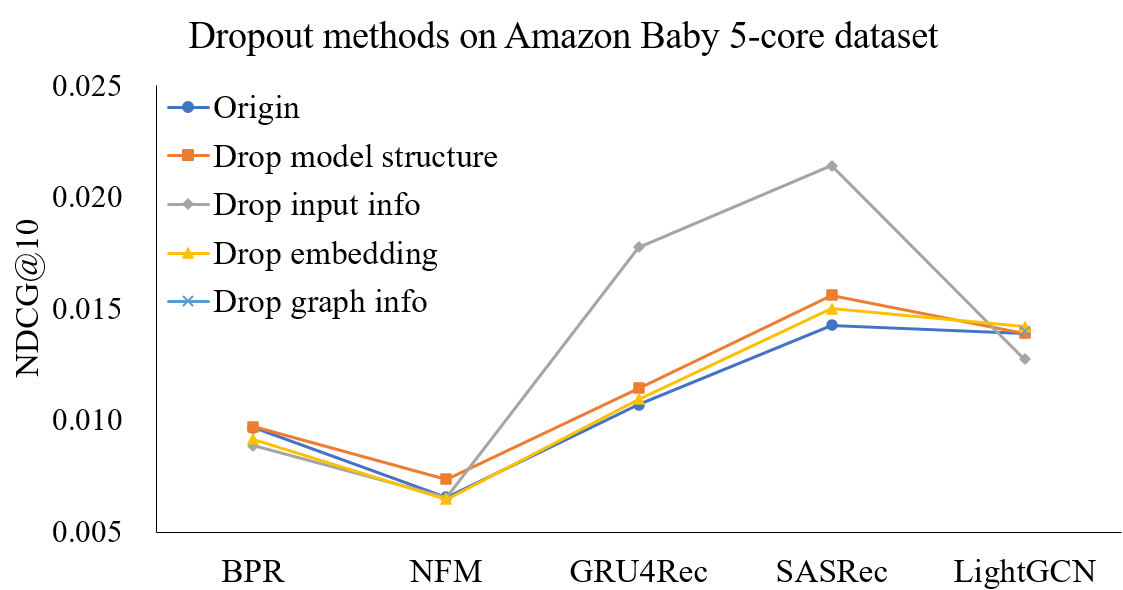}
    \vspace{-6pt}
    \caption{NDCG@10 on Amazon Baby 5-core}
    \label{fig:Baby_results}
    \vspace{-10pt}
\end{figure}

\subsubsection{Overall Results} \label{subsubsec:overall_results}

\begin{table}[ht]
\vspace{-2pt}
\centering
\caption{Overall NDCG@10 results}
\vspace{-4pt}
\begin{threeparttable}
\begin{tabular}{clllll}
\toprule
 & Origin & \makecell[c]{Drop Model\\Structure} & \makecell[c]{Drop Input\\Info} & \makecell[c]{Drop\\Embedding}  \\
\midrule
\multicolumn{5}{c}{Amazon Baby 5-core}\\
 \multicolumn{1}{l|}{BPR}      & 0.00969 & 0.00973 & 0.00888**  & 0.00916*  \\
 \multicolumn{1}{l|}{NFM}      & 0.00657 & 0.00737** & 0.00654  & 0.00647 \\
 \multicolumn{1}{l|}{GRU4Rec}  & 0.01071 & 0.01146 & 0.01777**  & 0.01096 \\
 \multicolumn{1}{l|}{SASRec}   & 0.01428 & 0.01562* & 0.02143**  & 0.01502 \\
 \multicolumn{1}{l|}{LightGCN\dag} & 0.01392 & 0.01392 & 0.01275  & 0.01420 \\ \hline \\
% \rule{5pt}{9pt}
\multicolumn{5}{c}{ml1m-cold}\\
 \multicolumn{1}{l|}{BPR}      & 0.0339  & 0.0364**  & 0.0331  & 0.0332 \\
 \multicolumn{1}{l|}{NFM}      & 0.0335  & 0.0353*  & 0.0334  & 0.0354** \\
 \multicolumn{1}{l|}{GRU4Rec}  & 0.0964  & 0.1010**  & 0.1084**  & 0.1013** \\
 \multicolumn{1}{l|}{SASRec}   & 0.1064  & 0.1092**  & 0.1093 & 0.1063 \\
 \multicolumn{1}{l|}{LightGCN\dag} & 0.0377  & 0.0388  & 0.0361** & 0.0376 \\
\bottomrule
\end{tabular}
\begin{tablenotes}
\footnotesize
\item[]*: $p < 0.05$, **: $p < 0.01$, compared to the origin value (not using any dropout methods)
\item[]\dag: Drop graph info for LightGCN on Amazon Baby 5-core is 0.01403, on ml1m-cold is 0.0383
\end{tablenotes}
\end{threeparttable}
\label{tab:results_all}
\vspace{-2pt}
\end{table}

\begin{figure}
    \centering
    \includegraphics[width=0.9\linewidth]{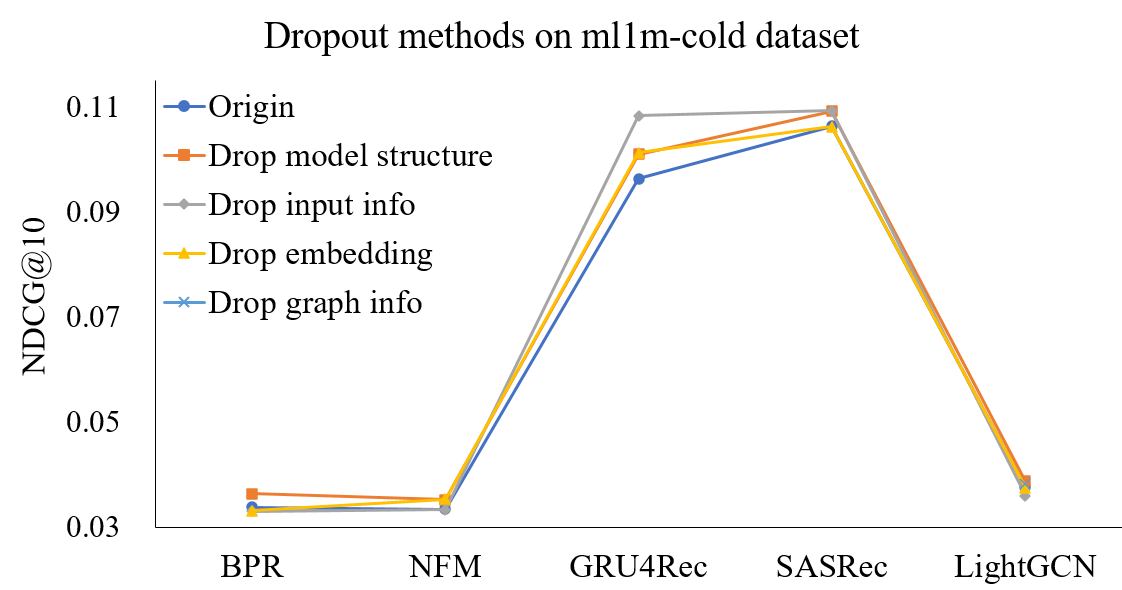}
    \vspace{-6pt}
    \caption{NDCG@10 on ml1m-cold}
    \label{fig:ml1m-cold_results}
    \vspace{-10pt}
\end{figure}

We present the overall results of NDCG@10 in Table \ref{tab:results_all}, and each value in the table is the best result for one dropout method on one model with the dropout ratio among $\{0.1, 0.2, 0.3\}$. We plot the results in lines showing in Figure \ref{fig:Baby_results} and \ref{fig:ml1m-cold_results}. All detailed experimental results of other evaluation metrics (NDCG@5, 20, 50; HR@10, 20) and each dropout ratio are in Appendix \ref{append:exp data}.

According to the experimental results, we summarize the effect of different dropout methods in the five recommendation models.

For traditional recommendation models that do not use neural networks, dropping model structure can have a regularizing effect and improve model performance. Dropping input information and dropping embeddings can be detrimental to the performance of the model.

For neural network models using contextual information, both dropping model structure and dropping embeddings can improve the model performance. This can be because the former acts as a regularizer, and the latter allows the model to take advantage of multiple aspects of information to better cope with cold start situations. Dropping input information does not affect model effectiveness.

For sequential models, all three dropout methods lead to improved model effects. Among them, dropping input information has the most significant improvement if dropout ratio is properly chosen. This is because dropping items in input sequences can be viewed as a type of sequence augmentation \cite{liu2021augmenting}. Dropping model structure has the most stable improvement. Dropping embeddings also allows the model to get improved or, at least, remain unchanged.

For the graph recommendation model, this paper only explores the lightweight model LightGCN, which contains a small quantity of parameters thus does not require too much regularization according to the original paper. So all the four dropout methods do not affect the model performance, or even have a detrimental effect.

\subsubsection{Discussion on the Applications of the Four Dropout Methods}\label{subsubsec:conclude 2}

As described in Section \ref{sec:survey}, the four dropout methods belong to different levels and may have different contributions, and Section \ref{subsubsec:overall_results} shows their performances vary in different scenarios. Therefore, it is not feasible to simply determine which of them is better or worse generally, but we can analyze their features. This section provides some analysis and discussions on the properties of the four dropout methods.

Dropping model structure is the most stable dropout method in our experiment. It makes the model performance gain or at least unchanged for all models, where most of them have significant improvement. This reveals that this classical dropout method is still the most effective way of regularization, and dropout according to the structural properties of the model can achieve good results, except for the model with very few parameters (like LightGCN) which does not need much regularization.

Dropping input information has significant side effects on the performance of traditional models and graph recommendation models, and has no effect on the neural model utilizing contextual information. This indicates that recommendation models utilizing less information and having fewer parameters, such as BPRMF and LightGCN in this paper, do not need too much regularization, and dropping input information will harm their performances. However, it significantly improves sequential recommendation models and is the most effective one among the four methods. This is because dropping input information of sequential models can be viewed as a way of sequence augmentation.

\begin{figure}
    \centering
    \includegraphics[width=\linewidth]{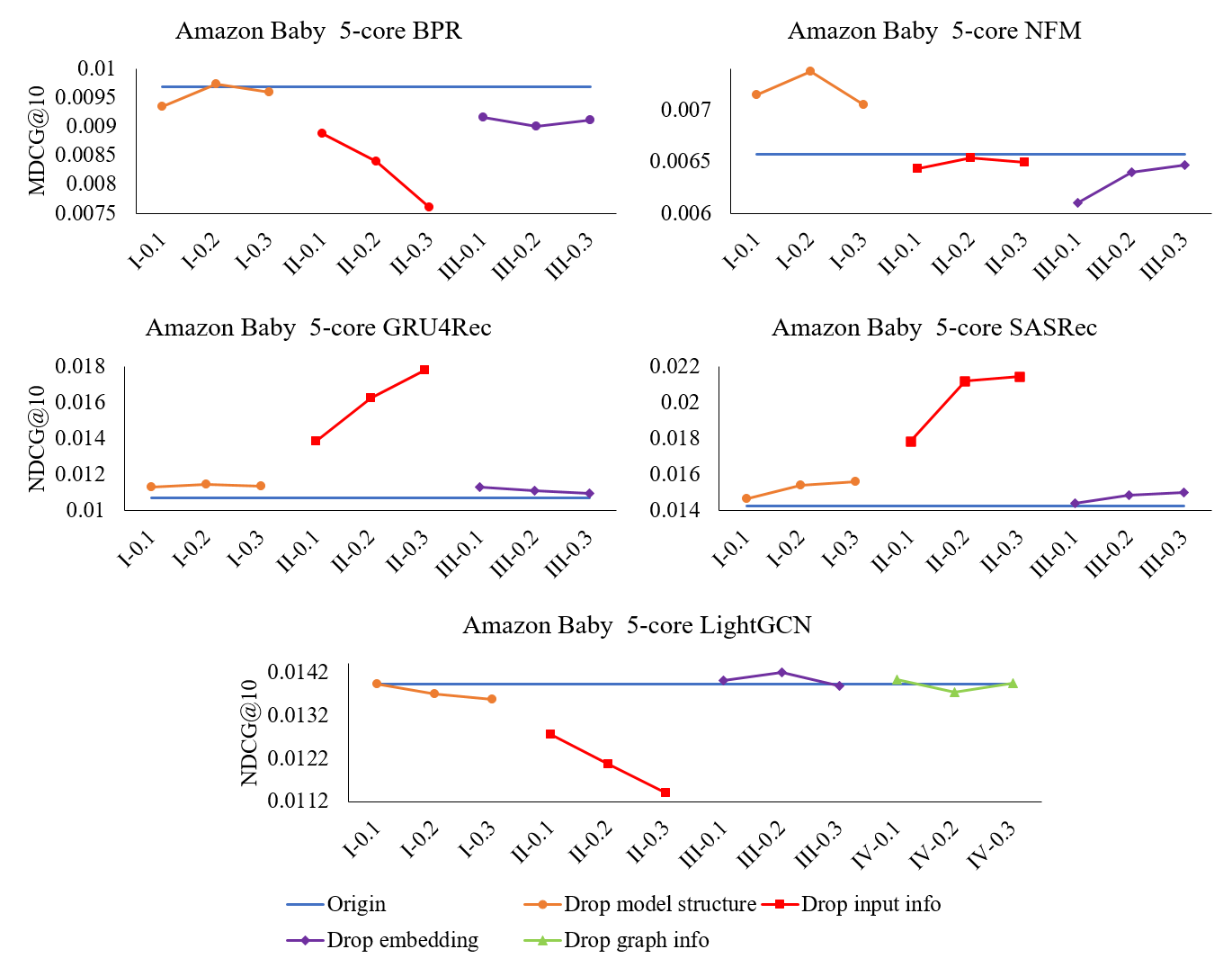}
    \vspace{-14pt}
    \caption{Effect of dropout ratio on Amazon Baby 5-core}
    \label{fig:drop_rate_Baby}
    \vspace{-10pt}
\end{figure}

Dropping embedding has no significant effect on model performance in most cases, while slightly improves NFM. The method of embedding dropout is from \cite{shi2018attention, shi2019adaptive}, which use the attention mechanism to make the model select the information to be exploited automatically. Then the model can automatically rely on the information that is kept after dropping part of the embeddings. In contrast, the model used in this experiment does not have this attention mechanism, so the improving effect is limited.

For dropping edges in graph, there is neither an improvement nor a decrease on model performance. This is determined by the nature of LightGCN \cite{he2020lightgcn}, which only has a small quantity of parameters. Dropping edges or nodes may be effective to other models with a larger number of parameters such as NGCF \cite{wang2019neural}.

\subsection{Effect of Dropout Ratio}\label{subsec:effect_of_dropout_ratio}

\begin{figure}
    \centering
    \includegraphics[width=\linewidth]{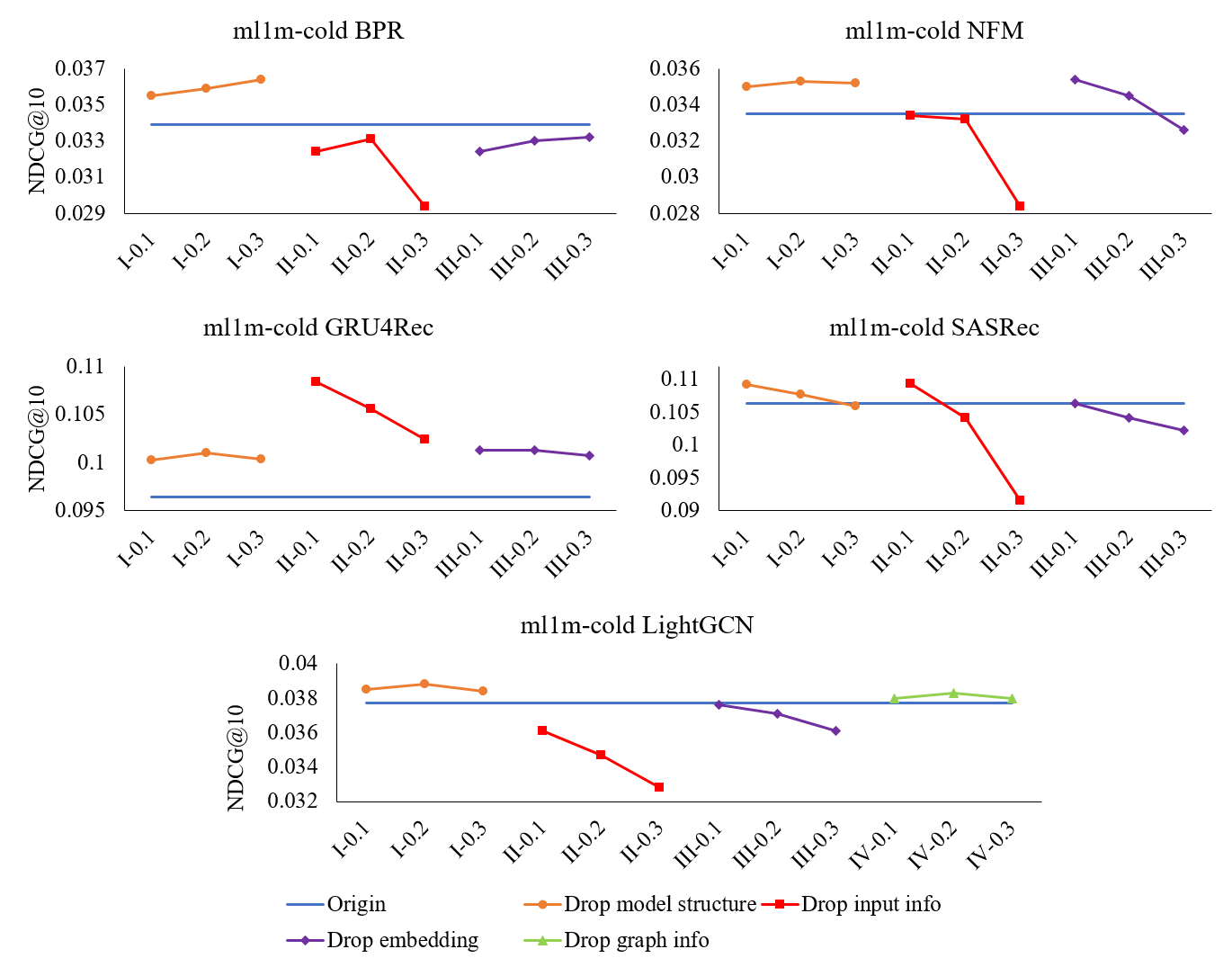}
    \vspace{-14pt}
    \caption{Effect of dropout ratio on ml1m-cold}
    \label{fig:drop_rate_ml1m-cold}
    \vspace{-10pt}
\end{figure}

In this section, we analyze the parameter sensitivity: how does dropout ratio affect model performances?

For each dataset, each model, and each evaluation metric, we test dropout ratios of $\{0.1, 0.2, 0.3\}$. We plotted the values of NDCG@10 in Figure \ref{fig:drop_rate_Baby} and \ref{fig:drop_rate_ml1m-cold}.

As can be seen in Figure \ref{fig:drop_rate_Baby} and \ref{fig:drop_rate_ml1m-cold}, 
% the performance of each recommendation model and each dropout method is relatively consistent for both datasets.
the orange line keeps staying above the blue line, indicating that dropping model structure almost always leads to a stable improvement on model performances. For the two sequential recommendation models, GRU4Rec and SASRec, the red line is higher than the other three colored lines, indicating that dropping input information has the most significant improvement effect on sequential recommendation models, when choosing dropout ratio properly. The appropriate range of dropout rate for SASRec on ml1m-cold is low, and 0.2 and 0.3 are too large, causing its performance to decline. The purple line improves NFM on ml1m-cold because ml1m-cold contains rich contextual information, and dropping embeddings allows NFM to utilize multifaceted information more robustly.

% \subsection{Summary}

% Section \ref{sec:experiments} presented dropout experiments in recommendation models under an unified evaluation experiment framework. We analyze and compare experiment results comprehensively, providing a reference for later works in this area.

% !TeX root = ../main.tex

\section{Future Directions}\label{sec:discussion2}

Based on our review and experiments, in this section, we further discuss several topics about dropout and analyzes some potential research directions in this field.

\subsection{Transfer of Dropout Strategies in Different Domains}
Most dropout methods that drop input information are domain specific, designed for a certain domain like NLP or CV, as we have reviewed in Section \ref{subsec:drop inputs}. However recently, MAE \cite{he2021masked} brought the randomly masking strategy from NLP pre-training to CV and achieved good results. Some dropout methods dropping model structure are also linked to the ideas of other domain. SimCSE \cite{gao2021simcse} adopted dropout in NLP self-supervised learning tasks, in a way like the data augmentation techniques in self-supervised learning in CV. This indicates that the applications of dropout methods in different domain could be embarking on similar trajectories.

\subsection{Selection of Dropout Methods}
% \noindent $\bullet$ \textbf{Selection of dropout methods.} 
Though specific dropout methods achieve impressive improvement on certain neural models, deciding which type of dropout method to use is not an easy task. Based on our experimental results in Section \ref{sec:experiments}, distinct dropout methods work best for various types of models, like dropping input information for sequential recommendation models, dropping embedding for recommendation models that utilize contextual information, and dropping model structure for all models. But can we determine which dropout method should be used according to the characteristic of the model before we conduct experiments? If so, lots of time on enumerating dropout methods during training could be saved. 

\subsection{Hyperparameter Optimization of Dropout}
% \noindent $\bullet$ \textbf{Hyperparameter optimization of dropout.} 
In the beginning, dropout methods require manually setting dropout ratio and patterns. Standard Dropout \cite{hinton2012improving} and DropConnect \cite{wan2013regularization} need manually setting dropout ratio, and the dropout patterns of the methods like DropBlock \cite{ghiasi2018dropblock} and GridMask \cite{chen2020gridmask} need to be deliberately designed.
Some later methods make progress to automate this process to some extent. As more data is exposed throughout the training process, Variational Dropout \cite{kingma2015variational}, Concrete Dropout \cite{gal2017concrete}, and Curriculum Dropout \cite{morerio2017curriculum} can automatically adjust dropout ratio towards more suitable values. Instead of randomly dropping neurons, Targeted Dropout \cite{gomez2018targeted} and Ising-Dropout \cite{salehinejad2019ising} calculate the most suitable neurons to drop, making the dropout pattern design more automatic.
% Weighted Channel Dropout \cite{hou2019weighted} and CorrDrop \cite{zeng2021correlation} can also decide the dropout pattern based on their calculation.
AutoDropout \cite{pham2021autodropout} uses reinforcement learning to train a controller, which decides the dropout patterns in CNN and Transformers. In the future, more efficient ways of optimizing dropout hyperparameters could be explored. 

\subsection{Efficient Dropout}
% \noindent $\bullet$ \textbf{Efficient dropout.} 
Besides effectiveness, efficiency is also needed to be considered when using dropout, since the dropout operation itself takes time. Some aforementioned methods in Section \ref{sec:survey} add additional attention parts to calculate dropout patterns, which may prolong the training time; edge dropping operation in Section \ref{sec:experiments} slows down the training; and the methods that use reinforcement learning \cite{pham2021autodropout} requires excessive computational time and resources.

Fast Dropout \cite{wang2013fast} takes the first step of improving the efficiency of dropout operation itself, and many methods have tried to improve dropout efficiency: Concrete Dropout optimizes the model uncertainty estimation process of Monte Carlo Dropout; the series of GCN-based dropout methods \cite{hamilton2017inductive, huang2018adaptive, feng2020graph} have been making improvement on node-dropping training. However, as more complicated and time-consuming techniques like reinforcement learning have been adopted for dropout, improving their efficiency to speed up the dropout operation is still worth to be further explored.
% Since now more advanced techniques like reinforcement learning have been adopted for dropout,

\subsection{Understanding Dropout Theoretically}
% \noindent $\bullet$ \textbf{Understanding dropout theoretically.} 
The effectiveness of dropout has been irrefutably verified by experiments of hundreds of works. However, mathematical proofs of the validity of dropout has been rare. Baldi and Sadowski in 2013 gave a mathematical formality of dropout and use it to analyze the averaging and regularization properties of dropout \cite{baldi2013understanding}. Gal and Ghahramani in 2016 cast dropout network training as approximate inference of Bayesian neural networks, achieving a significant improvement in experiment results without increasing time complexity \cite{gal2015bayesian}. Gal and Ghahramani's other works \cite{gal2016dropout, gal2016theoretically} also perform mathematical derivation on the validity of proposed dropout methods. In the future, proving the effectiveness of dropout not only from intuitive explanation and experimental verification but also from mathematical proof could be a challenging research direction.

% Monte Carlo Dropout\cite{gal2016dropout}
% Variational RNN Dropout\cite{gal2016theoretically}

% \subsection{Effect of Other Regularization Methods on Dropout}
% $L_2$ regularization 

% \subsection{Unified Evaluations under Different Scenarios}
% The analysis of broader applicability of dropout is also needed. We evaluate the effectiveness of dropout methods in recommendation scenarios, but how well does each one of aforementioned three types of dropout methods work in other scenarios? In the future, it is important for unified evaluations under other scenarios such as natural language processing, computer vision, and graph neural networks, to analyze the general applicability of different types of dropout methods. 

% !TeX root = ../main.tex

\section{Conclusions}\label{sec:conclusion}

In this paper, we investigate more than seventy dropout methods in neural network models and classify them into three major categories and six subcategories according to the stage where the dropout operation is performed. We discuss their applications in neural models, their contributions and interconnections.

We conduct experiments on five recommendation models to verify the effectiveness of each type of dropout method under our classification framework, and find that dropping model structure has the most general and stable improvement effect on the models, while dropping input information and dropping embeddings are model-specific.

Finally, we present some open problems and potential research directions, hoping to promote the research in this field. Dropout methods are basic and universally used training techniques in today's neural model, helping with our steps towards better machine learning and artificial intelligence. We hope this survey paper can help readers better understand the works in this research area. 

% In the future, we will further improve our classification framework. We will test the effects of various dropout methods on more datasets and models to make our conclusions more representative. Moreover, we will also conduct experiments in scenarios other than recommendation systems to generalize our conclusions.

% use section* for acknowledgment
\ifCLASSOPTIONcompsoc
  % The Computer Society usually uses the plural form
  \section*{Acknowledgments}
\else
  % regular IEEE prefers the singular form
  \section*{Acknowledgment}
\fi
% The authors would like to thank...
This work is supported by the National Key Research and Development Program of China (2018YFC0831900), Natural Science Foundation of China (Grant No. 62002191, 61672311, 61532011) and Tsinghua University Guoqiang Research Institute.

% Can use something like this to put references on a page
% by themselves when using endfloat and the captionsoff option.
\ifCLASSOPTIONcaptionsoff
  \newpage
\fi

% trigger a \newpage just before the given reference
% number - used to balance the columns on the last page
% adjust value as needed - may need to be readjusted if
% the document is modified later
%\IEEEtriggeratref{8}
% The "triggered" command can be changed if desired:
%\IEEEtriggercmd{\enlargethispage{-5in}}

% references section

% can use a bibliography generated by BibTeX as a .bbl file
% BibTeX documentation can be easily obtained at:
% http://mirror.ctan.org/biblio/bibtex/contrib/doc/
% The IEEEtran BibTeX style support page is at:
% http://www.michaelshell.org/tex/ieeetran/bibtex/
\bibliographystyle{IEEEtran}
% argument is your BibTeX string definitions and bibliography database(s)
% \bibliography{ref/refs}
\bibliography{main.bbl}
%
% <OR> manually copy in the resultant .bbl file
% set second argument of \begin to the number of references
% (used to reserve space for the reference number labels box)
% \begin{thebibliography}{109}

% \bibitem{IEEEhowto:kopka}
% H.~Kopka and P.~W. Daly, \emph{A Guide to \LaTeX}, 3rd~ed.\hskip 1em plus
%   0.5em minus 0.4em\relax Harlow, England: Addison-Wesley, 1999.

% \end{thebibliography}

% biography section
% 
% If you have an EPS/PDF photo (graphicx package needed) extra braces are
% needed around the contents of the optional argument to biography to prevent
% the LaTeX parser from getting confused when it sees the complicated
% \includegraphics command within an optional argument. (You could create
% your own custom macro containing the \includegraphics command to make things
% simpler here.)
% \begin{IEEEbiography}[{\includegraphics[width=1in,height=1.25in,clip,keepaspectratio]{mshell}}]{Michael Shell}
% Biography text here.
% \end{IEEEbiography}
% or if you just want to reserve a space for a photo:
\begin{IEEEbiography}[{\includegraphics[width=1in,height=1.25in,clip,keepaspectratio]{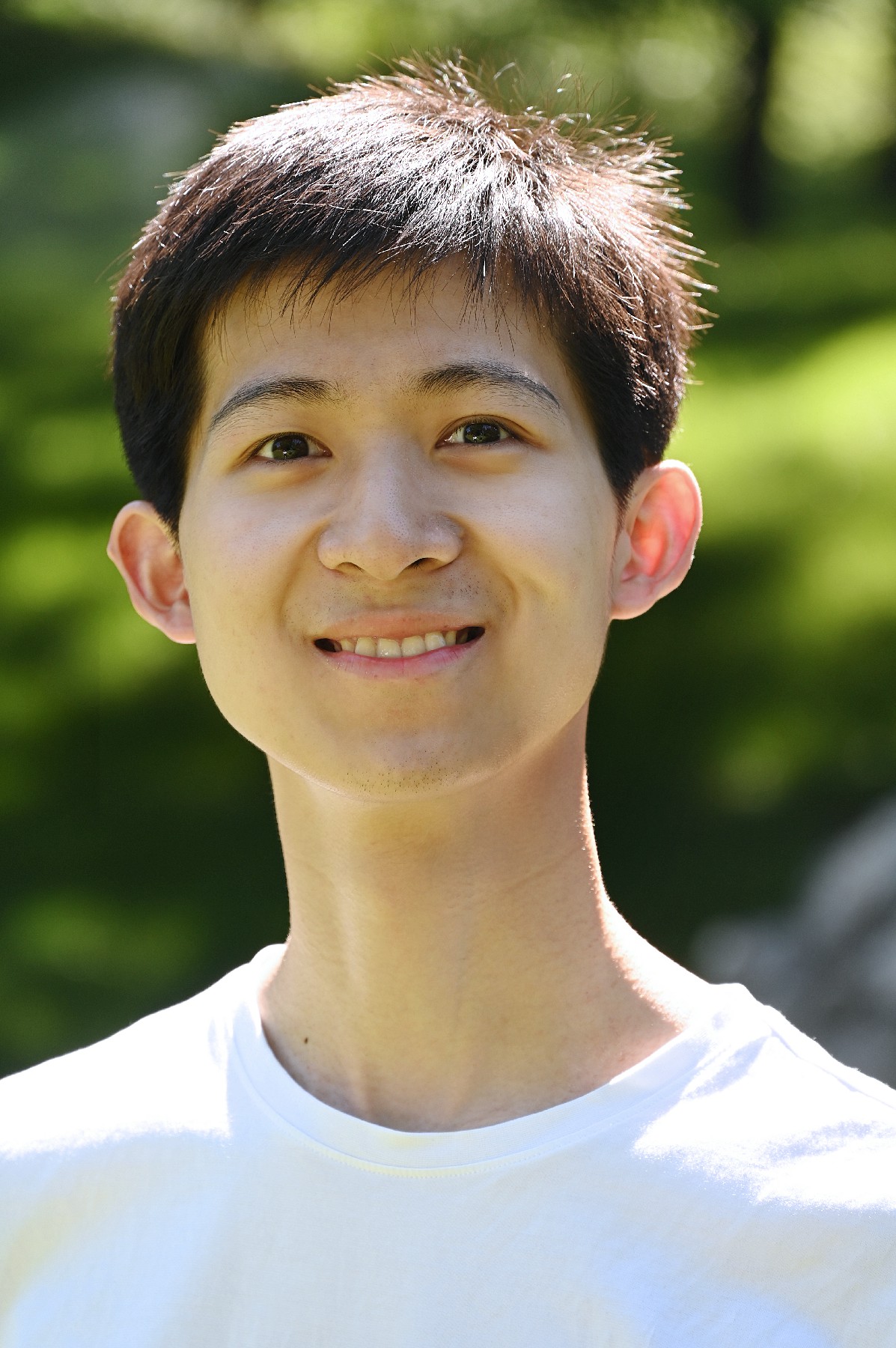}}] {Yangkun Li} is currently pursuing a Ph.D. degree in Department of Computer Science and Technology at Tsinghua University. He received his B.E. from Department of Computer Science and Technology, Tsinghua University in 2021. His research interests focus on personalized recommender systems, user modeling, and data mining.
\vspace{-10pt}
\end{IEEEbiography}
\begin{IEEEbiography}[{\includegraphics[width=1in,height=1.25in,clip,keepaspectratio]{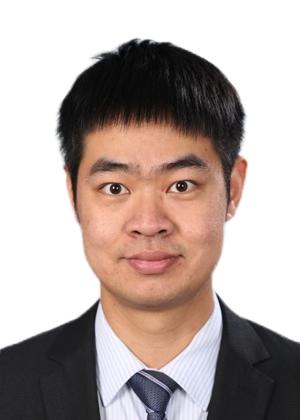}}]{Weizhi Ma} is currently a research assistant professor at the Institute for AI Industry Research (AIR), Tsinghua University. He received his Ph.D. from Tsinghua University in 2019. His major research interests are in recommender systems, user modeling, and information retrieval. He has served as PC member for top conferences including KDD, WSDM, AAAI, theWebConf, SIGIR, EMNLP, COLING, and CIKM, and as reviewer for journals including TKDE, TOIS, and TNNLS.
\vspace{-10pt}
\end{IEEEbiography}
\begin{IEEEbiography}[{\includegraphics[width=1in,height=1.25in,clip,keepaspectratio]{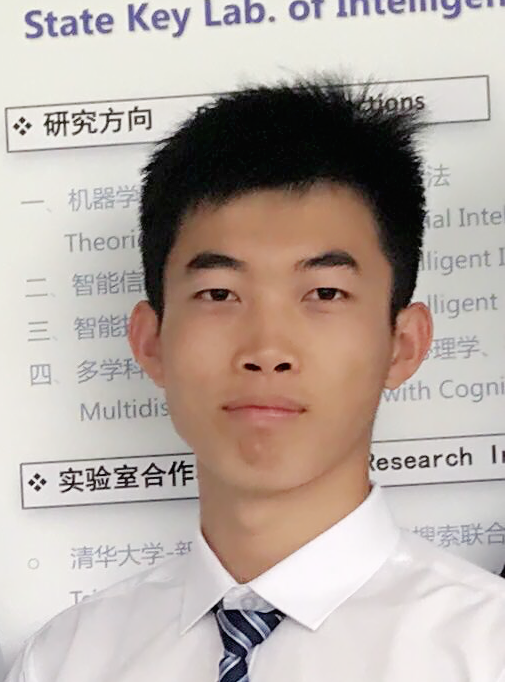}}]{Chong Chen} is now a final year Ph.D. student in Dept. of Computer Science and Technology in Tsinghua University. He received his B.E. from Tsinghua University in 2017. He has over 10 publications appeared in top conferences and journals including AAAI, WSDM, theWebConf, SIGIR, and TOIS. His research interests include deep learning, user modeling, and recommendation.
\vspace{-10pt}
\end{IEEEbiography}
\begin{IEEEbiography}[{\includegraphics[width=1in,height=1.25in,clip,keepaspectratio]{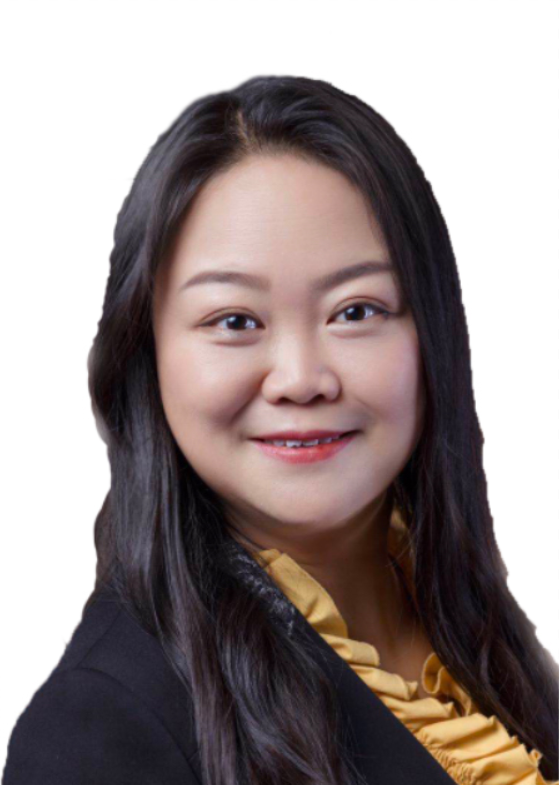}}]{Min Zhang} is currently an associate professor of Dept. of Computer Science and Technology, Tsinghua University. She received her Ph.D. in Tsinghua University in 2003. Her research interests include web information retrieval and recommendation, user behavior analysis and profiling, machine learning, and data mining. She is currently serving as Editor-in-Chief of ACM TOIS, and has served as co-chair, senoir PC member, or area chair for top conferences including WSDM, theWebConf, SIGIR, and CIKM.
\vspace{-10pt}
\end{IEEEbiography}
\begin{IEEEbiography}[{\includegraphics[width=1in,height=1.25in,clip,keepaspectratio]{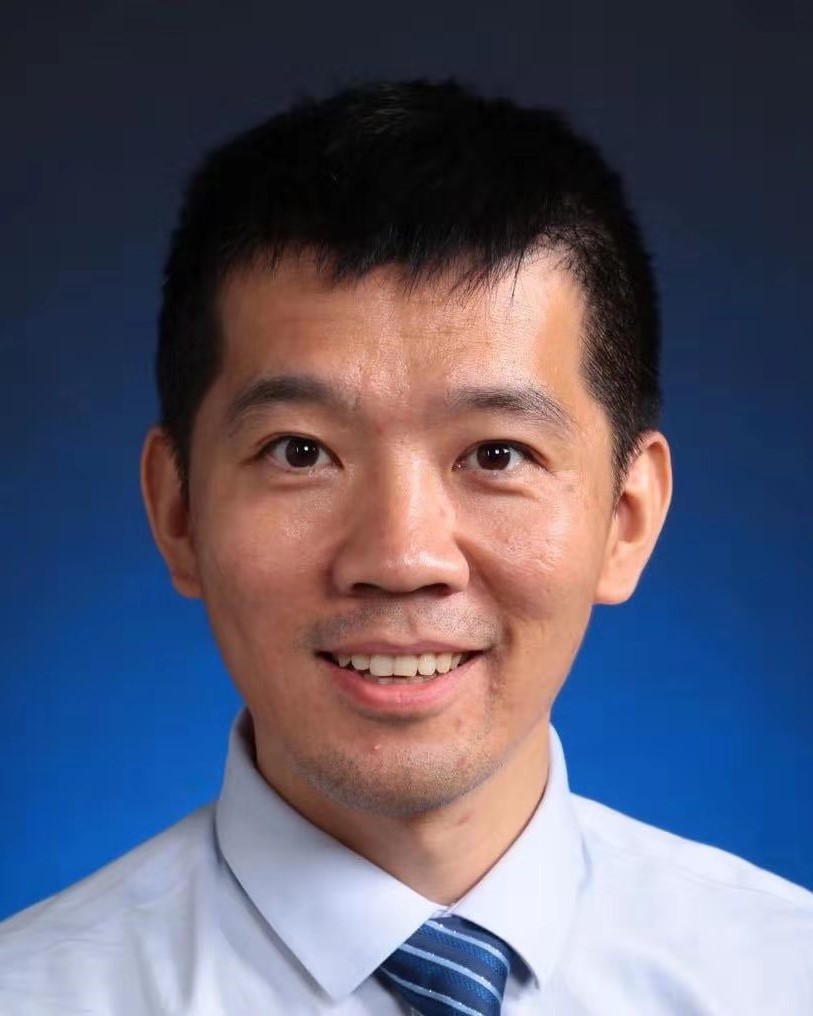}}]{Yiqun Liu} is working as a professor and department co-chair at Dept. of Computer Science and Technology in Tsinghua University. He received his Ph.D. in Tsinghua University in 2007. His major research interests are in web search, user behavior analysis, and information retrieval. He serves as co-Editor-in-Chief of FnTIR and has served as co-chair, senior PC member, or area chair for top conferences including IJCAI, theWebConf, SIGIR, CIKM, and NTCIR.
\vspace{-10pt}
\end{IEEEbiography}
\begin{IEEEbiography}[{\includegraphics[width=1in,height=1.25in,clip,keepaspectratio]{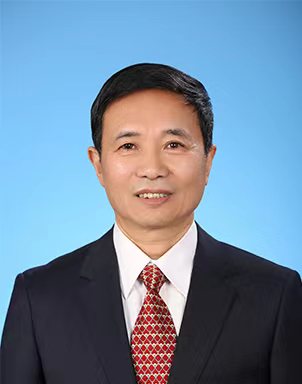}}]{Shaoping Ma} is a professor of Dept. of Computer Science and Technology in Tsinghua University. He received his Ph.D. in Tsinghua University in 1997. He devotes his life to exploring and researching intelligent information processing, including information retrieval, recommendation and machine learning. He has authored over 300 book chapters, journals, and top conference papers in his research area.
\vspace{-10pt}
\end{IEEEbiography}
\begin{IEEEbiography}[{\includegraphics[width=1in,height=1.25in,clip,keepaspectratio]{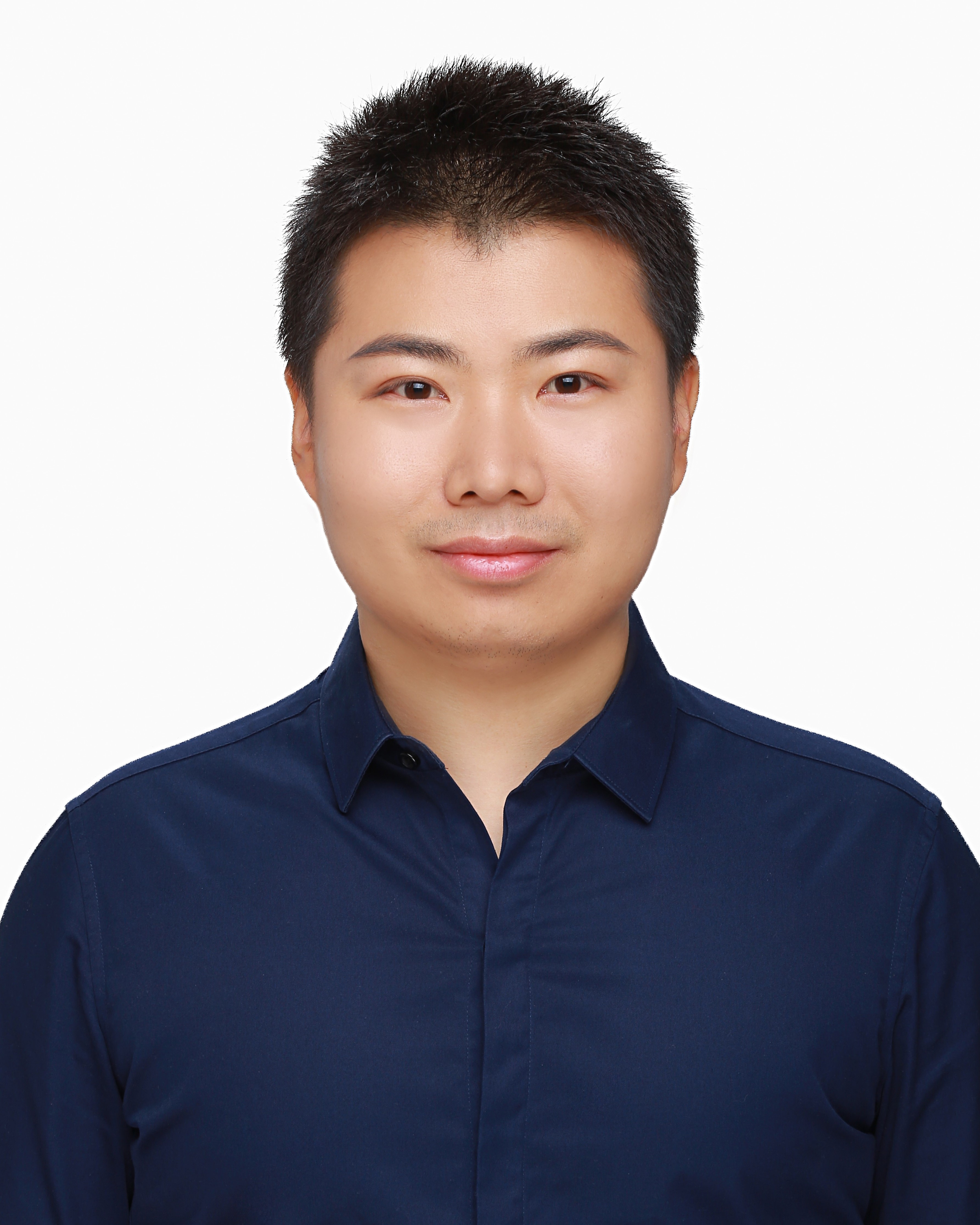}}]{Yuekui Yang} is currently pursuing an Eng.D. degree in Department of Computer Science and Technology, Tsinghua University. He received his B.E. from Taiyuan University of Technology in 2006. His research interests focus on recommendation systems and computational advertising. He is now working as a director in Tencent AI Platform Department.
\vspace{-10pt}
\end{IEEEbiography}

% if you will not have a photo at all:
% \begin{IEEEbiographynophoto}{John Doe}
% Biography text here.
% \end{IEEEbiographynophoto}

% insert where needed to balance the two columns on the last page with
% biographies
%\newpage

% \begin{IEEEbiographynophoto}{Jane Doe}
% Biography text here.
% \end{IEEEbiographynophoto}

% You can push biographies down or up by placing
% a \vfill before or after them. The appropriate
% use of \vfill depends on what kind of text is
% on the last page and whether or not the columns
% are being equalized.

%\vfill

% Can be used to pull up biographies so that the bottom of the last one
% is flush with the other column.
%\enlargethispage{-5in}

% if have a single appendix:
% \appendix[Experiment Data]\label{append:exp data}
% or
%\appendix  % for no appendix heading
% do not use \section anymore after \appendix, only \section*
% is possibly needed
% \input{data/appendix}

% use appendices with more than one appendix
% then use \section to start each appendix
% you must declare a \section before using any
% \subsection or using \label (\appendices by itself
% starts a section numbered zero.)
%

\newpage
\appendices
% !TeX root = ../main.tex

\section{Protocol used in the selection process of the articles in this survey}\label{append:protocol}
We searched with the query “dropout” on Google Scholar, selected the papers that were about dropout methods in neural models by checking the title and abstract of the articles, and we got around 200 articles. We further considered each article carefully, selecting the articles that met the following requirements:

$\bullet$ The article was published on top AI conferences or journals.

$\bullet$ The article had proposed new dropout methods, not just applied existing dropout methods.

$\bullet$ Some articles available on arxiv but not published on conferences or journals, which have proposed good dropout methods with some citations, were also included.

Then we checked the references of these articles, added the articles they cited and meeting the above requirements but missed by the search engine into our list. In this way, we finally got the current about 80 articles.

We sort the articles first according to our classification taxonomy. Within the same category, the articles are sorted mainly based on their publication date. Some highly related papers which are introduced together are placed in adjacent positions.

% !TeX root = ../main.tex

\section{Implementation Details of Dropout Methods on Recommendation Models}\label{append:implementation}

\begin{figure}
  \centering
  \includegraphics[width=0.75\linewidth]{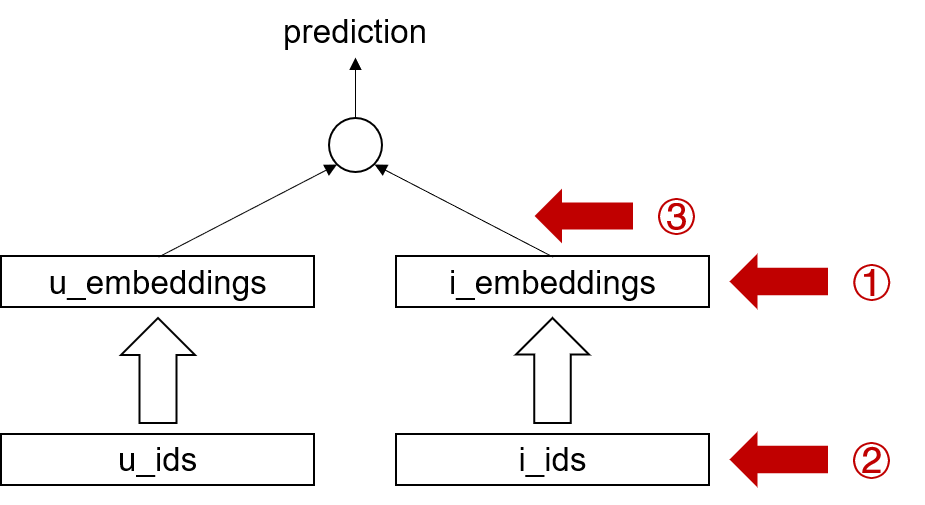}
  \caption{Dropout on BPR.}
  \label{fig:BPR_1}
\end{figure}

We first introduce the criterion we adopt to select the implementations of dropout methods, then we introduce the implementation details of dropout methods on each recommendation model.

\subsection{Criterion of the Implementations}
For dropping model structures, we implement dropout according to the original papers of the recommendation models: NFM \cite{he2017neural}, GRU4Rec \cite{hidasi2015session}, SASRec \cite{kang2018self}, and LightGCN \cite{he2020lightgcn}. Because dropping model structure is a universal type of dropout, all the neural recommendation models have the corresponding implementations. The original paper of BPR \cite{rendle2009bpr} is published before dropout was proposed, so we randomly drops the parameters in BPR model, which is consistent with the definition of dropping model structure. Our implementation of dropping graph information is also according to the original paper of LightGCN, which referred its implementation to NGCF \cite{wang2019neural}.

For dropping embeddings, we implement dropout according to the original papers (ACCM \cite{shi2018attention} and AFS \cite{shi2019adaptive}) where embedding dropout was first proposed as a formal method.

For dropping input information, we implement dropout according to the definition of “input information” in recommendation models, i.e., user ids, item ids, and features.

Following subsections are the implementation details of dropout methods on each recommendation model.

\subsection{Dropout on BPR}

\noindent $\bullet$ \textbf{Drop model structure}: standard dropout, achieved by adding a dropout layer after the user and item embedding matrix.

\noindent $\bullet$ \textbf{Drop input information}: randomly set some of the user and item numbers in each batch to random numbers.

\noindent $\bullet$ \textbf{Drop embedding}: randomly set the user and item embeddings in each batch to random embeddings.

The schematic diagram is shown in figure \ref{fig:BPR_1}. The \ding{172} in the figure indicates the dropout of model structure, \ding{173} indicates the dropout of input information, and \ding{174} indicates the dropout of embeddings.

\subsection{Dropout on NFM}

\begin{figure}
  \centering
  \includegraphics[width=0.7\linewidth]{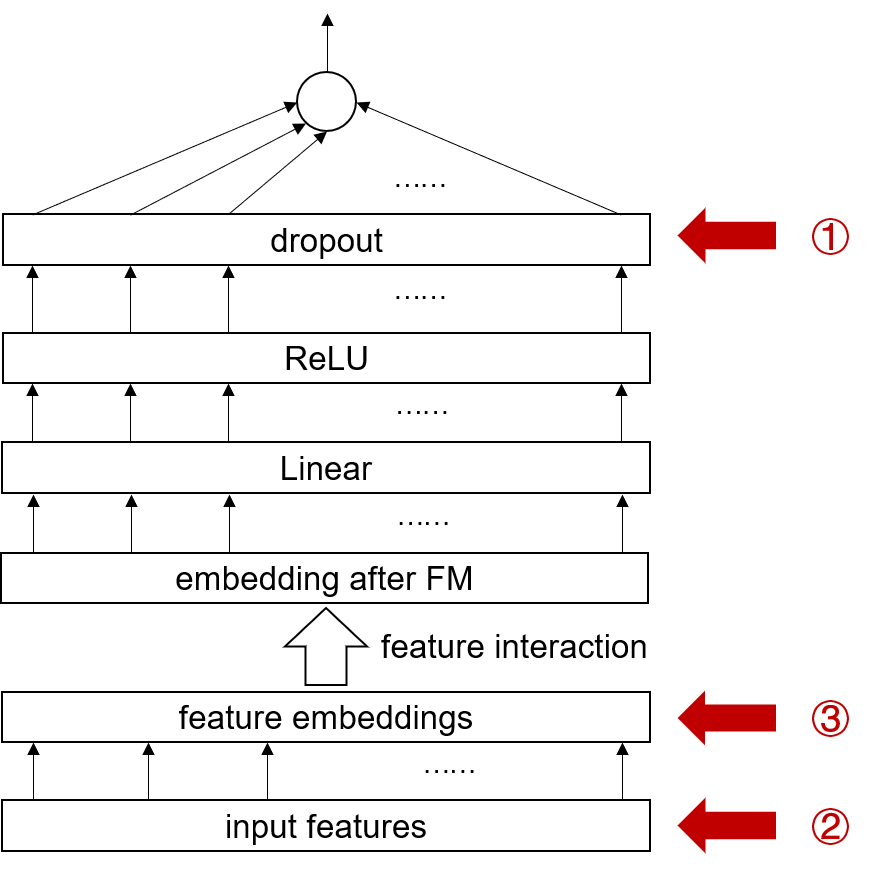}
  \caption{Dropout on NFM.}
  \label{fig:NFM_1}
\end{figure}

\noindent $\bullet$ \textbf{Drop model structure}: standard dropout, achieved by adding a dropout layer after the ReLU layer in the deep part.

\noindent $\bullet$ \textbf{Drop input information}: randomly drops the attribute information of users and items in each batch by setting them to random values.

\noindent $\bullet$ \textbf{Drop embedding}: randomly drops the embeddings corresponding to the attribute information of users and items in each batch by setting them to random embeddings.

The schematic diagram is shown in figure \ref{fig:NFM_1}. The \ding{172} in the figure indicates the dropout of model structure, \ding{173} indicates the dropout of input information, and \ding{174} indicates the dropout of embeddings.

\subsection{Dropout on GRU4Rec}

\begin{figure}
  \centering
  \includegraphics[width=\linewidth]{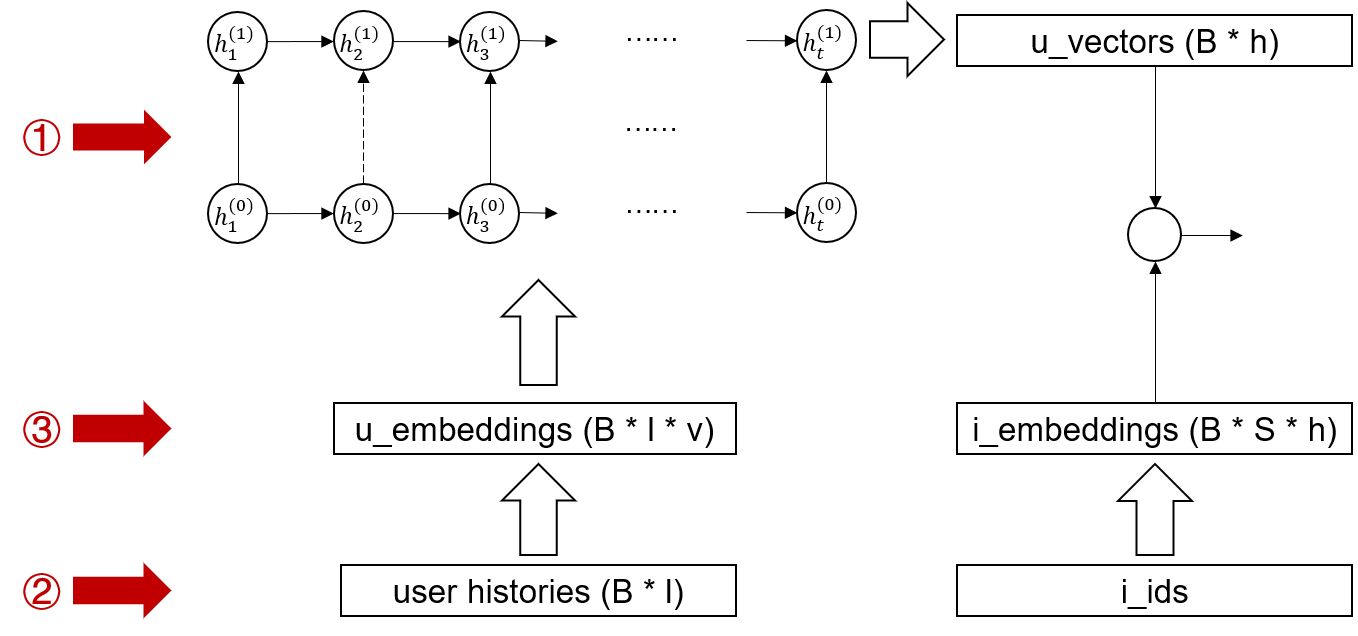}
  \caption{Dropout on GRU4Rec.}
  \label{fig:GRU4Rec_1}
\end{figure}

\noindent $\bullet$ \textbf{Drop model structure}: randomly dropout of feed-forward connections, following Zaremba et al. \cite{zaremba2014recurrent}.

\noindent $\bullet$ \textbf{Drop input information}: randomly set some of the user and item numbers in each batch to random numbers.

\noindent $\bullet$ \textbf{Drop embedding}: randomly set the user and item embeddings in each batch to random embeddings.

The schematic diagram is shown in figure \ref{fig:GRU4Rec_1}. The \ding{172} in the figure indicates the dropout of model structure, \ding{173} indicates the dropout of input information, and \ding{174} indicates the dropout of embeddings.

\subsection{Dropout on SASRec}

\begin{figure}
  \centering
  \includegraphics[width=\linewidth]{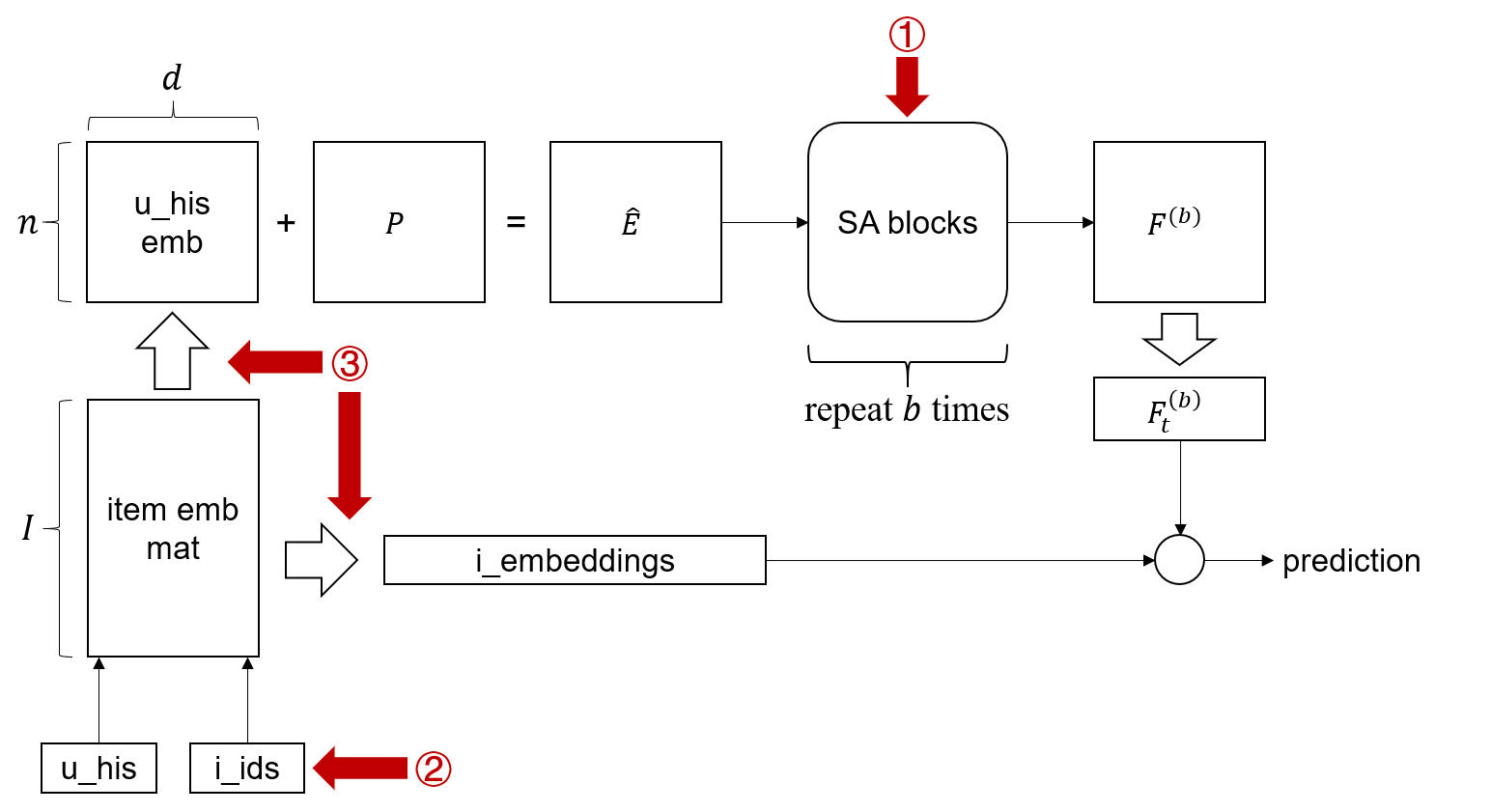}
  \caption{Dropout on SASRec.}
  \label{fig:SASRec_1}
\end{figure}

\noindent $\bullet$ \textbf{Drop model structure}: adding dropout layers within the self-attentive block to dropout neuron outputs, following the original article \cite{kang2018self}.

\noindent $\bullet$ \textbf{Drop input information}: randomly set some of the user and item numbers in each batch to random numbers.

\noindent $\bullet$ \textbf{Drop embedding}: randomly set the user and item embeddings in each batch to random embeddings.

The schematic diagram is shown in figure \ref{fig:SASRec_1}. The \ding{172} in the figure indicates the dropout of model structure, \ding{173} indicates the dropout of input information, and \ding{174} indicates the dropout of embeddings.

\subsection{Dropout on LightGCN}

\begin{figure}
  \centering
  \includegraphics[width=0.8\linewidth]{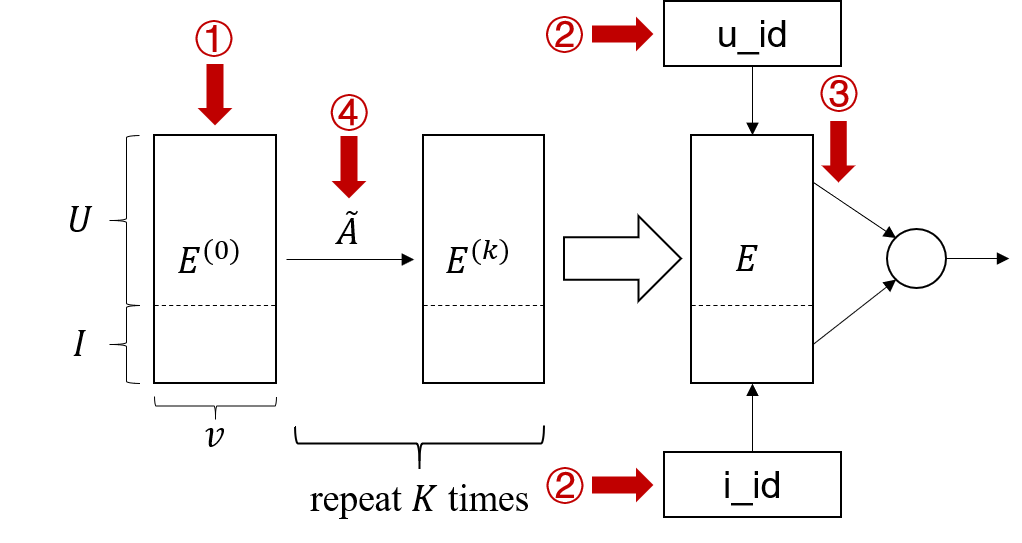}
  \caption{Dropout on LightGCN.}
  \label{fig:LightGCN_1}
\end{figure}

\noindent $\bullet$ \textbf{Drop model structure}: add a dropout layer after the user-item embedding matrix.

\noindent $\bullet$ \textbf{Drop input information}: randomly set some of the user and item numbers in each batch to random numbers.

\noindent $\bullet$ \textbf{Drop embedding}: randomly set the user and item embeddings in each batch to random embeddings.

\noindent $\bullet$ \textbf{Drop graph information}: for each batch, randomly drop some edges in the graph. This can be achieved by randomly dropping elements of the symmetrically normalized adjacency matrix $\Tilde{\mathbf{A}}$ \cite{he2020lightgcn}.

The schematic is shown in figure \ref{fig:LightGCN_1}. The \ding{172} in the figure indicates the dropout of model structure, \ding{173} indicates the dropout of input information, \ding{174} indicates the dropout of embeddings, and \ding{175} indicates the dropout of the edges.

% !TeX root = ../main.tex

\section{Experimental Settings and Model Parameters}\label{append:model_param}

Parameter values taken for all models in common are shown in Table \ref{tab:ExperimentSettings}. Parameters specific to each model are shown in Table \ref{tab:ExperimentSettings2}.

\begin{table}[ht]
\vspace{-4pt}
\centering
\caption{Global Parameters}
\vspace{-4pt}
\begin{tabular}{lll}
\toprule
Parameter           & Value            \\
\midrule
Learning rate                   & 0.001                         \\
Optimizer                       & Adam                          \\
Batch size                      & 128                           \\
Early stop                      & 50                            \\
Validation metrics              & NDCG@10                       \\
% \rule{-2.3pt}{9pt}
Evaluation metrics                    & NDCG@5,10,20,50; HR@10,20 \\
% \multirow{2}{*}{test metrics}   & NDCG@5,10,20,50;\\
%                                 & HR@10,20 \\
% \rule{-2.3pt}{9pt}
Neg. sample during training & 1                             \\
Neg. sample during testing  & all\cite{krichene2020sampled}       \\
Embedding size                  & 64                           \\
Loss function & BPR loss\cite{rendle2009bpr} \\
\bottomrule
\end{tabular}
\label{tab:ExperimentSettings}
\vspace{-2pt}
\end{table}

\begin{table*}[ht]
\vspace{-4pt}
\centering
\caption{Parameters specific to each model}
\vspace{-4pt}
\begin{tabular}{ccc}
\toprule
Model                       & Parameter         & Value   \\
\midrule
\multirow{2}{*}{NFM}     & Number of layers  & 1    \\
                         & Hidden state size & 64   \\
\rule{-2.3pt}{9pt}
\multirow{3}{*}{GRU4Rec} & User history length     & 20   \\
                         & Number of layers         & 2    \\
                         & Hidden vector size      & 64   \\
\rule{-2.3pt}{9pt}
\multirow{2}{*}{SASRec}  & User history length     & 20   \\
                         & Number of self-attention heads   & 1    \\
\rule{-2.3pt}{9pt}
LightGCN                 & Number of layers       & 3   \\
\bottomrule
\end{tabular}
\label{tab:ExperimentSettings2}
\vspace{-2pt}
\end{table*}

We grid search all main parameters and choose the parameter set that achieves the best performance on validation set. We use leave-one-out strategy to get the validation set and the test set. We adopt early stop, stopping training when model performance has not increased on validation set for 50 epochs. We run each experiment for ten times with ten random seeds, and do significance test using t-test.

% !TeX root = ../main.tex

\section{Experiment Data}\label{append:exp data}

\begin{table*}[ht]\normalsize
\renewcommand{\arraystretch}{1.1}
\centering
\caption{Detailed results for dropout methods on BPRMF, ml1m-cold dataset.}
\begin{threeparttable}
\begin{tabular}{ccllllll}
\toprule
\makecell[c]{Dropout\\Methods} & \makecell[c]{Dropout\\Ratio} & \makecell[c]{NDCG\\@5} & \makecell[c]{NDCG\\@10} & \makecell[c]{NDCG\\@20} & \makecell[c]{NDCG\\@50} & \makecell[c]{HR\\@10} & \makecell[c]{HR\\@20} \\
\midrule
Origin & \multicolumn{1}{c|}{-} & 0.0251 & 0.0339   & 0.0458   & 0.0667   & 0.0678   & 0.1155   \\ \cline{3-8}
\rule{-2.3pt}{12pt}
\multirow{3}{*}{\makecell[c]{Drop Model\\Structure}}   & \multicolumn{1}{c|}{0.1} & 0.0258   & 0.0355*  & 0.0479*  & 0.0697** & 0.0718*  & 0.1213*  \\
                             & \multicolumn{1}{c|}{0.2} & \textbf{0.0267}*  & 0.0359** & 0.0487** & \textbf{0.0709}** & 0.0715** & \textbf{0.1225}** \\
                             & \multicolumn{1}{c|}{0.3} & \textbf{0.0267}** & \textbf{0.0364}** & \textbf{0.0488}** & 0.0701** & \textbf{0.0721}** & 0.1220** \\
\multirow{3}{*}{\makecell[c]{Drop Input\\Info}}    & \multicolumn{1}{c|}{0.1} & 0.0233** & 0.0324*  & 0.0444*  & 0.0658   & 0.0668   & 0.1148   \\
                             & \multicolumn{1}{c|}{0.2} & 0.0241*  & 0.0331   & 0.0448   & 0.0660   & 0.0671   & 0.1138   \\
                             & \multicolumn{1}{c|}{0.3} & 0.0217** & 0.0294** & 0.0394** & 0.0573** & 0.0586** & 0.0988** \\
\multirow{3}{*}{\makecell[c]{Drop\\Embedding}} & \multicolumn{1}{c|}{0.1} & 0.0236*  & 0.0324*  & 0.0450   & 0.0660   & 0.0658   & 0.1162   \\
                             & \multicolumn{1}{c|}{0.2} & 0.0236*  & 0.0330   & 0.0451   & 0.0660   & 0.0672   & 0.1155  \\
                             & \multicolumn{1}{c|}{0.3} & 0.0241   & 0.0332   & 0.0453   & 0.0658   & 0.0672   & 0.1154 \\
\bottomrule
\end{tabular}
\begin{tablenotes}
\footnotesize
\item[]* for $p<0.05$, ** for $p<0.01$, compared to the origin (not using any dropout methods). Bold numbers are the best results of each column.
\end{tablenotes}
\end{threeparttable}
\label{tab:ml1m-cold-BPR}
\end{table*}

\begin{table*}[ht]\normalsize
\centering
\caption{Detailed results for dropout methods on BPRMF, Amazon Baby 5-core dataset.}
\begin{threeparttable}
\begin{tabular}{ccllllll}
\toprule
\makecell[c]{Dropout\\Methods} & \makecell[c]{Dropout\\Ratio}                      & \makecell[c]{NDCG\\@5}   & \makecell[c]{NDCG\\@10}  & \makecell[c]{NDCG\\@20}  & \makecell[c]{NDCG\\@50}  & \makecell[c]{HR\\@10}    & \makecell[c]{HR\\@20}    \\
\midrule
Origin                          & \multicolumn{1}{c|}{-}    & 0.00709   & 0.00969   & 0.01300   & 0.01936   & 0.01942   & 0.03263   \\ \cline{3-8}
\rule{-2.3pt}{12pt}
    \multirow{3}{*}{\makecell[c]{Drop Model\\Structure}}   & \multicolumn{1}{c|}{0.1} & 0.00693   & 0.00935  & 0.01253  & 0.01946 & 0.01872  & 0.03145  \\
                                & \multicolumn{1}{c|}{0.2} & \textbf{0.00723}  & \textbf{0.00973} & \textbf{0.01311} & \textbf{0.01991} & \textbf{0.01943} & 0.03292 \\
                                & \multicolumn{1}{c|}{0.3} & 0.00706 & 0.00959 & \textbf{0.01311} & 0.01984* & 0.01907 & \textbf{0.03315}\\
    \multirow{3}{*}{\makecell[c]{Drop Input\\Info}}    & \multicolumn{1}{c|}{0.1} & 0.00631*  & 0.00888** & 0.01229* & 0.01892 & 0.01839* & 0.03202   \\
                                & \multicolumn{1}{c|}{0.2} & 0.00610**  & 0.00839** & 0.01180** & 0.01836** & 0.01720** & 0.03078*   \\
                                & \multicolumn{1}{c|}{0.3} & 0.00544** & 0.00761** & 0.01077** & 0.01671** & 0.01581** & 0.02844**\\
    \multirow{3}{*}{\makecell[c]{Drop\\Embedding}} & \multicolumn{1}{c|}{0.1} & 0.00664*  & 0.00916* & 0.01271 & 0.01933 & 0.01864 & 0.03283   \\
                                & \multicolumn{1}{c|}{0.2} & 0.00641**  & 0.00900** & 0.01241** & 0.01943 & 0.01868* & 0.03231  \\
                                & \multicolumn{1}{c|}{0.3} & 0.00661 & 0.00911* & 0.01241* & 0.01914 & 0.01862 & 0.03177\\
\bottomrule
\end{tabular}
\begin{tablenotes}
\footnotesize
\item[]* for $p<0.05$, ** for $p<0.01$, compared to the origin (not using any dropout methods). Bold numbers are the best results of each column.
\end{tablenotes}
\end{threeparttable}
\label{tab:Amazon-Baby-5-core-BPR}
\end{table*}

\begin{table*}[ht]\normalsize
\centering
\caption{Detailed results for dropout methods on NFM, ml1m-cold dataset.}
\begin{threeparttable}
\begin{tabular}{ccllllll}
\toprule
\makecell[c]{Dropout\\Methods} & \makecell[c]{Dropout\\Ratio}                      & \makecell[c]{NDCG\\@5}   & \makecell[c]{NDCG\\@10}  & \makecell[c]{NDCG\\@20}  & \makecell[c]{NDCG\\@50}  & \makecell[c]{HR\\@10}    & \makecell[c]{HR\\@20}    \\
\midrule
Origin                          & \multicolumn{1}{c|}{-}    & 0.0246   & 0.0335   & 0.0454   & 0.0664   & 0.0680   & 0.1153   \\ \cline{3-8}
\rule{-2.3pt}{12pt}
    \multirow{3}{*}{\makecell[c]{Drop Model\\Structure}}   & \multicolumn{1}{c|}{0.1} & 0.0255   & 0.0350*  & 0.0472*  & 0.0682** & 0.0708*  & \textbf{0.1193}*  \\
                                & \multicolumn{1}{c|}{0.2} & \textbf{0.0260}*  & 0.0353* & 0.0473* & 0.0687** & 0.0704* & 0.1184 \\
                                & \multicolumn{1}{c|}{0.3} & 0.0257*  & 0.0352** & \textbf{0.0474}** & 0.0684** & 0.0705*  & 0.1191* \\
    \multirow{3}{*}{\makecell[c]{Drop Input\\Info}}    & \multicolumn{1}{c|}{0.1} & 0.0242  & 0.0334 & 0.0457 & 0.0678** & 0.0677 & 0.1167   \\
                                & \multicolumn{1}{c|}{0.2} & 0.0240  & 0.0332 & 0.0458 & 0.0675* & 0.0670 & 0.1172   \\
                                & \multicolumn{1}{c|}{0.3} & 0.0135** & 0.0184** & 0.0244** & 0.0371** & 0.0369** & 0.0608** \\
    \multirow{3}{*}{\makecell[c]{Drop\\Embedding}} & \multicolumn{1}{c|}{0.1} & 0.0259**  & \textbf{0.0354}** & 0.0473** & \textbf{0.0691}** & \textbf{0.0716}** & 0.1188*   \\
                                & \multicolumn{1}{c|}{0.2} & 0.0251  & 0.0345 & 0.0470 & 0.0686** & 0.0695 & \textbf{0.1193}*  \\
                                & \multicolumn{1}{c|}{0.3} & 0.0231** & 0.0326   & 0.0449   & 0.0666   & 0.0667   & 0.1159 \\
\bottomrule
\end{tabular}
\begin{tablenotes}
\footnotesize
\item[]* for $p<0.05$, ** for $p<0.01$, compared to the origin (not using any dropout methods). Bold numbers are the best results of each column.
\end{tablenotes}
\end{threeparttable}
\label{tab:ml1m-cold-NFM}
\end{table*}

\begin{table*}[ht]\normalsize
\centering
\caption{Detailed results for dropout methods on NFM, Amazon Baby 5-core dataset.}
\begin{threeparttable}
\begin{tabular}{ccllllll}
\toprule
\makecell[c]{Dropout\\Methods} & \makecell[c]{Dropout\\Ratio}                      & \makecell[c]{NDCG\\@5}   & \makecell[c]{NDCG\\@10}  & \makecell[c]{NDCG\\@20}  & \makecell[c]{NDCG\\@50}  & \makecell[c]{HR\\@10}    & \makecell[c]{HR\\@20}    \\
\midrule
Origin                          & \multicolumn{1}{c|}{-}    & 0.00458   & 0.00657   & 0.00926   & 0.01444   & 0.01376   & 0.02451   \\ \cline{3-8}
\rule{-2.3pt}{12pt}
    \multirow{3}{*}{\makecell[c]{Drop Model\\Structure}}   & \multicolumn{1}{c|}{0.1} & 0.00515*   & 0.00715*  & 0.01010**  & 0.01592** & 0.01496**  & 0.02677**  \\
                                & \multicolumn{1}{c|}{0.2} & \textbf{0.00518}*  & \textbf{0.00737}** & \textbf{0.01046}** & \textbf{0.01618}** & \textbf{0.01545}** & \textbf{0.02778}** \\
                                & \multicolumn{1}{c|}{0.3} & 0.00499 & 0.00705 & 0.00985 & 0.01534 & 0.01475 & 0.02588\\
    \multirow{3}{*}{\makecell[c]{Drop Input\\Info}}    & \multicolumn{1}{c|}{0.1} & 0.00460  & 0.00643 & 0.00898 & 0.01441 & 0.01357 & 0.02386   \\
                                & \multicolumn{1}{c|}{0.2} & 0.00465  & 0.00654 & 0.00923 & 0.01489 & 0.01388 & 0.02469   \\
                                & \multicolumn{1}{c|}{0.3} & 0.00468 & 0.00649 & 0.00899 & 0.01467 & 0.01379 & 0.02381\\
    \multirow{3}{*}{\makecell[c]{Drop\\Embedding}} & \multicolumn{1}{c|}{0.1} & 0.00424  & 0.00610 & 0.00858 & 0.01348 & 0.01264 & 0.02255   \\
                                & \multicolumn{1}{c|}{0.2} & 0.00460  & 0.00640 & 0.00892 & 0.01440 & 0.01352 & 0.02359  \\
                                & \multicolumn{1}{c|}{0.3} & 0.00473 & 0.00647 & 0.00913 & 0.01479 & 0.01369 & 0.02435\\
\bottomrule
\end{tabular}
\begin{tablenotes}
\footnotesize
\item[]* for $p<0.05$, ** for $p<0.01$, compared to the origin (not using any dropout methods). Bold numbers are the best results of each column.
\end{tablenotes}
\end{threeparttable}
\label{tab:Amazon-Baby-5-core-NFM}
\end{table*}

\begin{table*}[ht]\normalsize
\centering
\caption{Detailed results for dropout methods on GRU4Rec, ml1m-cold dataset.}
\begin{threeparttable}
\begin{tabular}{ccllllll}
\toprule
\makecell[c]{Dropout\\Methods} & \makecell[c]{Dropout\\Ratio}                      & \makecell[c]{NDCG\\@5}   & \makecell[c]{NDCG\\@10}  & \makecell[c]{NDCG\\@20}  & \makecell[c]{NDCG\\@50}  & \makecell[c]{HR\\@10}    & \makecell[c]{HR\\@20}    \\
\midrule
Origin                          & \multicolumn{1}{c|}{-}    & 0.0752   & 0.0964   & 0.1196   & 0.1496   & 0.1818   & 0.2739   \\ \cline{3-8}
\rule{-2.3pt}{12pt}
    \multirow{3}{*}{\makecell[c]{Drop Model\\Structure}}   & \multicolumn{1}{c|}{0.1} & 0.0782*   & 0.1003**  & 0.1233**  & 0.1536** & 0.1892**  & 0.2805**  \\
                                & \multicolumn{1}{c|}{0.2} & 0.0792**  & 0.1010** & 0.1245** & 0.1544** & 0.1895** & 0.2830** \\
                                & \multicolumn{1}{c|}{0.3} & 0.0780*  & 0.1004** & 0.1239** & 0.1538** & 0.1902** & 0.2834** \\
    \multirow{3}{*}{\makecell[c]{Drop Input\\Info}}    & \multicolumn{1}{c|}{0.1} & \textbf{0.0859}**  & \textbf{0.1084}** & \textbf{0.1323}** & \textbf{0.1625}** & \textbf{0.2012}** & \textbf{0.2957}**   \\
                                & \multicolumn{1}{c|}{0.2} & 0.0829**  & 0.1056** & 0.1297** & 0.1598** & 0.1973** & 0.2931**   \\
                                & \multicolumn{1}{c|}{0.3} & 0.0811** & 0.1024** & 0.1255** & 0.1555** & 0.1904** & 0.2818** \\
    \multirow{3}{*}{\makecell[c]{Drop\\Embedding}} & \multicolumn{1}{c|}{0.1} & 0.0791**  & 0.1013** & 0.1246** & 0.1552** & 0.1911** & 0.2840**   \\
                                & \multicolumn{1}{c|}{0.2} & 0.0784**  & 0.1012** & 0.1245** & 0.1554** & 0.1925** & 0.2852**  \\
                                & \multicolumn{1}{c|}{0.3} & 0.0779*  & 0.1007** & 0.1248** & 0.1557** & 0.1920** & 0.2875** \\
\bottomrule
\end{tabular}
\begin{tablenotes}
\footnotesize
\item[]* for $p<0.05$, ** for $p<0.01$, compared to the origin (not using any dropout methods). Bold numbers are the best results of each column.
\end{tablenotes}
\end{threeparttable}
\label{tab:ml1m-cold-GRU4Rec}
\end{table*}

\begin{table*}[ht]\normalsize
\centering
\caption{Detailed results for dropout methods on GRU4Rec, Amazon Baby 5-core dataset.}
\begin{threeparttable}
\begin{tabular}{ccllllll}
\toprule
\makecell[c]{Dropout\\Methods} & \makecell[c]{Dropout\\Ratio}                      & \makecell[c]{NDCG\\@5}   & \makecell[c]{NDCG\\@10}  & \makecell[c]{NDCG\\@20}  & \makecell[c]{NDCG\\@50}  & \makecell[c]{HR\\@10}    & \makecell[c]{HR\\@20}    \\
\midrule
Origin                          & \multicolumn{1}{c|}{-}    & 0.00762   & 0.01071   & 0.01487   & 0.02261   & 0.02199   & 0.03856   \\ \cline{3-8}
\rule{-2.3pt}{12pt}
    \multirow{3}{*}{\makecell[c]{Drop Model\\Structure}}   & \multicolumn{1}{c|}{0.1} & 0.00786   & 0.01132*  & 0.01555*  & 0.02337* & 0.02347**  & 0.04036  \\
                                & \multicolumn{1}{c|}{0.2} & 0.00810  & 0.01146 & 0.01579* & 0.02342 & 0.02374* & 0.04101* \\
                                & \multicolumn{1}{c|}{0.3} & 0.00800 & 0.01138* & 0.01565** & 0.02348* & 0.02340** & 0.04043**\\
    \multirow{3}{*}{\makecell[c]{Drop Input\\Info}}    & \multicolumn{1}{c|}{0.1} & 0.01013**  & 0.01383** & 0.01864** & 0.02698** & 0.02781** & 0.04700**   \\
                                & \multicolumn{1}{c|}{0.2} & 0.01185**  & 0.01622** & 0.02170** & 0.03073** & 0.03257** & 0.05440**   \\
                                & \multicolumn{1}{c|}{0.3} & \textbf{0.01318}** & \textbf{0.01777}** & \textbf{0.02343}** & \textbf{0.03301}** & \textbf{0.03543}** & \textbf{0.05797}**\\
    \multirow{3}{*}{\makecell[c]{Drop\\Embedding}} & \multicolumn{1}{c|}{0.1} & 0.00817  & 0.01129 & 0.01556 & 0.02327 & 0.02311 & 0.04010   \\
                                & \multicolumn{1}{c|}{0.2} & 0.00792  & 0.01109 & 0.01536 & 0.02311 & 0.02294 & 0.03991  \\
                                & \multicolumn{1}{c|}{0.3} & 0.00785 & 0.01096 & 0.01514 & 0.02275 & 0.02255 & 0.03926\\
\bottomrule
\end{tabular}
\begin{tablenotes}
\footnotesize
\item[]* for $p<0.05$, ** for $p<0.01$, compared to the origin (not using any dropout methods). Bold numbers are the best results of each column.
\end{tablenotes}
\end{threeparttable}
\label{tab:Amazon-Baby-5-core-GRU4Rec}
\end{table*}

\begin{table*}[ht]\normalsize
\centering
\caption{Detailed results for dropout methods on SASRec, ml1m-cold dataset.}
\begin{threeparttable}
\begin{tabular}{ccllllll}
\toprule
\makecell[c]{Dropout\\Methods} & \makecell[c]{Dropout\\Ratio}                      & \makecell[c]{NDCG\\@5}   & \makecell[c]{NDCG\\@10}  & \makecell[c]{NDCG\\@20}  & \makecell[c]{NDCG\\@50}  & \makecell[c]{HR\\@10}    & \makecell[c]{HR\\@20}    \\
\midrule
Origin                          & \multicolumn{1}{c|}{-}    & 0.0840   & 0.1064   & 0.1296   & 0.1593   & 0.1981   & 0.2903   \\ \cline{3-8}
\rule{-2.3pt}{12pt}
    \multirow{3}{*}{\makecell[c]{Drop Model\\Structure}}   & \multicolumn{1}{c|}{0.1} & 0.0864*   & 0.1092**  & 0.1326*  & 0.1627** & 0.2013  & 0.2941  \\
                                & \multicolumn{1}{c|}{0.2} & 0.0852  & 0.1077 & 0.1308 & 0.1606 & 0.1996 & 0.2912 \\
                                & \multicolumn{1}{c|}{0.3} & 0.0836   & 0.1059   & 0.1290   & 0.1585   & 0.1961   & 0.2878 \\
    \multirow{3}{*}{\makecell[c]{Drop Input\\Info}}    & \multicolumn{1}{c|}{0.1} & \textbf{0.0868}*  & \textbf{0.1093} & \textbf{0.1330}* & \textbf{0.1632}* & \textbf{0.2019} & \textbf{0.2957}*   \\
                                & \multicolumn{1}{c|}{0.2} & 0.0816*  & 0.1041* & 0.1282 & 0.1589 & 0.1942* & 0.2901   \\
                                & \multicolumn{1}{c|}{0.3} & 0.0705** & 0.0915** & 0.1142** & 0.1445** & 0.1742** & 0.2645** \\
    \multirow{3}{*}{\makecell[c]{Drop\\Embedding}} & \multicolumn{1}{c|}{0.1} & 0.0835  & 0.1063 & 0.1301 & 0.1606 & 0.1980 & 0.2923   \\
                                & \multicolumn{1}{c|}{0.2} & 0.0814*  & 0.1042 & 0.1278 & 0.1588 & 0.1962 & 0.2901  \\
                                & \multicolumn{1}{c|}{0.3} & 0.0797*  & 0.1022*  & 0.1259*  & 0.1572   & 0.1935   & 0.2876 \\
\bottomrule
\end{tabular}
\begin{tablenotes}
\footnotesize
\item[]* for $p<0.05$, ** for $p<0.01$, compared to the origin (not using any dropout methods). Bold numbers are the best results of each column.
\end{tablenotes}
\end{threeparttable}
\label{tab:ml1m-cold-SASRec}
\end{table*}

\begin{table*}[ht]\normalsize
\centering
\caption{Detailed results for dropout methods on SASRec, Amazon Baby 5-core dataset.}
\begin{threeparttable}
\begin{tabular}{ccllllll}
\toprule
\makecell[c]{Dropout\\Methods} & \makecell[c]{Dropout\\Ratio}                      & \makecell[c]{NDCG\\@5}   & \makecell[c]{NDCG\\@10}  & \makecell[c]{NDCG\\@20}  & \makecell[c]{NDCG\\@50}  & \makecell[c]{HR\\@10}    & \makecell[c]{HR\\@20}    \\
\midrule
Origin                          & \multicolumn{1}{c|}{-}    & 0.01039   & 0.01428   & 0.01913   & 0.02787   & 0.02889   & 0.04824   \\ \cline{3-8}
\rule{-2.3pt}{12pt}
    \multirow{3}{*}{\makecell[c]{Drop Model\\Structure}}   & \multicolumn{1}{c|}{0.1} & 0.01048   & 0.01468  & 0.01990  & 0.02873 & 0.03016  & 0.05100*  \\
                                & \multicolumn{1}{c|}{0.2} & 0.01140**  & 0.01542* & 0.02067** & 0.02961** & 0.03076* & 0.05170** \\
                                & \multicolumn{1}{c|}{0.3} & 0.01145* & 0.01562* & 0.02095** & 0.03004** & 0.03154* & 0.05277**\\
    \multirow{3}{*}{\makecell[c]{Drop Input\\Info}}    & \multicolumn{1}{c|}{0.1} & 0.01308**  & 0.01783** & 0.02375** & 0.03326** & 0.03600** & 0.05961**   \\
                                & \multicolumn{1}{c|}{0.2} & 0.01595**  & 0.02119** & 0.02714** & 0.03750** & 0.04182** & 0.06563**   \\
                                & \multicolumn{1}{c|}{0.3} & \textbf{0.01612}** & \textbf{0.02143}** & \textbf{0.02768}** & \textbf{0.03828}** & \textbf{0.04207}** & \textbf{0.06701}**\\
    \multirow{3}{*}{\makecell[c]{Drop\\Embedding}} & \multicolumn{1}{c|}{0.1} & 0.01045  & 0.01440 & 0.01964 & 0.02849 & 0.02934 & 0.05020*   \\
                                & \multicolumn{1}{c|}{0.2} & 0.01071  & 0.01484 & 0.01985 & 0.02904 & 0.03010 & 0.05008  \\
                                & \multicolumn{1}{c|}{0.3} & 0.01094 & 0.01502 & 0.02009 & 0.02950* & 0.03030 & 0.05056*\\
\bottomrule
\end{tabular}
\begin{tablenotes}
\footnotesize
\item[]* for $p<0.05$, ** for $p<0.01$, compared to the origin (not using any dropout methods). Bold numbers are the best results of each column.
\end{tablenotes}
\end{threeparttable}
\label{tab:Amazon-Baby-5-core-SASRec}
\end{table*}

\begin{table*}[ht]\normalsize
\centering
\caption{Detailed results for dropout methods on LightGCN, ml1m-cold dataset.}
\begin{threeparttable}
\begin{tabular}{ccllllll}
\toprule
\makecell[c]{Dropout\\Methods} & \makecell[c]{Dropout\\Ratio}                      & \makecell[c]{NDCG\\@5}   & \makecell[c]{NDCG\\@10}  & \makecell[c]{NDCG\\@20}  & \makecell[c]{NDCG\\@50}  & \makecell[c]{HR\\@10}    & \makecell[c]{HR\\@20}    \\
\midrule
Origin                          & \multicolumn{1}{c|}{-}    & 0.0281   & 0.0377   & 0.0504   & 0.0722   & 0.0748   & 0.1255   \\ \cline{3-8}
\rule{-2.3pt}{12pt}
    \multirow{3}{*}{\makecell[c]{Drop Model\\Structure}}   & \multicolumn{1}{c|}{0.1} & 0.0287   & 0.0385  & 0.0510  & 0.0732 & 0.0761  & 0.1261  \\
                                & \multicolumn{1}{c|}{0.2} & \textbf{0.0291}*  & \textbf{0.0388} & \textbf{0.0516} & \textbf{0.0737}* & \textbf{0.0765} & \textbf{0.1277} \\
                                & \multicolumn{1}{c|}{0.3} & 0.0286 & 0.0384 & 0.0513 & 0.0730 & 0.0757 & 0.1273 \\
    \multirow{3}{*}{\makecell[c]{Drop Input\\Info}}    & \multicolumn{1}{c|}{0.1} & 0.0264**  & 0.0361** & 0.0484** & 0.0705** & 0.0721** & 0.1213**   \\
                                & \multicolumn{1}{c|}{0.2} & 0.0254**  & 0.0347** & 0.0466** & 0.0674** & 0.0690** & 0.1163**   \\
                                & \multicolumn{1}{c|}{0.3} & 0.0239** & 0.0328** & 0.0443** & 0.0642** & 0.0653** & 0.1113** \\
    \multirow{3}{*}{\makecell[c]{Drop\\Embedding}}    & \multicolumn{1}{c|}{0.1} & 0.0280  & 0.0376 & 0.0505 & 0.0723 & 0.0748 & 0.1259   \\
                                & \multicolumn{1}{c|}{0.2} & 0.0272  & 0.0371 & 0.0497 & 0.0709 & 0.0741 & 0.1242   \\
                                & \multicolumn{1}{c|}{0.3} & 0.0264* & 0.0361** & 0.0482** & 0.0696** & 0.0724* & 0.1205* \\
    \multirow{3}{*}{\makecell[c]{Drop Graph\\Info}} & \multicolumn{1}{c|}{0.1} & 0.0282  & 0.0380 & 0.0508 & 0.0723 & 0.0754 & 0.1265   \\
                                & \multicolumn{1}{c|}{0.2} & 0.0284  & 0.0383 & 0.0511 & 0.0729 & 0.0764 & 0.1273  \\
                                & \multicolumn{1}{c|}{0.3} & 0.0283 & 0.0380 & 0.0509 & 0.0727 & 0.0756 & 0.1269 \\
\bottomrule
\end{tabular}
\begin{tablenotes}
\footnotesize
\item[]* for $p<0.05$, ** for $p<0.01$, compared to the origin (not using any dropout methods). Bold numbers are the best results of each column.
\end{tablenotes}
\end{threeparttable}
\label{tab:ml1m-cold-LightGCN}
\end{table*}

\begin{table*}[ht]\normalsize
\centering
\caption{Detailed results for dropout methods on LightGCN, Amazon Baby 5-core dataset.}
\begin{threeparttable}
\begin{tabular}{ccllllll}
\toprule
\makecell[c]{Dropout\\Methods} & \makecell[c]{Dropout\\Ratio}                      & \makecell[c]{NDCG\\@5}   & \makecell[c]{NDCG\\@10}  & \makecell[c]{NDCG\\@20}  & \makecell[c]{NDCG\\@50}  & \makecell[c]{HR\\@10}    & \makecell[c]{HR\\@20}    \\
\midrule
Origin                          & \multicolumn{1}{c|}{-}    & 0.01032   & 0.01392   & 0.01843   & 0.02738   & 0.02779   & 0.04577   \\ \cline{3-8}
\rule{-2.3pt}{12pt}
    \multirow{3}{*}{\makecell[c]{Drop Model\\Structure}}   & \multicolumn{1}{c|}{0.1} & 0.01017   & 0.01392  & 0.01835  & 0.02743 & 0.02809  & 0.04577  \\
                                & \multicolumn{1}{c|}{0.2} & 0.01001  & 0.01370 & 0.01807 & 0.02730 & 0.02763 & 0.04502 \\
                                & \multicolumn{1}{c|}{0.3} & 0.01001 & 0.01357 & 0.01811 & 0.02703 & 0.02693 & 0.04509\\
    \multirow{3}{*}{\makecell[c]{Drop Input\\Info}}    & \multicolumn{1}{c|}{0.1} & 0.00936**  & 0.01275** & 0.01710** & 0.02598** & 0.02559** & 0.04301**   \\
                                & \multicolumn{1}{c|}{0.2} & 0.00872**  & 0.01207** & 0.01651** & 0.02532** & 0.02435** & 0.04209**   \\
                                & \multicolumn{1}{c|}{0.3} & 0.00824** & 0.01140** & 0.01583** & 0.02456** & 0.02321** & 0.04095**\\
    \multirow{3}{*}{\makecell[c]{Drop\\Embedding}}    & \multicolumn{1}{c|}{0.1} & \textbf{0.01043}  & 0.01402 & 0.01866 & 0.02754 & 0.02781 & 0.04633   \\
                                & \multicolumn{1}{c|}{0.2} & 0.01034  & \textbf{0.01420} & \textbf{0.01869} & \textbf{0.02773}* & \textbf{0.02853} & \textbf{0.04651}   \\
                                & \multicolumn{1}{c|}{0.3} & 0.01029 & 0.01389 & 0.01841 & 0.02757 & 0.02798 & 0.04595\\
    \multirow{3}{*}{\makecell[c]{Drop Graph\\Info}} & \multicolumn{1}{c|}{0.1} & 0.01020  & 0.01403 & 0.01838 & 0.02733 & 0.02821 & 0.04552   \\
                                & \multicolumn{1}{c|}{0.2} & 0.01005*  & 0.01375 & 0.01811* & 0.02722 & 0.02766 & 0.04503  \\
                                & \multicolumn{1}{c|}{0.3} & 0.01021 & 0.01395 & 0.01843 & 0.02736 & 0.02799 & 0.04584\\
\bottomrule
\end{tabular}
\begin{tablenotes}
\footnotesize
\item[]* for $p<0.05$, ** for $p<0.01$, compared to the origin (not using any dropout methods). Bold numbers are the best results of each column.
\end{tablenotes}
\end{threeparttable}
\label{tab:Amazon-Baby-5-core-LightGCN}
\end{table*}

% \section{Proof of the First Zonklar Equation}
% Appendix one text goes here.

% you can choose not to have a title for an appendix
% if you want by leaving the argument blank
% \section{}
% Appendix two text goes here.

% that's all folks
\end{document}